\documentclass[review,12pt]{elsarticle}
\usepackage{amsthm}
\usepackage{lineno,hyperref}
\usepackage{moreverb}
\usepackage{graphicx,subfig,wrapfig}
\usepackage{algorithm}
\usepackage{algpseudocode}
\usepackage{color}
\usepackage{multirow}
\usepackage{amsmath}
\usepackage{amssymb}
\usepackage{txfonts}
\usepackage{afterpage}
\usepackage[margin=0.8in]{geometry}
\usepackage{mathrsfs}
\usepackage{tikz,amsmath}
\usepackage{pstricks}
\usepackage{diagbox}
\usepackage{arydshln}
\usepackage{dsfont}
\usepackage{smartdiagram}
\usepackage[normalem]{ulem}
\usepackage{todonotes}
\usepackage{booktabs}
\usepackage{appendix}
\usepackage{verbatim}

\usetikzlibrary{calc,trees,positioning,arrows,chains,shapes.geometric,%
    decorations.pathreplacing,decorations.pathmorphing,shapes,%
    matrix,shapes.symbols}

\tikzset{
>=stealth',
  punktchain/.style={
    rectangle, 
    rounded corners, 
    draw=black, very thick,
    text width=10em, 
    minimum height=2em, 
    text centered, 
    on chain},
  line/.style={draw, thick, <-},
  element/.style={
    tape,
    top color=white,
    bottom color=blue!50!black!60!,
    minimum width=6em,
    draw=blue!40!black!90, very thick,
    text width=10em, 
    minimum height=3.5em, 
    text centered, 
    on chain},
  every join/.style={->, thick,shorten >=1pt},
  decoration={brace},
  tuborg/.style={decorate},
  tubnode/.style={midway, right=2pt},
}

\tikzstyle{arrow} = [thick,->,>=stealth]

\newdefinition{rmk}{Remark}
\newproof{pf}{Proof}

\newcommand{\pB}{\mathfrak{b}}

\renewcommand{\mathbf}{\boldsymbol}
\newcommand{\xs}{\mathbb}

\newcommand{\comm}[1]{}

\makeatletter
\def\ps@pprintTitle{%
   \let\@oddhead\@empty
   \let\@evenhead\@empty
   \let\@oddfoot\@empty
   \let\@evenfoot\@oddfoot
}
\makeatother

\modulolinenumbers[5]

\journal{Journal of Computational Physics}

\begin{document}

\begin{frontmatter}
		
            \title{Dimension-reduced KRnet maps for high-dimensional Bayesian inverse problems}
		
		
		\author[mymainaddress]{Yani Feng}
		\ead{fengyn@shanghaitech.edu.cn}
		\author[mysecondaryaddress]{Kejun Tang}
		\ead{tangkejun@icode.pku.edu.cn}
		
		\author[mythirdaddress]{Xiaoliang Wan}
		\ead{xlwan@lsu.edu}
		
		\author[mymainaddress]{Qifeng Liao\corref{mycorrespondingauthor}}
		\cortext[mycorrespondingauthor]{Corresponding author}
		\ead{liaoqf@shanghaitech.edu.cn}
		
		\address[mymainaddress]{School of Information Science and Technology, ShanghaiTech University, Shanghai 201210, China}
		\address[mysecondaryaddress]{Changsha Institute for Computing and Digital Economy, Peking University, Changsha 410205, China}
		\address[mythirdaddress]{Department of Mathematics and Center for Computation 
			and Technology, 
			Louisiana State University, Baton Rouge 70803, USA}

		\begin{abstract}
We present a dimension-reduced KRnet map approach (DR-KRnet) for high-dimensional Bayesian inverse problems, which is based on an explicit construction of a map that pushes forward the prior measure to the posterior measure in the latent space. 
Our approach consists of two main components: data-driven VAE prior and density approximation of the posterior of the latent variable. 
In reality, it may not be trivial to initialize a prior distribution that is consistent with available prior data; in other words, the complex prior information is often beyond simple hand-crafted priors. 
We employ variational autoencoder (VAE) to approximate the underlying distribution of the prior dataset, which is achieved through a  latent variable and a decoder. Using the decoder provided by the VAE prior, we reformulate the problem in a low-dimensional latent space. In particular, we seek an invertible transport map given by KRnet to approximate the posterior distribution of the latent variable. 
Moreover, an efficient physics-constrained surrogate model without any labeled data is constructed to reduce the computational cost of solving both forward and adjoint problems involved in likelihood computation.
With numerical experiments, we demonstrate the accuracy and efficiency of DR-KRnet for high-dimensional Bayesian inverse problems.
		\end{abstract}
		
		
	\end{frontmatter}

\section{Introduction}\label{section_intro}
Bayesian inverse problems arise frequently in science and engineering, with applications ranging from subsurface and atmospheric transport to chemical kinetics. The primary task of such problems is to recover spatially varying unknown parameters from noisy and incomplete observations. Quantifying the uncertainty in the unknown parameters \cite{efendiev2006preconditioning,marzouk2007stochastic,wang2016gaussian,huan2013simulation,li2015adaptive,lieberman2010parameter,cui2016dimension} is then essential for predictive modeling and simulation-based decision-making.

The Bayesian statistical approach provides a foundation for inference from data and past knowledge. Indeed, the Bayesian setting casts the inverse solution as a posterior probability distribution over the unknown parameters. Though conceptually straightforward, characterizing the posterior, e.g., sample generation, marginalization, computation of moments, etc., is often computationally challenging especially when the dimensionality of the unknown parameters is large. 
The most commonly used method for posterior simulation is Markov Chain Monte Carlo (MCMC) \cite{robert1999monte}. MCMC is an exact inference method and easy to implement. However, MCMC suffers from many limitations. An efficient MCMC algorithm depends on the design of effective proposal distributions, which becomes difficult when the target distribution contains strong correlations, particularly in high-dimensional cases.  Moreover, MCMC often requires a large number of iterations, where the forward model needs to be solved at each iteration. If the model is computationally intensive, e.g., a PDE with high-dimensional spatially-varying parameter, MCMC becomes prohibitively expensive.  
While considerable efforts have been
devoted to reducing the computational cost, e.g., \cite{lieberman2010parameter,li2014adaptive,cui2015data,jiang2017multiscale,liao2019adaptive,wang2018adaptive}, many challenges still remain in inverse problems.
Furthermore, the iteration process of MCMC is not associated with a clear convergence criterion to imply when the process has adequately captured the posterior.

As an alternative strategy to MCMC sampling, variational inference (VI) is widely used to approximate posterior distributions in Bayesian inference. Compared to MCMC, VI tends to be faster and easier to scale to large data. The idea of VI is to seek the best approximation of the posterior distribution within a family of 
parameterized density models. In \cite{goh2022solving, xia2023vi, tewari-deeppriors-2022}, coupling deep generative priors with VI to solve Bayesian inverse problems is studied.  
However, for high-dimensional distributions, it is still very challenging to
obtain an accurate posterior approximation due to the curse of dimensionality. For instance, the commonly used mean-field approach \cite{blei2017variational} assumes mutual independence between dimensions to achieve a tractable density model, which in general results in underestimated second-order moments.
To remedy the issue, more capable density models are needed, where the mutual-independence assumption is relaxed. One strategy to do this is to seek 
a map that pushes the prior to the posterior, where the conditional dependence structure can be exploited and encoded into the map for more efficiency \cite{spantini2018inference}.  
More specifically, the map transforms a random variable $z$, distributed according to the prior, into a random variable $y$, distributed according to the posterior. Such transformations can be viewed as transport maps between probability measures, 
whose existence is not unique. A certain structure needs to be introduced to determine a map. 
Some typical structures include the Knothe–Rosenblatt (K-R) rearrangement \cite{el2012bayesian}, neural ODE \cite{chen2018neural} and the composition of many simple maps used in flow-based deep generative models such as NICE \cite{Dinh_2014}, real NVP \cite{dinh2016density}, KRnet \cite{tang2020deep,adda_2022,wan2022vae}, to name a few.

Two challenges need to be addressed for many practical Bayesian inverse problems. 
First, prior knowledge is often available in terms of historical data or previously acquired solutions, which should be consistent with the prior distribution of Bayesian inference. Unfortunately, the true prior may be much more complex than any commonly used explicit density models. The second challenge is the curse of dimensionality, which demands a trade-off between density approximation and sample generation for high-dimensional cases. 
To deal with these two challenges, data-driven priors and dimension reduction can be incorporated.  For example, VAE-priors \cite{goh2022solving, xia2023vi, tewari-deeppriors-2022, xia2022bayesian,zhihang2023domain} and GAN priors \cite{patel2022solution} are proposed to learn the prior distribution from data, where a mapping from the low-dimensional latent space to the high-dimensional parameter space is established and MCMC or VI is subsequently implemented in terms of the latent random variable. To further increase efficiency, a surrogate model can be employed to avoid the expensive forward problem and the adjoint problem  \cite{tromp2005seismic,cao2003adjoint} that are needed for either sampling or variational inference approaches. One popular choice for surrogate modeling is the deep neural network which is able to provide a good approximation of high-dimensional parametric PDEs. For example, physics-informed neural networks (PINN) \cite{raissi2019physics} has attracted broad attention for solving PDEs, which embeds the laws of physics into the loss function. Zhu et al. \cite{zhu2018bayesian,zhu2019physics} propose a dense convolutional encoder-decoder network for PDEs with high-dimensional random inputs.

In this work, we propose a dimension-reduced KRnet map approach (DR-KRnet) for high-dimensional Bayesian inverse problems.
The main idea is to approximate the posterior distribution in terms of the latent variable that is learned from historical data, where VAE is used for dimension reduction and KRnet is used for density approximation. We first use abundant historical data to train a VAE prior, where we use the learned decoder to transfer the inference to the latent variable. We then minimize the Kullback-Leibler divergence between the posterior for the latent variable and the density model induced by KRnet. Using the decoder and the approximate posterior of the latent variable, we can compute the desired statistics efficiently because KRnet defines a transport map that provides exact samples with neglected costs. To further increase the efficiency, we also develop a convolutional encoder-decoder network as the surrogate model \cite{zhu2018bayesian,zhu2019physics}. Compared to sampling-based approaches, the main advantages of our strategy are twofold: First, we take advantages of two capable deep generative models, i.e., VAE and KRnet, to obtain an explicit density model that is sufficiently expressive for a high-dimensional posterior distribution. Second, KRnet, a normalizing flow model, is able to effectively deal with a moderately large number of dimensions and can be more robust than MCMC. 

This paper is organized as follows. In section \ref{section_problem}, we describe the formulation of the Bayesian inverse problems that will be considered in this work. In section \ref{section_method}, our dimension-reduced KRnet map approach (DR-KRnet) for high-dimensional Bayesian inverse problems is presented, where we provide a scheme for building the neural network structure of VAE priors, introduce the KRnet to construct the map between the prior and posterior and build the physics-constrained surrogate model. In section \ref{section_experiments}, with two numerical experiments we demonstrate that our DR-KRnet can infer high-dimensional parameters efficiently. The paper is concluded in section \ref{section_conclude}.
\section{Bayesian inverse problems}\label{section_problem}
To begin with, details of the forward  model considered in this paper are addressed as follows. Let $\mathcal{S}$ denote a spatial domain that is bounded, connected and with a polygonal boundary $\partial \mathcal{S}$, and $s\in \mathcal{S}$ is  a spatial variable. The physics of the problem considered is governed by a PDE over the spatial domain $\mathcal{S}$: find $u(s,y(s))$ such that
\begin{equation}\label{physical_problem}
	\begin{aligned}
		&\mathcal{L}\Big(s,u(s,y(s));y(s)\Big)=h(s),\quad \forall s\in \mathcal{S},\\
		&\pB\Big(s,u(s,y(s));y(s)\Big)=g(s),\quad \forall s\in \partial\mathcal{S},
	\end{aligned}
\end{equation}
where $\mathcal{L}$ is a partial differential operator and $\pB$ is a boundary operator, both of which can depend on the unknown spatial-varying parameter $y(s)$, $h(s)$ is the source function, and  $g(s)$ specifies boundary conditions.

\subsection{Bayesian framework}
We consider the task of inferring the parameter $y\in \mathbb{R}^n$ from observations $\mathcal{D}_{obs}\in \mathbb{R}^m$ under the assumption that there exists a forward model $\mathcal{F}$ determined by \eqref{physical_problem} that maps the unknown parameter $y$ to the observations $\mathcal{D}_{obs}$:
\begin{align}
    \mathcal{D}_{obs}=\mathcal{F}(y)+\epsilon,
\end{align}
where $\epsilon\in \mathbb{R}^m$ is the measurement noise. Let $\pi_{\epsilon}(\epsilon)$ be the distribution of $\epsilon$, and one can obtain the distribution of $\mathcal{D}_{obs}$ conditioned on $y$:
\begin{align}
    \pi(\mathcal{D}_{obs}|y)=\pi_{\epsilon}(\mathcal{D}_{obs}-\mathcal{F}(y)).
\end{align}

Since often $m\ll n$, inverse problems are in general ill-posed, i.e.,\ one may not be able to uniquely recover the parameter $y$ given the noisy observations $\mathcal{D}_{obs}$. In the Bayesian setting, the parameters to be inferred are treated as random variables. Given the observations $\mathcal{D}_{obs}$, one assigns a prior distribution $\pi(y)$ encoding the prior information on the parameter of interest, and the posterior $\pi(y|\mathcal{D}_{obs})$ can then be calculated via the Bayes' rule:
\begin{equation}\label{yposterior}
\pi(y|\mathcal{D}_{obs})=\frac{\pi(\mathcal{D}_{obs}|y)\pi(y)}{C} \,\,\,
	\propto \,\,\,\underbrace{\pi(\mathcal{D}_{obs}|y)\pi(y)}_{\hat{\pi}(y)},
\end{equation}
where $\pi(\mathcal{D}_{obs}|y)$ is the likelihood function, and the evidence or marginal likelihood $C=\int \pi(\mathcal{D}_{obs}|y)\pi(y) dy$ is a normalization constant. 

Since the map $\mathcal{F}$ from $y$ to $\mathcal{D}_{obs}$ is typically nonlinear and the evidence is often intractable, especially for high-dimensional problems, the posterior, in general, cannot be obtained in a closed form. Therefore, the Bayesian inference needs to characterize the unnormalized posterior, which is usually achieved by variational inference (VI) or sampling approaches such as MCMC. However, extra assumptions are often introduced in VI, e.g., the family of parameterized density models for approximating the posterior distribution is a diagonal covariance Gaussian distribution, and sampling approaches such as MCMC become less efficient for sufficiently large $n$ and $m$. To improve efficiency, dimension reduction can be introduced such that VI \cite{goh2022solving,xia2023vi} or MCMC \cite{xia2022bayesian,patel2022solution} can be implemented in a low-dimensional latent space, where the dimension reduction is achieved, either explicitly or implicitly, by deep generative models. In this work we replace MCMC or VI with a normalizing flow to develop a dimension-reduced KRnet map approach (DR-KRnet) that is completely based on deep generative modeling.

\subsection{Inference with a map}
The core idea of our approach is to find a map that pushes forward the prior to the posterior in the latent space. Before taking into account dimension reduction and surrogate modeling, we look at how normalizing flows approximate the posterior of $y$. 
Let $z \in \xs{R}^n$ be a random variable that has a known distribution, e.g., the standard Gaussian. 
We seek an invertible map $f: \mathbb{R}^n\to\mathbb{R}^n$
\begin{align}
	z=f(y),
\end{align}
which depends on the observations $\mathcal{D}_{obs}$, the forward model $\mathcal{F}$, and the distribution of the measurement noise $\epsilon$. Assuming the map $f$ exists, we have the posterior by the change of variables 
\begin{align}
	p_y(y)=p_{z}(f(y)) \left |\det\nabla_{y} f \right|.
\end{align}
In practice, we will learn the map $f(\cdot)$ by minimizing the Kullback-Leibler divergence between $p_y(y)$ and the posterior:
\begin{align}
	D_{KL}\left(p_y||\pi\left(y|\mathcal{D}_{obs}\right)\right)&=\int p_{y}\log \frac{p_{y}}{\pi(y|\mathcal{D}_{obs})} dy \nonumber\\
	&=\int p_{y}\log \frac{p_{y}}{\hat{\pi}(y)} dy+\log C \nonumber\\
	&=\int p_{z}\log \frac{p_{y}(f^{-1}(z))}{\hat{\pi}(f^{-1}(z))} dz +\log C\nonumber\\
	&\approx\frac{1}{I}\sum_{i=1}^I\log p_{y}\left(f^{-1}\left(z^{(i)}\right)\right)-\frac{1}{I}\sum_{i=1}^I\log \hat{\pi}\left(f^{-1}\left(z^{(i)}\right)\right)+\log C,\quad z^{(i)} \sim p_{z}.\label{highkl}
\end{align}
It is noted that normalizing flows provide an explicit density model and an efficient way to generate exact samples of $y$ through the invertible map $y=f^{-1}(z)$. Normalizing flows can be much more expressive than classical density models, e.g., the Gaussian model subject to a diagonal covariance matrix used in the mean-field approach. Yet the construction, representation, and evaluation of these generative models grow challenging in high-dimensional cases.
Moreover, the prior $\pi(y)$ is often provided through historical data $\{y^{(i)}\}_{i=1}^N$, which may be significantly different from the commonly used prior such as the Gaussian distribution and needs to be modeled explicitly. Furthermore, 
each sample $z$ requires an evaluation of the computationally expensive forward function $\mathcal{F}$. 
To address these problems, we use a dimension-reduced VAE prior to model $\pi(y)$ through  historical data and then, in the low-dimensional latent space, apply KRnet to seek an invertible map $f$ with respect to a surrogate model for the forward problem.
\section{Dimension-reduced KRnet maps}\label{section_method}
In this section, we present a dimension-reduced KRnet map approach (DR-KRnet) in detail, which consists of three parts (choices of prior, likelihood computation, and posterior approximation).
First, the VAE prior is introduced to capture the features of $\{y^{(i)}\}_{i=1}^N$. Next, the KRnet map is adopted for pushing forward the prior to the posterior in the low-dimensional latent space. In addition, physics-constrained surrogate modeling is used to compute the likelihood function efficiently.
\subsection{VAE priors for dimension reduction}\label{vae_gan_section}
As a dimension reduction method, variational autoencoder (VAE) builds the relationship between the latent space and the original high-dimensional parameter. We briefly recall the VAE. Assume that there exists a latent random variable $x\in \mathbb{R}^d$ ($d\ll n$) with a marginal distribution $p_{x,\theta}$, where $\theta$ includes the model parameters. The joint distribution $p_{x,y,\theta}$ of $x$ and $y$ is then described by the conditional distribution $p_{y|x,\theta}$, i.e., $p_{x,y,\theta}=p_{y|x,\theta}p_{x,\theta}$.
 According to Bayes' rule,
\begin{align}
    p_{y,\theta}=\frac{p_{x,y,\theta}}{p_{x|y,\theta}}=\frac{p_{y|x,\theta}p_{x,\theta}}{p_{x|y,\theta}}.
\end{align}
The posterior distribution $p_{x|y,\theta}$ is in general
intractable, and then an approximation model $q_{x|y,\phi}$ is needed, where $\phi$ includes the model parameters. The optimal parameters $\theta$ and $\phi$ are determined by minimizing the KL divergence
\begin{align}
    D_{KL}(q_{x|y,\phi}||p_{x|y,\theta})=D_{KL}(q_{x|y,\phi}||p_{x,\theta})-\mathbb{E}_{q_{x|y,\phi}}[\log p_{y|x,\theta} ]+\log p_{y,\theta}\geq 0.
\end{align}
The minimization of $D_{KL}(q_{x|y,\phi}||p_{x|y,\theta})$ is equivalent to the maximization of
the variational lower bound of $\log p_{y,\theta}$, which is defined as 
\begin{align}\label{eq_vae_loss}
    \mathcal{L}_{\theta,\phi}(y)=\mathbb{E}_{q_{x|y,\phi}}[\log p_{y|x,\theta} ]-D_{KL}(q_{x|y,\phi}||p_{x,\theta}).
\end{align}
In the canonical VAE, we specify the PDF models respectively for $p_{y|x,\theta},\,q_{x|y,\phi}$ and $p_{x,\theta}$ as follows:
\begin{equation}
	\begin{aligned}
		p_{y|x,\theta}&=\mathcal{N}\left(\mu_{de,\theta}\left(x\right), \text{diag}\left(\sigma_{de,\theta}^{\odot 2}\left(x\right)\right)\right),\\
		q_{x|y,\phi}&=\mathcal{N}\left(\mu_{en,\phi}\left(y\right), \text{diag}\left(\sigma_{en,\phi}^{\odot 2}\left(y\right)\right)\right), \label{caen}\\
		p_{x,\theta}&=\mathcal{N}(0,\mathbf{I}),
	\end{aligned}
\end{equation}
where ${*}^{\odot2}$ means the component-wise square operation. The tuples $(\mu_{en,\theta}(y), \sigma_{en,\theta}(y))$ and $\left(\mu_{de,\theta}(x), \sigma_{de,\theta}(x)\right)$ are modeled via neural networks, i.e.,
\begin{align}
&\left(\mu_{en,\theta}(y), \sigma_{en,\theta}(y)\right)=\text{NN}_{en}(y;\theta),\label{vae1}\\
&x=\mu_{en,\phi}(y)+\sigma_{en,\phi}(y)\odot \varepsilon,\quad \varepsilon \sim \mathcal{N}(0,\mathbf{I}),\label{vae2}\\
&\left(\mu_{de,\theta}(x), \sigma_{de,\theta}(x)\right)=\text{NN}_{de}(x;\phi),\label{vae3}\\
&\hat{y}=\mu_{de,\theta}(x),
\end{align}
where $\text{NN}_{de}$ and $\text{NN}_{en}$ characterize the encoder and decoder neural networks to describe the relation between a data sample $y\in\mathbb{R}^n$ and a latent representation $x\in\mathbb{R}^d$, and $\hat{y}$ is the reconstruction of  $y$. 

Given a prior dataset $Y:=\{y^{(i)}\}_{i=1}^N$, the expectation of the variational lower bound  \eqref{eq_vae_loss} can be approximated via 
the Monte Carlo estimation
\begin{align}
    \mathbb{E}_{p_{y,\theta}}\left[\mathcal{L}_{\theta,\phi}(y)\right]&\approx \frac{1}{N}\sum_{i=1}^N\mathcal{L}_{\theta,\phi}\left(y^{(i)}\right)\nonumber\\
    &\approx\underbrace{\frac{1}{N}\sum_{i=1}^N\left[\log p_{y^{(i)}|x^{(i)},\theta}-(\log q_{x^{(i)}|y^{(i)},\phi}-\log p_{x^{(i)},\theta})\right]}_{\hat{\mathcal{L}}_{\theta,\phi}(Y)}, \label{vae_loss_discri}
\end{align}
where $x^{(i)}$ can be generated by substituting $y^{(i)}$ into  \eqref{vae2}.
We pre-train the VAE priors by Algorithm \ref{alg_vae_gan}, of which the output  is the pre-trained decoder $p_{y|x,\theta^*}$. Here $\theta^*$ consists of the optimal parameters of the decoder.
The inference over $y$ in  \eqref{yposterior} is replaced by infering the latent variable $x$ from the observations, formulated as
\begin{align}
	\pi(x|\mathcal{D}_{obs})\propto &{\pi(\mathcal{D}_{obs}|x)\pi(x)}, \nonumber\\
	&=\left(\int \pi(\mathcal{D}_{obs}|y,x) \pi(y|x) dy\right) \pi(x)\nonumber\\
	&= \underbrace{\left(\int \pi(\mathcal{D}_{obs}|y,x) p_{y|x,\theta^*} dy\right) p_{x,\theta^*}}_{\hat{\pi}(x)}, \label{xposterior}
\end{align}
where $\pi(\mathcal{D}_{obs}|y,x) $ is the likelihood function, $p_{y|x,\theta^*}$ is the pre-trained decoder, and $p_{x,\theta^*}$ is a simple prior distribution of VAE, e.g., the standard Gaussian. 
\begin{algorithm}[H]
	\caption{Training the VAE priors}
	\label{alg_vae_gan}
	\begin{algorithmic}[1]
		\Require The prior dataset $Y:=\{y^{(i)}\}_{i=1}^N$, maximum epoch number $E$, batch size $n_{batch}$, learning rate $\eta$.
           \State Divide $Y$ into $N_b$ mini-batches $\{Y_j\}_{j=1}^{N_b}$ where $N_b=\frac{N}{n_{batch}}$.
           \State Initialize $\theta$ and $\phi$ for the encoder and decoder networks.		
           \For {$i = 1:E$}
		\For {$j=1:N_b$}
            \State Construct the noise set $S_j=\{\varepsilon^k\sim \mathcal{N}(0,\mathbf{I}),k=1,2,\dots,n_{batch}\}$.
            \State Apply $Y_j$ and $S_j$ to compute  \eqref{vae1}--\eqref{vae3}.
		\State Compute  $-\hat{\mathcal{L}}_{\theta,\phi}(Y_j)$ in  \eqref{vae_loss_discri} and its gradients $-\nabla_{\theta}\hat{\mathcal{L}}_{\theta,\phi}(Y_j),\,-\nabla_{\phi}\hat{\mathcal{L}}_{\theta,\phi}(Y_j)$.
		\State Update the parameters $(\theta,\phi)$ using gradient-based optimization algorithm (e.g., Adam optimizer \cite{kingma2014adam} with learning rate $\eta$).
		\EndFor
		\EndFor
         \State Let $\theta^*=\theta$, where $\theta$ includes the parameters of the decoder networks at the last epoch.
		\Ensure The probabilistic decoder $p_{y|x,\theta^*}$.
	\end{algorithmic}
\end{algorithm}

\subsection{KRnet map}
In  \eqref{xposterior}, let $\pi(x|\mathcal{D}_{obs})=C^{-1}\hat{\pi}(x)$, $x\in \mathbb{R}^d$, $d\ll n$. In the low-dimensional latent space of the pre-trained VAE prior, we intend to approximate the posterior $\pi(x|\mathcal{D}_{obs})$ by constructing a map that pushes forward the prior to the posterior. In other words, we seek a transport map $\mathcal{T}$: $z \mapsto x$ such that $\mathcal{T}_{\#} \mu_z = \mu_x$, where $d\mu_z=p_{z,\theta^*}dz$ and $d\mu_x=\pi(x|\mathcal{D}_{obs})dx$ are the probability measures of $z$ and $x$ respectively, and $\mathcal{T}_{\#} \mu_{z}$ is the push-forward of $\mu_{z}$ satisfying $\mu_{x}(B) = \mu_{z}(\mathcal{T}^{-1}(B))$ for every Borel set $B$. The Knothe-Rosenblatt rearrangement tells us that the transport map $\mathcal{T}$ may have a lower-triangular structure 
\begin{equation}
    {z} = \mathcal{T}^{-1}({x}) = \left[ 
    \begin{array}{l}
    \mathcal{T}_1(x_1) \\
    \mathcal{T}_2(x_1, x_2) \\
    \vdots \\
    \mathcal{T}_{d}(x_1, \ldots, x_d)
    \end{array}
    \right].
\end{equation} 
This mapping can be regarded as a limit of sequence of optimal transport maps when the quadratic cost degenerates \cite{carlier2010knothe}. 



The basic idea of KRnet is to define the structure of a normalizing flow $f({x})$ in terms of the Knothe-Rosenblatt rearrangement which results in KRnet as a generalization of real NVP \cite{dinh2016density}. Let ${x} = \left[{x}^{(1)}, \ldots, {x}^{(K)}\right]^\mathsf{T}$ be a partition of ${x}$, where ${x}^{(i)} =\left[x_{1}^{(i)}, \ldots, x_{m}^{(i)}\right]^\mathsf{T}$ with $1 \leq K \leq d, 1 \leq m \leq d$, and $\sum_{i=1}^K \mathrm{dim}\left({x}^{(i)}\right) = d$. Our KRnet takes an overall form
\begin{equation} \label{eqn_KR}
  {z} = f({x}) = \left[ 
    \begin{array}{l}
    f_1\left({x}^{(1)}\right) \\
    f_2\left({x}^{(1)}, {x}^{(2)}\right) \\
    \vdots \\
    f_{K}\left({x}^{(1)}, \ldots, {x}^{(K)}\right)
    \end{array}
    \right].
\end{equation}
Each $f_{i}$, $i=2,\ldots,K$, is constructed with real NVP by stacking a sequence of simple bijections. KRnet provides a more expressive density model than real NVP for the same model size. More details about KRnet can be found in \cite{tang2020deep,adda_2022}.



Let $q_{x,\alpha}$ be the PDF model induced by a KRnet with model parameters $\alpha$, and then  \eqref{eqn_KR} is reformulated into
\begin{equation}\label{eqn_KR_alpha}
    {z} = f_{\alpha}({x}).
\end{equation}
To approximate $\pi(x|\mathcal{D}_{obs})$ in  \eqref{xposterior}, we minimize the KL divergence between $q_{x,\alpha}$ and $\pi(x|\mathcal{D}_{obs})$
\begin{align*}
	D_{KL}\left(q_{x,\alpha}||\pi\left(x|\mathcal{D}_{obs}\right)\right)=\int q_{x,\alpha}\log \frac{q_{x,\alpha}}{\pi(x|\mathcal{D}_{obs})} dx
	=\int q_{x,\alpha}\log \frac{q_{x,\alpha}}{\hat{\pi}(x)} dx+\log C,
\end{align*}
which is equivalent to minimize the following functional
\begin{align}
	\int q_{x,\alpha}\log \frac{q_{x,\alpha}}{\hat{\pi}(x)} dx 
	&=\int p_{z,\theta^*}\log \frac{q_{x,\alpha}\left(f_{\alpha}^{-1}(z)\right)}{\hat{\pi}\left(f_{\alpha}^{-1}(z)\right)} dz \nonumber\\
	&\approx\frac{1}{I}\sum_{i=1}^I\log q_{x,\alpha}\left(f_{\alpha}^{-1}\left(z^{(i)}\right)\right)-\frac{1}{I}\sum_{i=1}^I\log \hat{\pi}\left(f_{\alpha}^{-1}\left(z^{(i)}\right)\right),\quad z^{(i)} \sim p_{z,\theta^*}. \label{firstloss}
\end{align}
Let $x^{(i)}=f_{\alpha}^{-1}\left(z^{(i)}\right)$. The second term of the right hand of  \eqref{firstloss} is obtained as
\begin{align*}
	-\frac{1}{I}\sum_{i=1}^I\log \hat{\pi}\left(x^{(i)}\right)
	&=-\frac{1}{I}\sum_{i=1}^I\log \left(\int \pi\left(\mathcal{D}_{obs}|y,x^{(i)}\right) p_{y|x^{(i)},\theta^*} dy\right) -\frac{1}{I}\sum_{i=1}^I\log p_{x^{(i)},\theta^*}\\
	&\leq -\frac{1}{I}\sum_{i=1}^I\int p_{y|x^{(i)},\theta^*} \log \pi\left(\mathcal{D}_{obs}|y,x^{(i)}\right) dy-\frac{1}{I}\sum_{i=1}^I\log p_{x^{(i)},\theta^*}\\
	&\approx -\frac{1}{I}\frac{1}{J}\sum_{i=1}^I\sum_{j=1}^J \log \pi\left(\mathcal{D}_{obs}|y^{(i,j)},x^{(i)}\right)-\frac{1}{I}\sum_{i=1}^I\log p_{x^{(i)},\theta^*},\quad y^{(i,j)}\sim p_{y|x^{(i)},\theta^*}, 
\end{align*}
where the Jensen's inequality is applied and $\pi\left(\mathcal{D}_{obs}|y^{(i,j)},x^{(i)}\right)$ is the likelihood function. Since the first term on the right-hand side corresponds to the expectation of $\log\pi(\mathcal{D}_{obs}|y,x)$ with respect to the joint distribution given by $p_{y|x,\theta^*}p_{x,\theta^*}$, we may simply let $J=1$.  
We then reach our objective function for minimization
\begin{align}
	\mathcal{L}_{KRnet}&=\frac{1}{I}\sum_{i=1}^I\log q_{x^{(i)},\alpha}-\frac{1}{I}\sum_{i=1}^I \log \pi\left(\mathcal{D}_{obs}|y^{(i)},x^{(i)}\right)-\frac{1}{I}\sum_{i=1}^I\log p_{x^{(i)},\theta^*}, \label{krnet_loss}
\end{align}
where $x^{(i)}=f_{\alpha}^{-1}\left(z^{(i)}\right),\, z^{(i)} \sim p_{z,\theta^*}$ and $y^{(i)}\sim p_{y|x^{(i)},\theta^*} $.

Once KRnet has been trained by minimizing $\mathcal{L}_{KRnet}$, we can estimate the moments of the posterior  $\pi(y|\mathcal{D}_{obs})$ through the pre-trained decoder,
\begin{align}
	\mathbb{E}[y]&=\int y \left(\int p_{y|x,\theta^*}q_{x,\alpha^*}dx \right) dy
	\approx \frac{1}{N_s}\sum_{i=1}^{N_s}\int y p_{y|x^{(i)},\theta^*} dy\nonumber\\
	&\approx \frac{1}{N_s}\sum_{i=1}^{N_s} \mu_{de,\theta^*}\left(x^{(i)}\right),\quad x^{(i)}=f_{\alpha^*}^{-1}\left(z^{(i)}\right),\, z^{(i)} \sim p_{z,\theta^*}, \label{mean_compute}
\end{align}
\begin{align}
	\mathbb{V}[y]&=\mathbb{E}\left[\left(y-\mathbb{E}[y]\right)\left(y-\mathbb{E}[y]\right)^\mathsf{T}\right]\nonumber\\
	&=\int (y-\mathbb{E}[y])(y-\mathbb{E}[y])^\mathsf{T}\left(\int p_{y|x,\theta^*}q_{x,\alpha}dx \right) dy\nonumber\\
	&\approx \frac{1}{N_s}\sum_{i=1}^{N_s}\int (y-\mathbb{E}[y])(y-\mathbb{E}[y])^\mathsf{T} p_{y|x^{(i)},\theta^*} dy\nonumber\\
	&\approx \frac{1}{N_s}\sum_{i=1}^{N_s} \text{diag}\left(\sigma_{de,\theta^*}^{\odot 2}\left(x^{(i)}\right)\right),\quad x^{(i)}=f_{\alpha^*}^{-1}\left(z^{(i)}\right),\, z^{(i)} \sim p_{z,\theta^*},\label{variance_compute}
\end{align}
where $ p_{y|x^{(i)},\theta^*}=\mathcal{N}\left(\mu_{de,\theta^*}\left(x^{(i)}\right), \text{diag}\left(\sigma_{de,\theta^*}^{\odot 2}\left(x^{(i)}\right)\right)\right)$ is the pre-trained decoder given in section \ref{vae_gan_section}, $\alpha^*$ represents the optimal parameters of KRnet, and $N_s$ is the number of posterior samples.
\subsection{Physics-constrained surrogate modeling}
Asides from the pre-trained decoder, we need to pre-train a surrogate model for the forward problem such that we may efficiently minimize $\mathcal{L}_{KRnet}$ given in  \eqref{krnet_loss} by stochastic gradient-based optimization \cite{bottou2018optimization}.
Assume that the governing 
equations are defined on a two-dimensional regular $H\times W$ grid, where $H$ and $W$ denote the number of grid points along the two axes of the spatial domain. We transform the surrogate modeling problem into an image-to-image regression problem through a mapping 
\begin{align}\label{surroagte_map}
	\hat{\mathcal{F}}:y\in \mathbb{R}^{d_y\times H\times W }\to u\in \mathbb{R}^{d_u\times H\times W }.
\end{align}
Here $d_y$ and $d_u$ are treated as the number of channels in the input and output images, similar to the RGB channels in natural images. More specifically, the surrogate model $u=\hat{\mathcal{F}}_{\Theta}(y)$ with model parameters $\Theta$ is composed of convolutional encoder and decoder networks, i.e.,\ $u=\text{decoder}\circ\text{encoder}(y)$. The surrogate model is trained without labeled data, in other words, the PDE will not be simulated for some chosen $y$. Similar to PINN, it is trained \cite{raissi2019physics} by enforcing the constraints given by  \eqref{physical_problem}, i.e., we minimize the following objective function:
\begin{align}\label{surrogate_loss}
	\mathcal{J}\left(\Theta;\{y^{(i)}\}_{i=1}^N\right)=\frac{1}{N}\sum_{i=1}^N\left[ {\left\Arrowvert \mathcal{R}\left(\hat{\mathcal{F}}_{\Theta}\left(y^{(i)}\right),y^{(i)}\right)\right\Arrowvert}_2^2+\beta{\left\Arrowvert\mathcal{B}\left(\hat{\mathcal{F}}_{\Theta}\left(y^{(i)}\right)\right)\right\Arrowvert}_2^2\right],
\end{align}
where 
$\mathcal{R}\left(\hat{\mathcal{F}}_{\Theta}\left(y^{(i)}\right),y^{(i)}\right)=\mathcal{L}\left(\hat{\mathcal{F}}_{\Theta}\left(y^{(i)}\right);y^{(i)}\right)-h$ and $\mathcal{B}\left(\hat{\mathcal{F}}_{\Theta}\left(y^{(i)}\right)\right)=\pB\left(\hat{\mathcal{F}}_{\Theta}\left(y^{(i)}\right);y^{(i)}\right)-g$ measure how well $\hat{\mathcal{F}}_{\Theta}\left(y^{(i)}\right)$ satisfies the partial differential equations and the boundary conditions, respectively, and $\beta>0$ is a penalty parameter. 
Both $\mathcal{R}\left(\hat{\mathcal{F}}_{\Theta}\left(y^{(i)}\right),y^{(i)}\right)$ and $\mathcal{B}\left(\hat{\mathcal{F}}_{\Theta}\left(y^{(i)}\right)\right)$ may involve
integration and differentiation with respect to the spatial coordinates, which are approximated with highly efficient discrete operations, e.g.,\ Sobel filters \cite{ma2004invitation,zhu2019physics}. The surrogate trained with the loss function  \eqref{surrogate_loss} is called
physics-constrained surrogate. The 
training process is summarized in Algorithm \ref{alg_surrogate}.

\begin{figure}
	\centering
	\includegraphics[width=1.0\textwidth]{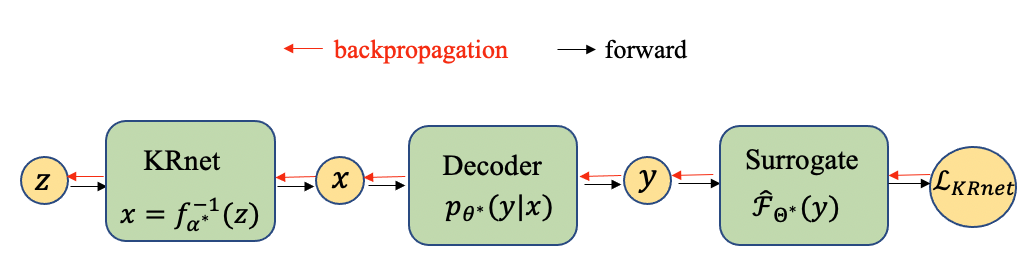}
	\caption{The full workflow of seeking the dimension-reduced KRnet map.}
	\label{krnet_flow}
\end{figure}

Once we obtain the pre-trained decoder $p_{y|x,\theta^*}$ and the pre-train surrogate model $\hat{\mathcal{F}}_{\Theta^*}(y)$, we can find the transport map from the prior to the posterior in the low-dimensional latent space, 
which is implemented in Algorithm \ref{alg_krnet}. The whole process of seeking the dimension-reduced KRnet map (DR-KRnet) is shown in Figure \ref{krnet_flow}.

\begin{algorithm}[H]
	\caption{Training the physics-constrained surrogate model}
	\label{alg_surrogate}
	\begin{algorithmic}[1]
		\Require The prior dataset $Y:=\{y^{(i)}\}_{i=1}^N$, maximum epoch number $E$, batch size $n_{batch}$, and learning rate $\eta$.
            \State Divide $Y$ into $N_b$ mini-batches $\{Y_j\}_{j=1}^{N_b}$ where $N_b=\frac{N}{n_{batch}}$.
            \State Initialize $\Theta$ for the surrogate networks.
		\For {$i = 1:E$}
		\For {$j=1:N_b$}
            \State Compute the objective function $\mathcal{J}(\Theta; Y_j)$ in  \eqref{surrogate_loss} and its gradient $\nabla_{\Theta}\mathcal{J}(\Theta; Y_j)$.
		\State Update the parameters $\Theta$ using gradient-based optimization algorithm (e.g., Adam optimizer \cite{kingma2014adam} with learning rate $\eta$).
		\EndFor
		\EndFor
  \State Let $\Theta^*=\Theta$, where $\Theta$ includes the parameters of the surrogate networks at the last epoch.
		\Ensure The surrogate model $u=\hat{\mathcal{F}}_{\Theta^*}(y)$ with optimal parameters $\Theta^*$.
	\end{algorithmic}
\end{algorithm}
\begin{algorithm}[H]
	\caption{Dimension-reduced KRnet maps (DR-KRnet)}
	\label{alg_krnet}
	\begin{algorithmic}[1]
		\Require Pre-trained decoder $p_{y|x,\theta^*}=\mathcal{N}\left(\mu_{de,\theta^*}\left(x\right), \text{diag}\left(\sigma_{de,\theta^*}^{\odot 2}\left(x\right)\right)\right)$, pre-trained surrogate model $\hat{\mathcal{F}}_{\Theta^*}$, sample size from $\mathcal{N}(0,\mathbf{I})$ $I$, sample size for posterior distribution $N_s$, batch size $n_{batch}$, maximum epoch number $E$, learning rate $\eta$.
  \State Generate the training dataset $Z:=\{z^{(i)}\}_{i=1}^I$ where $z^{(i)}\sim \mathcal{N}(0,\mathbf{I})$.
  \State Divide $Z$ into $N_b$ mini-batches $\{Z_j\}_{j=1}^{N_b}$ where $N_b=\frac{I}{n_{batch}}$.
  \State Initialize $\alpha$ of the KRnet map.
		\For {$i = 1:E$}
		\For {$j=1:N_b$}
		\State Compute $X_j=f_{\alpha}^{-1}(Z_j)$ in  \eqref{eqn_KR_alpha}.
            \State Compute the high-dimensional parameters: $Y_j=\mu_{de,\theta^*}(X_j).$
		\State Compute the surrogate model: $U_j=\hat{\mathcal{F}}_{\Theta^*}(Y_j).$
		\State Compute the loss function $\mathcal{L}_{KRnet}$ in  \eqref{krnet_loss} and its gradient $\nabla_{\alpha}\mathcal{L}_{KRnet}$.
		\State Update the parameters $\alpha$ using gradient-based optimization algorithm (e.g., Adam optimizer \cite{kingma2014adam} with learning rate $\eta$).
		\EndFor
		\EndFor
  \State Let $\alpha^*=\alpha$, where $\alpha$ includes the parameters of the KRnet map at the last epoch.
		\State Sample $\{z^{(i)}\}_{i=1}^{N_s} $ where $z^{(i)}\sim \mathcal{N}(0,\mathbf{I})$.
		\State $x^{(i)}=f_{\alpha^*}^{-1}\left(z^{(i)}\right)$, for $i=1,2,\dots,N_s$.
		\State Compute the posterior mean $\hat{\mathbb{E}}[y]=\frac{1}{N_s}\sum_{i=1}^{N_s} \mu_{de,\theta^*}\left(x^{(i)}\right)$ in  \eqref{mean_compute}.
             \State Compute the posterior variance $\hat{\mathbb{V}}[y]=\frac{1}{N_s}\sum_{i=1}^{N_s} \text{diag}\left(\sigma_{de,\theta^*}^{\odot 2}\left(x^{(i)}\right)\right)$ in  \eqref{variance_compute}.
		\Ensure The posterior mean $\hat{\mathbb{E}}[y]$ and the posterior variance $\hat{\mathbb{V}}[y]$.
	\end{algorithmic}
\end{algorithm}
\section{Numerical experiments}\label{section_experiments}
We consider single-phase, steady-state Darcy flows. Let $\alpha(s)$ denote an unknown permeability field. The pressure field $u(s,y(s))$ is defined by the following diffusion equation
\begin{align}\label{diffusion}
    -\nabla \cdot\left(\alpha(s)\nabla u(s,\alpha(s))\right)=h(s),\quad s\in \mathcal{S},
\end{align}
where the physical domain $\mathcal{S}=(0,1)^2\in\mathbb{R}^2$ is considered. We use homogeneous Dirichlet boundary conditions on the left and right boundaries and homogeneous Neumann boundary conditions on the top
and bottom boundaries, i.e.,
\begin{align*}
    &u(s,\alpha(s))=0,\quad s\in  \{0\} \times [0, 1],\\
     &u(s,\alpha(s))=0,\quad s\in  \{1\} \times [0, 1],\\
    &\alpha(s)\nabla u(s,\alpha(s))\cdot \mathbf{n}=0,\quad s\in \{(0, 1) \times \{0\}\} \cup \{(0, 1) \times \{1\}\},
\end{align*}
where $\mathbf{n}$ is the outward-pointing normal to the Neumann boundary. The source term is specified as $h(s)=3$.
In the following numerical experiments, the computation domain $\mathcal{S}$ is discretized by a uniform 64×64 grid, i.e., $H=64,W=64$ in \eqref{surroagte_map}. The goal in this paper is to infer the log-permeability field $y(s)=\log \alpha(s)$ from noisy and incomplete observations.


We assume that the log-permeability field $y(s)$ is a Gaussian random field (GRF), i.e.,\,$y(s)\sim \mathcal{GP}\left(m(s),k(s_1,s_2)\right)$, where $m(s)$ and $k(s_1,s_2)$ are the mean and covariance functions, respectively. Let $s_1=[s_{1,1},s_{1,2}]^\mathsf{T}$ and $s_2=[s_{2,1},s_{2,2}]^\mathsf{T}$ denote two arbitrary spatial locations. The covariance function $k(s_1,s_2)$ is taken as
\begin{align}\label{covariance}
    k(s_1,s_2)=\sigma^2\exp\left(-\sqrt{\left(\frac{s_{1,1}-s_{2,1}}{l_1}\right)^2+\left(\frac{s_{1,2}-s_{2,2}}{l_2}\right)^2}\right),
\end{align}
where $\sigma^2$ is the variance, and $l_1,\,l_2$ are the length scales. 
This random field can be approximated by
a truncated Karhunen-Lo{\`e}ve expansion (KLE),
\begin{align}
    y(s)\approx m(s)+\sum_{k=1}^{d_{KL}}\sqrt{\lambda_k}y_k(s)\xi_k,
\end{align}
where $d_{KL}\in\mathbb{N}_+$, $y_k(s)$ and $\lambda_k$ are the eigenfunctions and
eigenvalues of $k(s_1,s_2)$ and $\{\xi_k\}_{k=1}^{d_{KL}}$ are i.i.d. Gaussian random variables of zero mean and unit variance. 
We set $m(s)=1$ and $\sigma^2=0.5$ in the numerical experiments. 
We set $d_{KL}$ large enough such that $95\%$ of the total variance of the exponential covariance function are captured. 

We now generate the datasets as historical data for the training of VAE priors. 
 One can assume that the data are from random fields of different length scales. 
 More specifically, we consider two different experimental setups with an increasing difficulty. The length scales are set to be $l_1 = l_2 = 0.2 + 0.01i,i = 0, 1, 2,\dots, 9$ in test problem 1 and 
 $l_1 = l_2 = 0.1 + 0.01i,i = 0, 1, 2,\dots, 9$ in test problem 2.
We generate 2000 samples for each length scale and combine them to obtain the prior datasets $\{y^{(i)}\}_{i=1}^N$ for training VAE priors, where $N=20000$. Note that the KLE method
is inappropriate for dealing with varying correlation lengths but the VAE priors in section \ref{vae_gan_section} for characterizing the prior do not have such a limitation.  The architectures of the neural networks used in this paper have been summarized in Appendix. All neural network models are trained on a single NVIDIA GeForce GTX 1080Ti GPU card.

To access the accuracy of the estimated posterior mean field, relative errors are defined as
\begin{align}\label{error_compute}
    \epsilon_{relative}:={\left\Arrowvert\mathbb{E}[y]-y_{exact}\right\Arrowvert}_2/{\Arrowvert y_{exact}\Arrowvert}_2,
\end{align}
where $y_{exact}$ is the exact log-permeability field, and $\mathbb{E}[y]$ can be approximated by computing the posterior mean through \eqref{mean_compute}.

\subsection{Test problem 1}
\begin{figure}
	\centering
	\subfloat[][Prior sample 1]{\includegraphics[width=.28
 \textwidth]{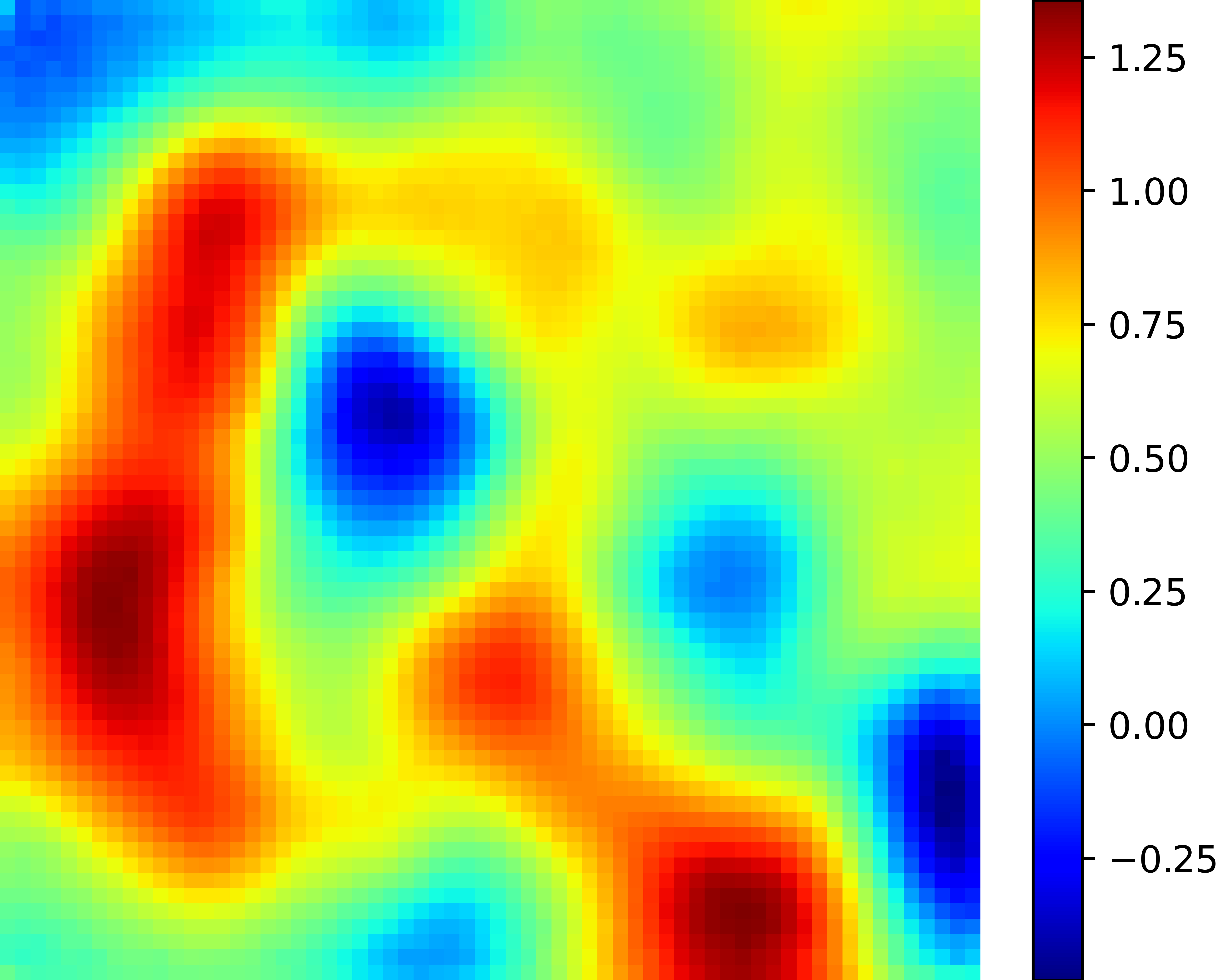}}\quad
	\subfloat[][Prior sample 2]{\includegraphics[width=.28\textwidth]{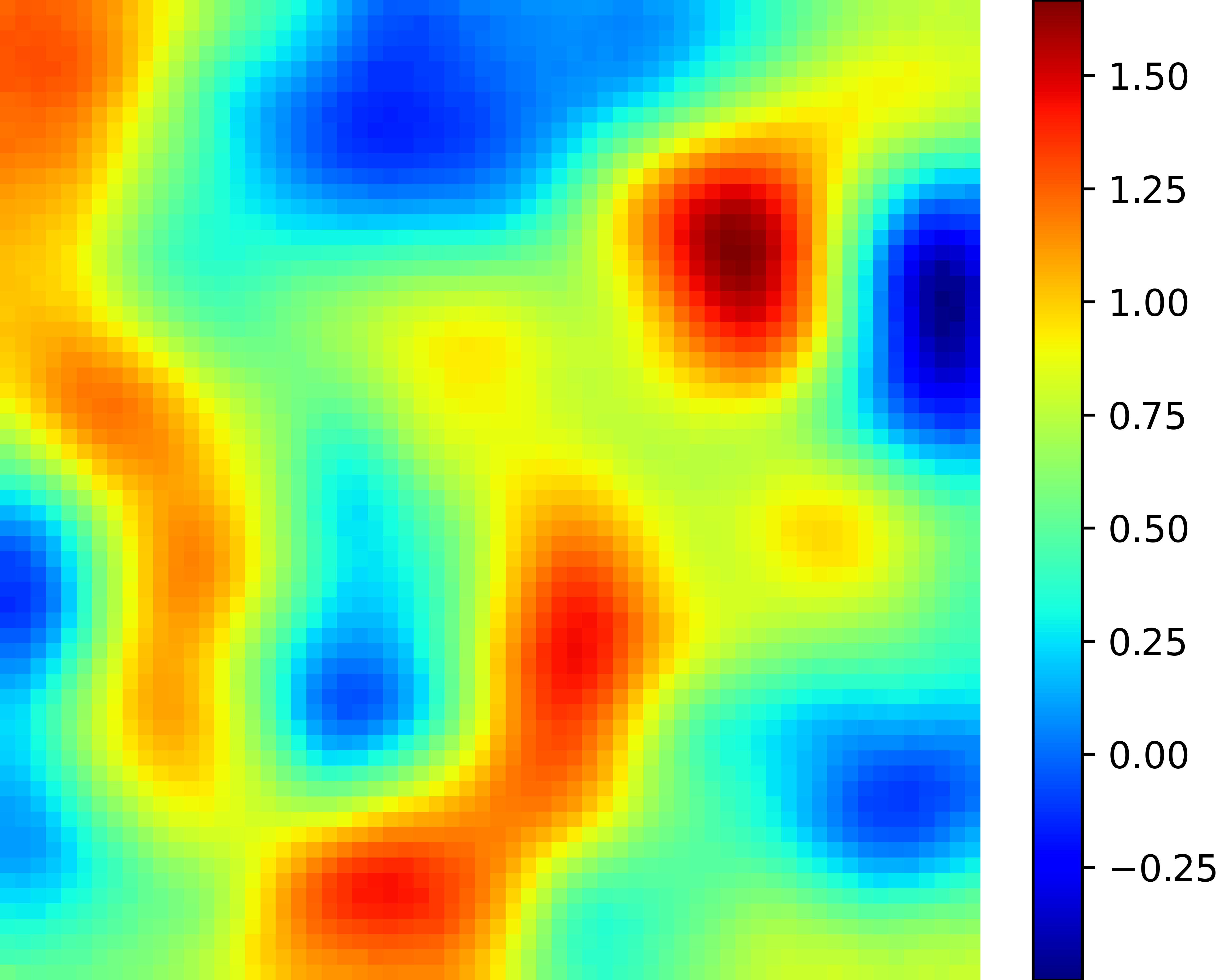}}\quad
	\subfloat[][Prior sample 3]{\includegraphics[width=.28\textwidth]{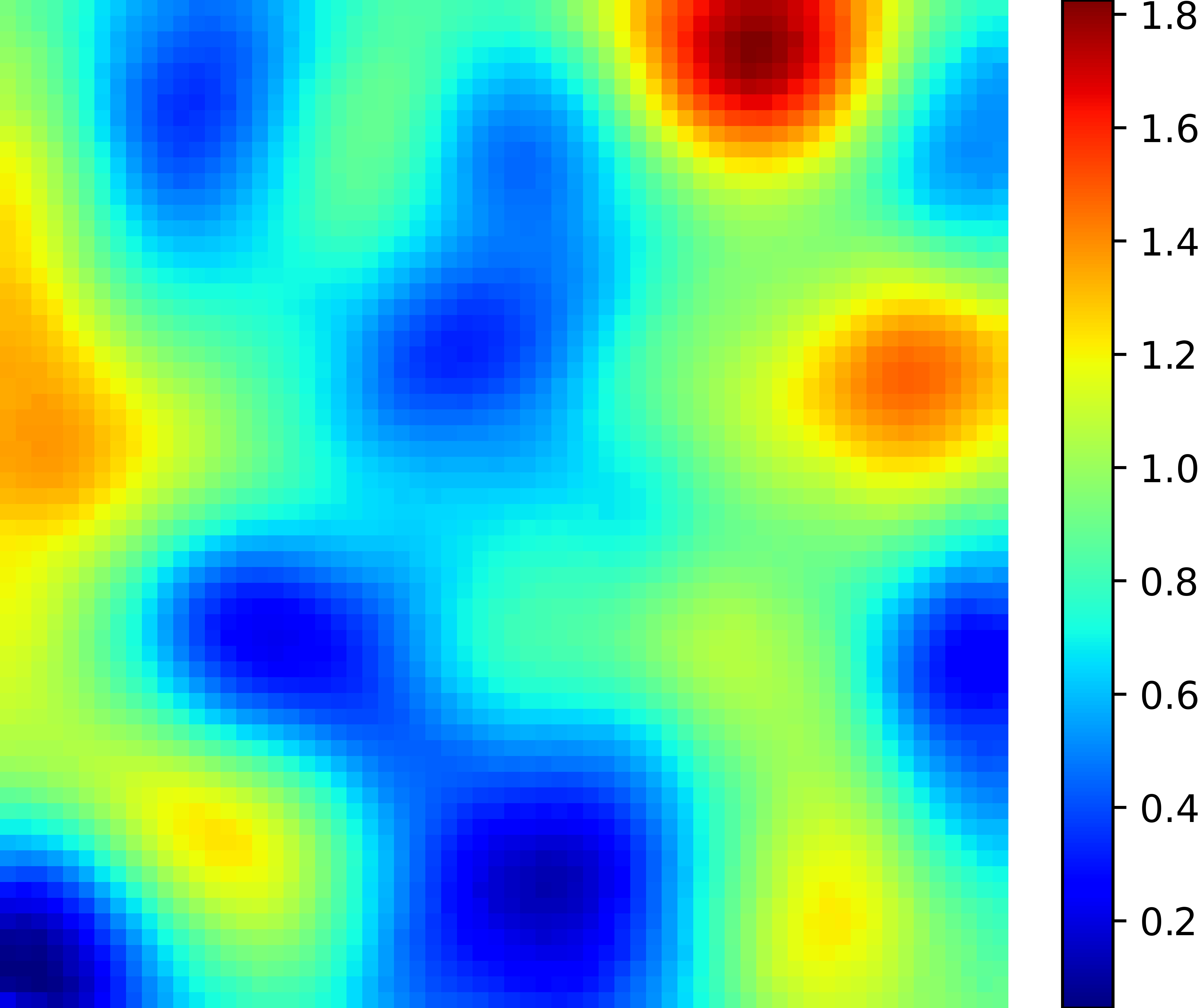}}
	\\
	\subfloat[][Prior sample 4]{\includegraphics[width=.28\textwidth]{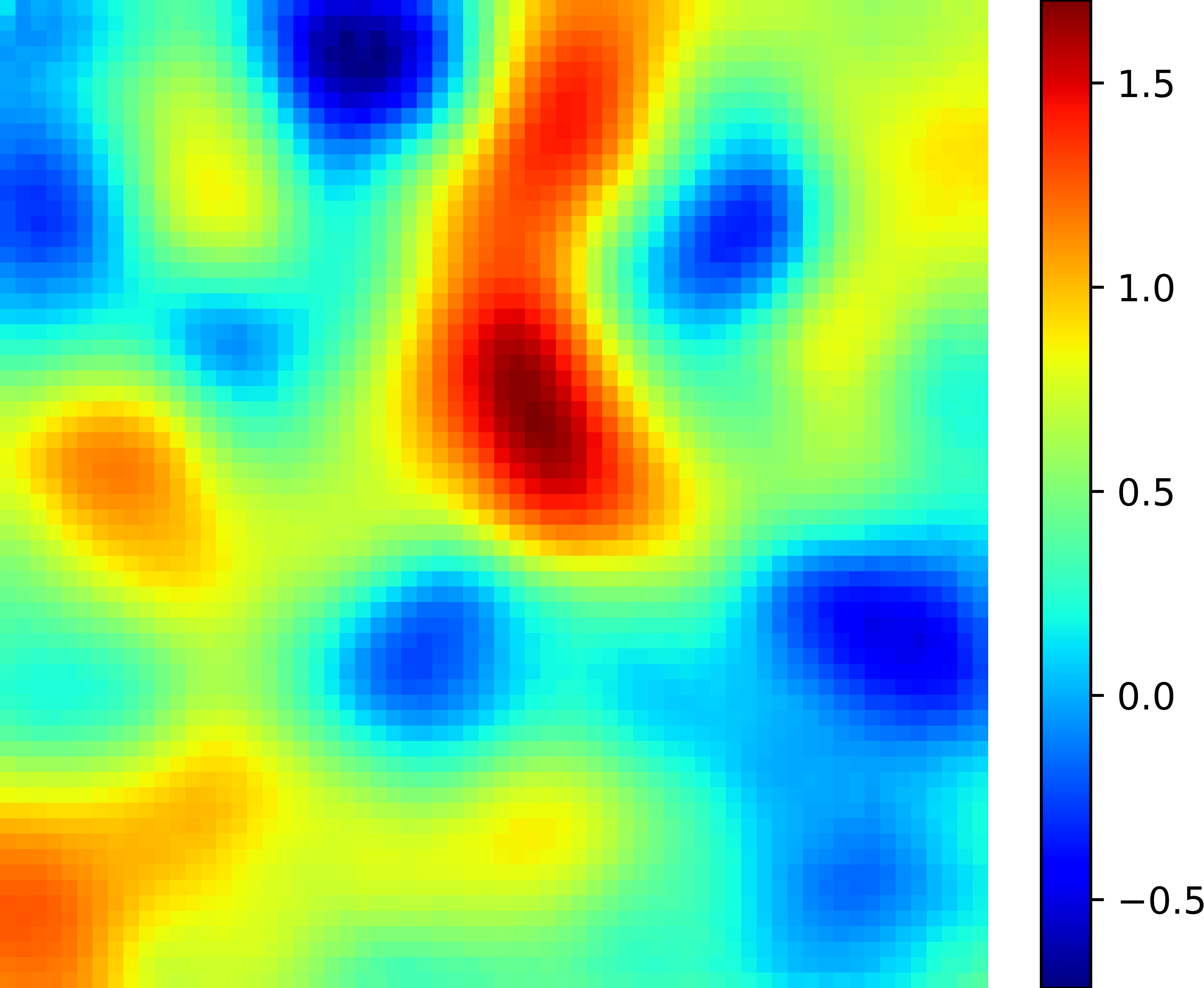}}\quad
	\subfloat[][Prior sample 5]{\includegraphics[width=.28\textwidth]{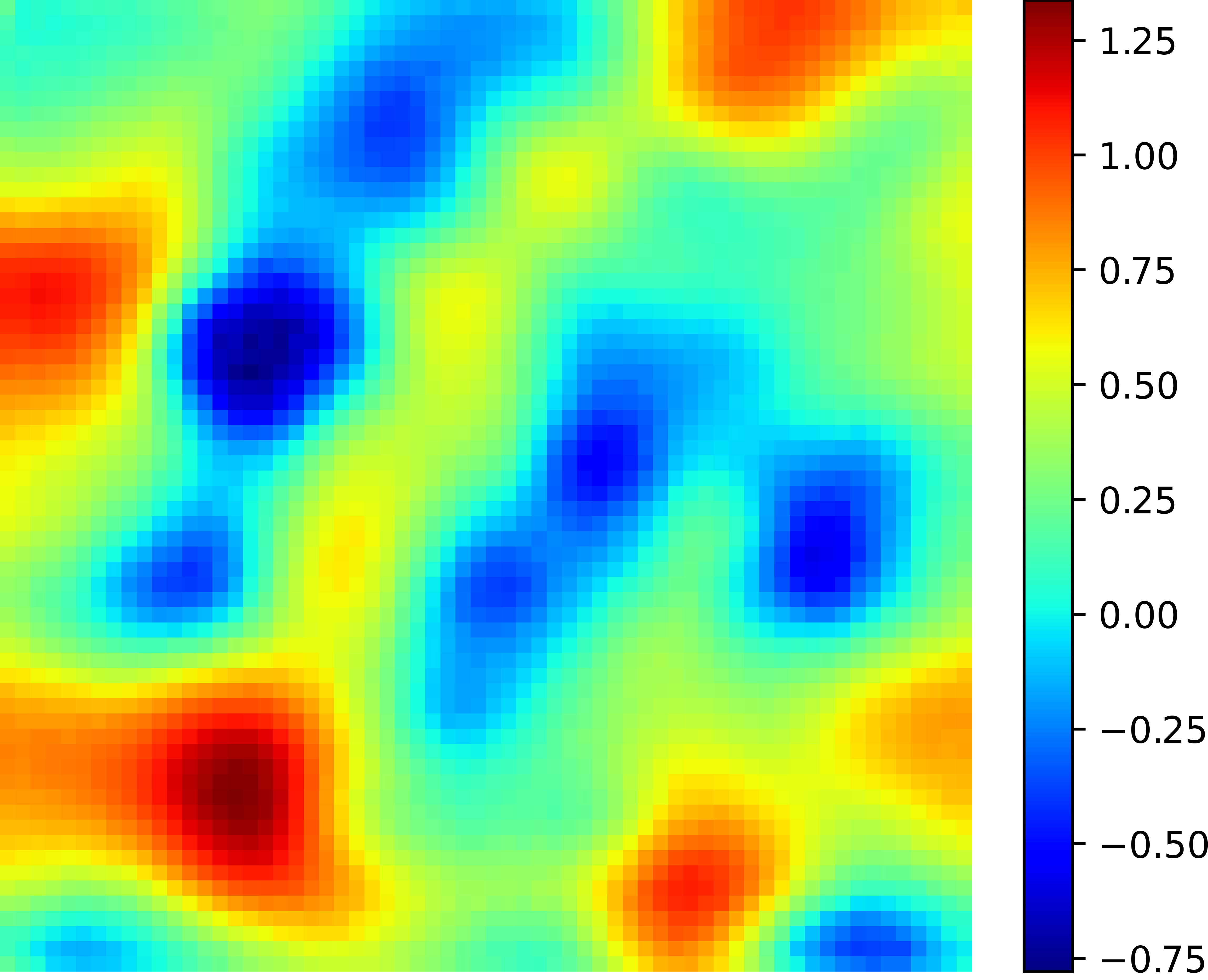}}
   \quad \subfloat[][Prior sample 6]{\includegraphics[width=.28\textwidth]{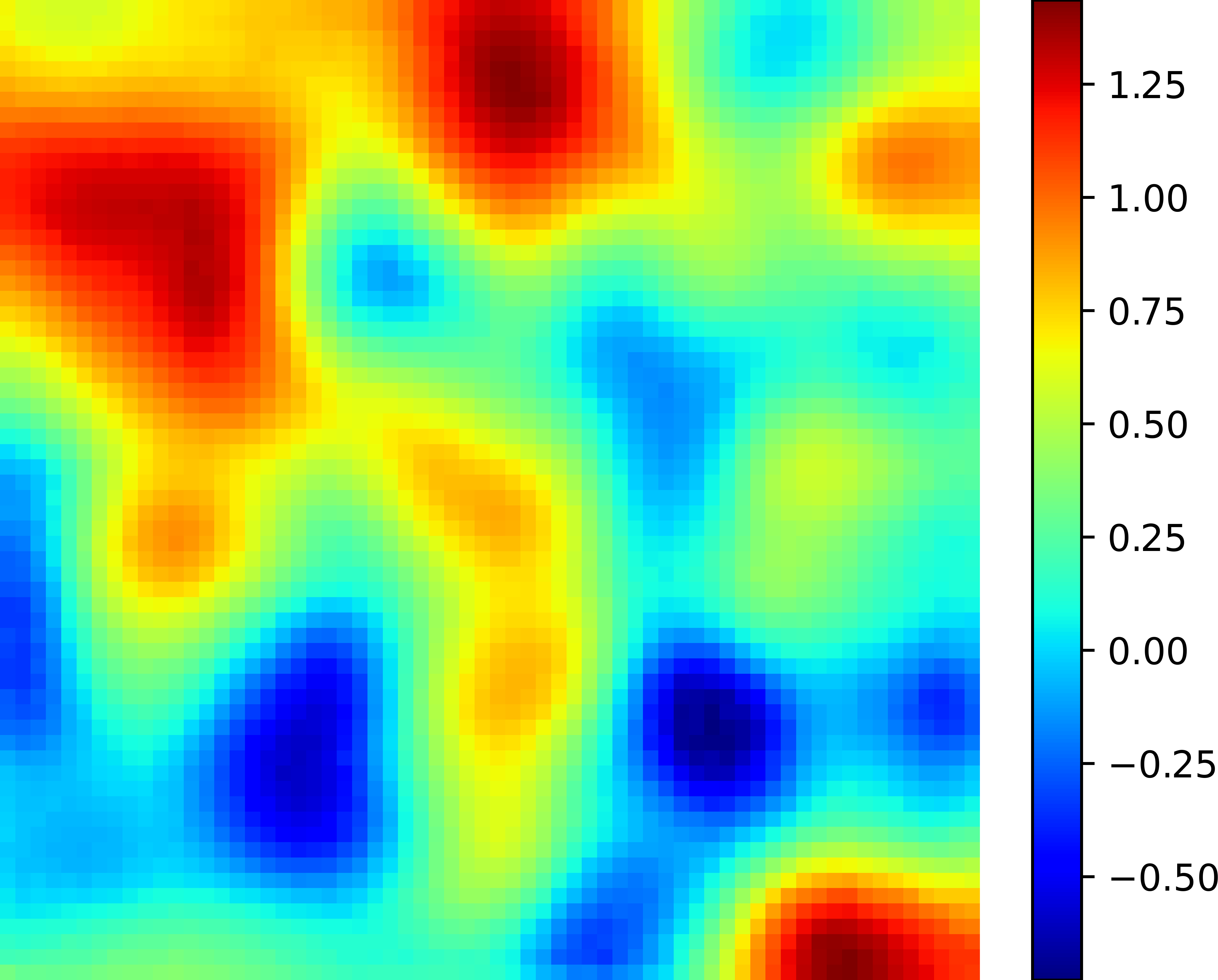}}
	\caption{64×64 resolution random samples generated from $p_{y|x,\theta^*}p_x$ for test problem 1, where $x\in \mathbb{R}^{36}$, $p_{y|x,\theta^*}$ is the decoder of the VAE prior and  $p_x=\mathcal{N} (0, \mathbf{I})$.}
    \label{samplesVAE}
\end{figure}

Given the prior dataset $\{y^{(i)}\}_{i=1}^N$ with $y^{(i)}\in\mathbb{R}^{ 64\times 64}$, the latent variable is set to $x\in \mathbb{R}^{36}$ and then we train the corresponding VAE prior. 
In Algorithm \ref{alg_vae_gan}, we assign the batch size $n_{batch} = 100$ and the maximum epoch number $E=200$, and 
employ the Adam optimizer with a learning rate $\eta=0.0001$. 
The architecture of the VAE prior is given in \ref{vae_nn}. 
To generate samples that are consistent with the prior dataset,  one can sample a latent variable $x$ from Gaussian distribution $\mathcal{N}(0,\mathbf{I})$, and then generate the samples of $y$ by the learned decoder $p_{y|x,\theta^*}$ , i.e., $y=\mu_{de,\theta^*}(x)$. Some samples generated by the VAE prior are shown in Figure \ref{samplesVAE}.

\begin{figure}
	\centering
	\subfloat[][The exact log-permeability]{\includegraphics[width=.28
 \textwidth]{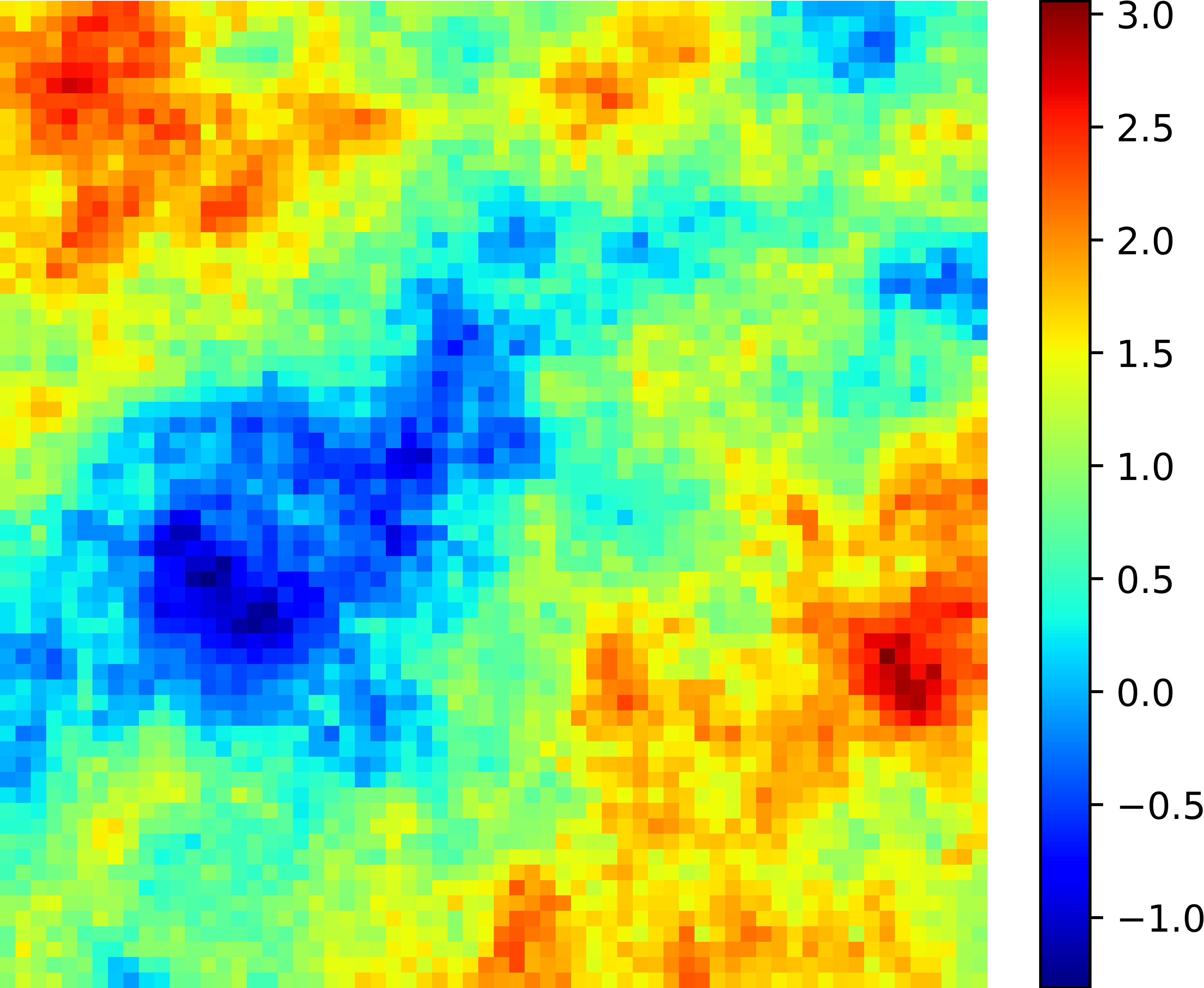}}\quad
	\subfloat[][Simulation pressure]{\includegraphics[width=.28\textwidth]{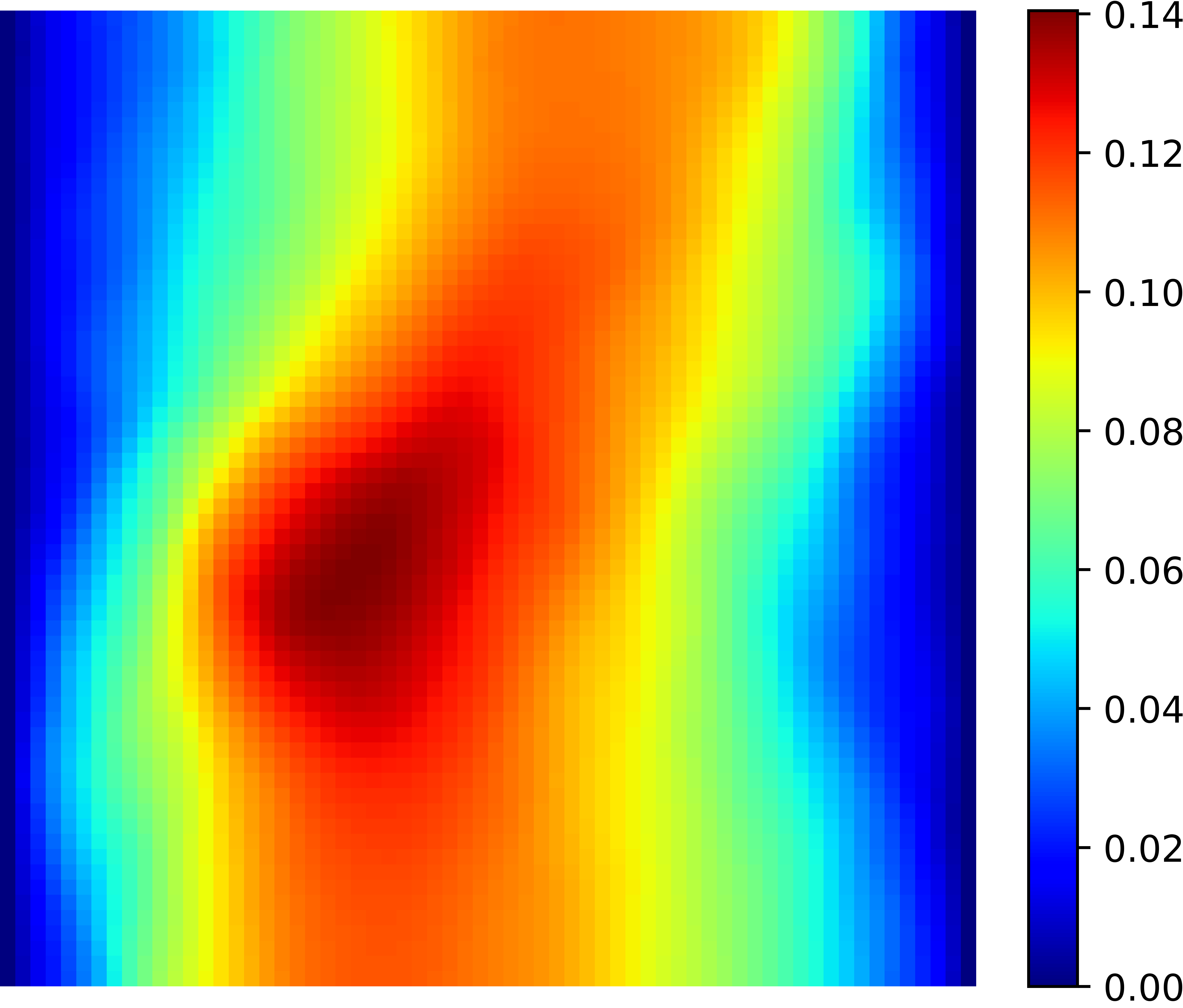}}\\
	\subfloat[][Surrogate pressure]{\includegraphics[width=.28\textwidth]{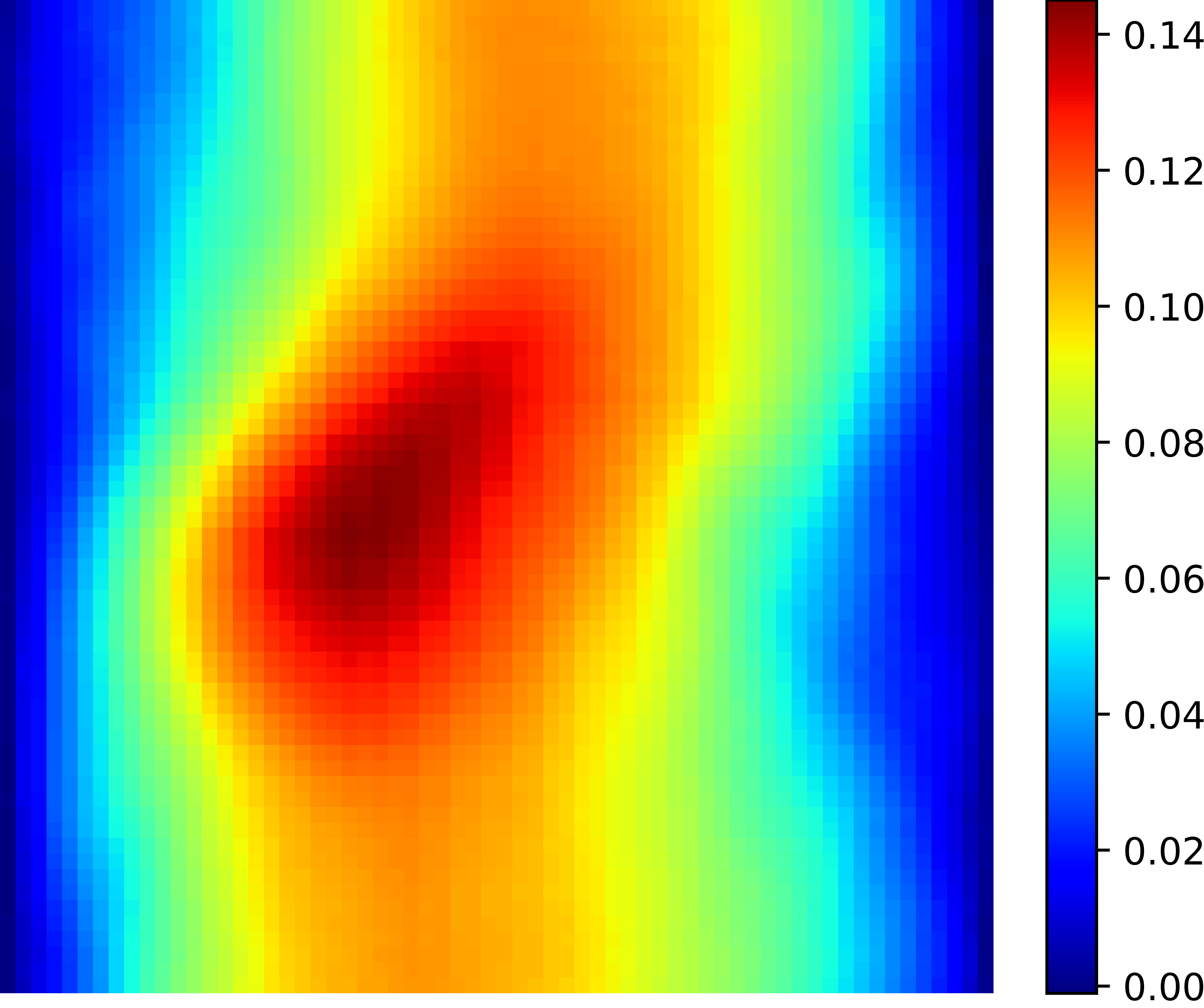}}\quad
	\subfloat[][Difference between simulation pressure and surrogate pressure]{\includegraphics[width=.28\textwidth]{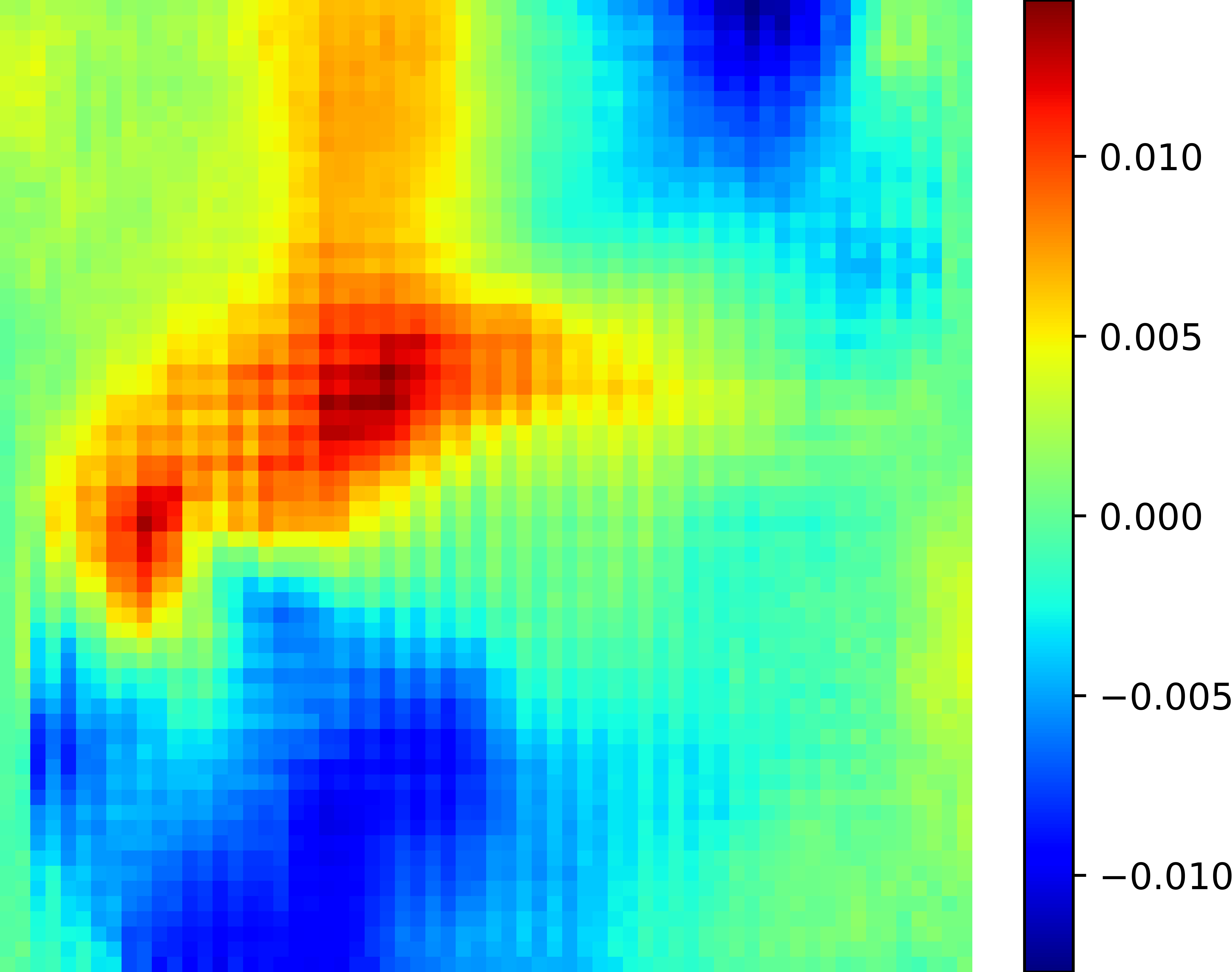}}
	\caption{Illustration of surrogate results for test problem 1.}
    \label{surrogate_plot}
\end{figure}

With the dataset $\{y^{(i)}\}_{i=1}^N$, we conduct Algorithm \ref{alg_surrogate} to train the surrogate model.  
The loss function for \eqref{diffusion} and the architecture of the surrogate model are given in \ref{surrogate_nn}. In Algorithm \ref{alg_vae_gan}, the batch size and 
the maximum epoch number
are set to $n_{batch}=100$ and  $E=100$ respectively, and the Adam optimizer is employed
 with a learning rate $\eta=0.001$. 
Figure \ref{surrogate_plot} shows the performance of the trained surrogate model by comparing the prediction of the surrogate model with the simulation given by the finite element method implemented in FEniCS \cite{langtangen2017solving}. 
The difference between the surrogate pressure $\hat{u}$ and the simulation pressure $u$ is defined by $\hat{u}-u$ (see Figure \ref{surrogate_plot}(d)). The relative errors ($\Arrowvert \hat{u}-u \Arrowvert_2/\Arrowvert u \Arrowvert_2$) is 0.05694.

Our DR-KRnet is compared with VAEprior-MCMC    
which uses MCMC to sample the posterior of the latent variable given by the same VAE prior as in DR-KRnet. We consider 64 pressure observations from locations $[0.0625 + 0.125i, 0.0625 +
0.125i],i = 0, 1, 2,\dots, 7$. 
Noisy observations are formulated by adding $5\%$ independent additive Gaussian noise to the simulated pressure field (see Figure\ \ref{surrogate_plot}(b)).

Using the pre-trained decoder and the surrogate, we seek a KRnet with Algorithm \ref{alg_krnet} to approximate the posterior in the latent space.  For KRnet, we partition the components of $x\in\mathbb{R}^{36}$ to 6 equal groups and deactivate one group after 8 affine coupling layers, where the bijection given by each coupling layer is based on the outputs of a neural network with two fully connected hidden layers of 48 neurons (More detailed about the structure of KRnet can be found in \cite{tang2020deep,adda_2022}). In Algorithm \ref{alg_krnet}, the batch size and 
the maximum epoch number
are set to $n_{batch}=100$ and  $E=5$ respectively, and the Adam optimizer is applied
 with a learning rate $\eta=0.01$. 
The sample size from standard Gaussian distribution is $I=5000$. The sample size for posterior distribution is $N_s=2000$. 
For VAEprior-MCMC, we consider the preconditioned Crank Nicolson MCMC (pCN-MCMC) method \cite{beskos2008mcmc,cotter2013mcmc} 
and then run 10000 iterations to ensure its convergence. 
For all implementations of the MCMC algorithm, the last 2000 states are retained and regarded as the posterior samples.  The corresponding acceptance rate (numbers of accepted samples
divided by the total sample size) is $30.64\%$. 

\begin{figure}
	\centering
	\subfloat[][The exact log-permeability field]{\includegraphics[width=.28
 \textwidth]{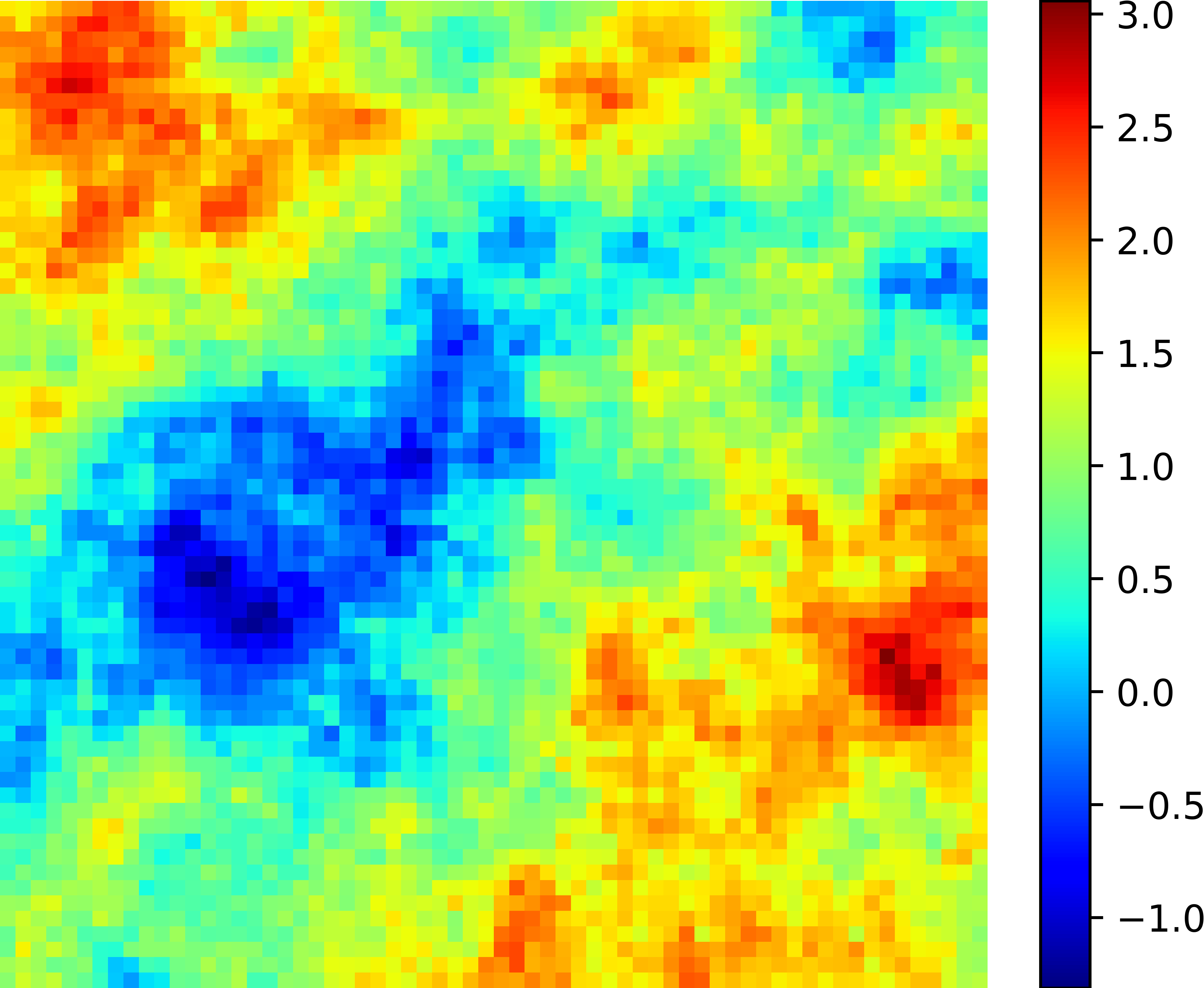}}\quad
	\subfloat[][Posterior mean]{\includegraphics[width=.28\textwidth]{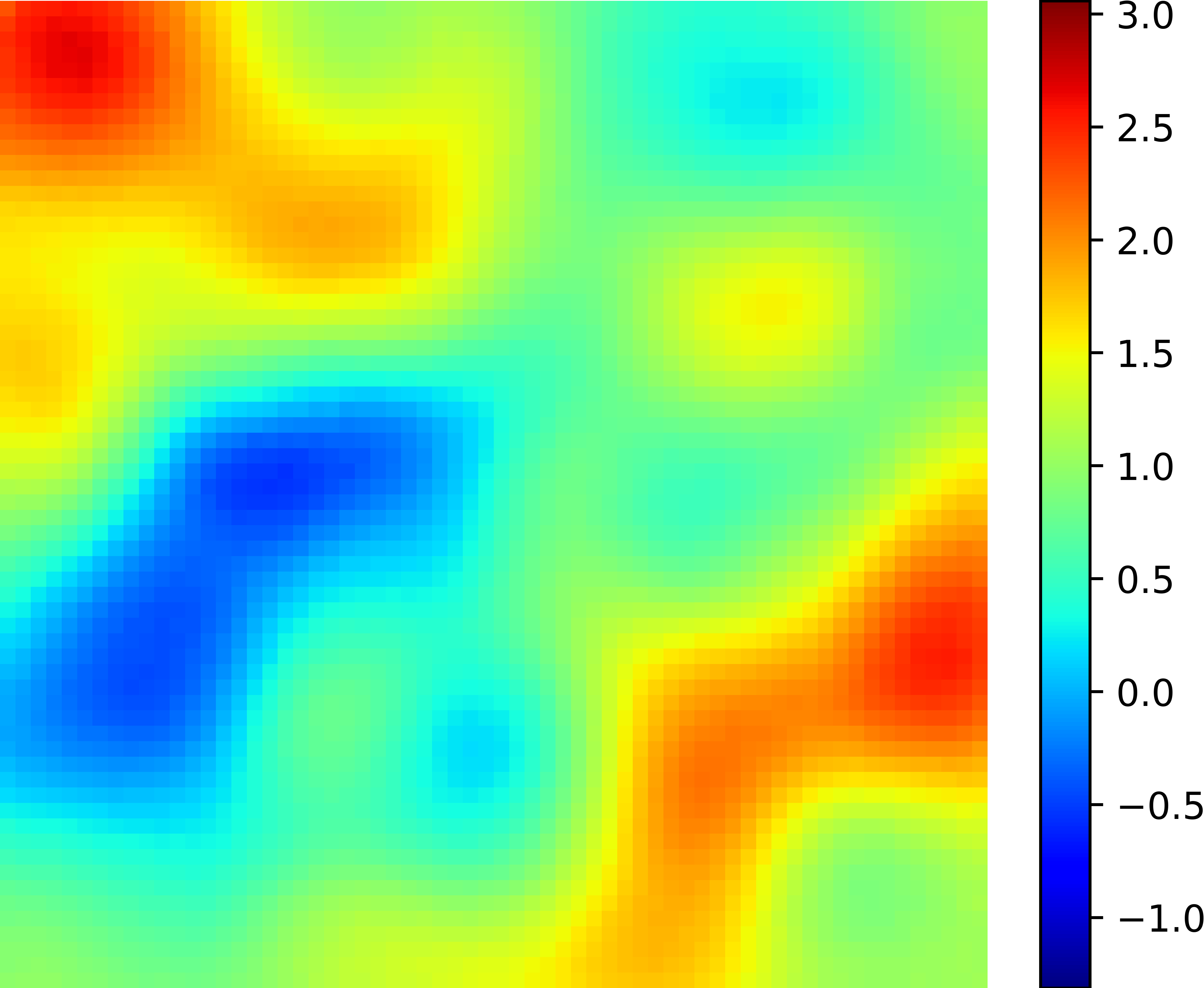}}\quad
	\subfloat[][Posterior variance]{\includegraphics[width=.28\textwidth]{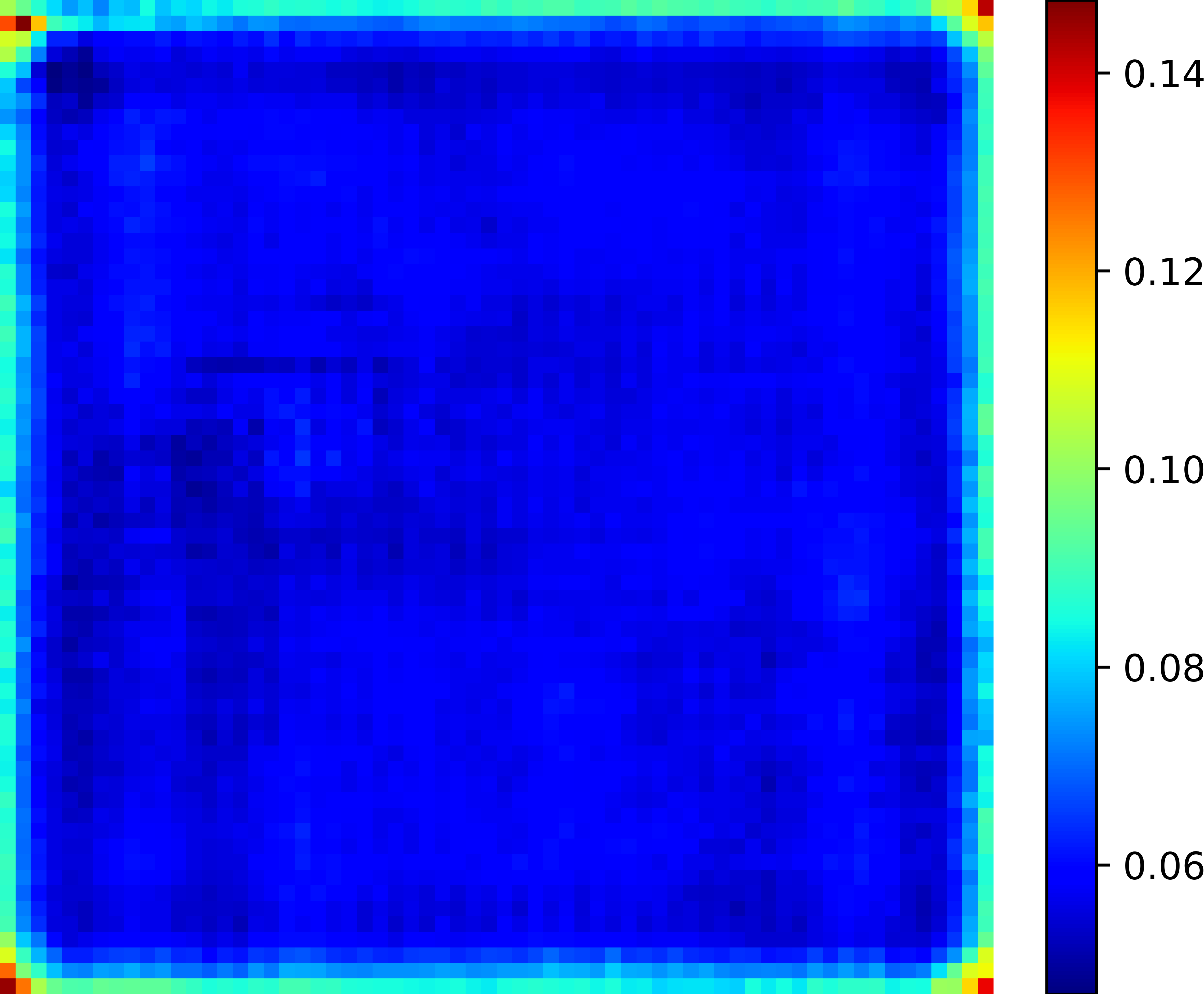}}
	\\
	\subfloat[][Posterior sample 1]{\includegraphics[width=.28\textwidth]{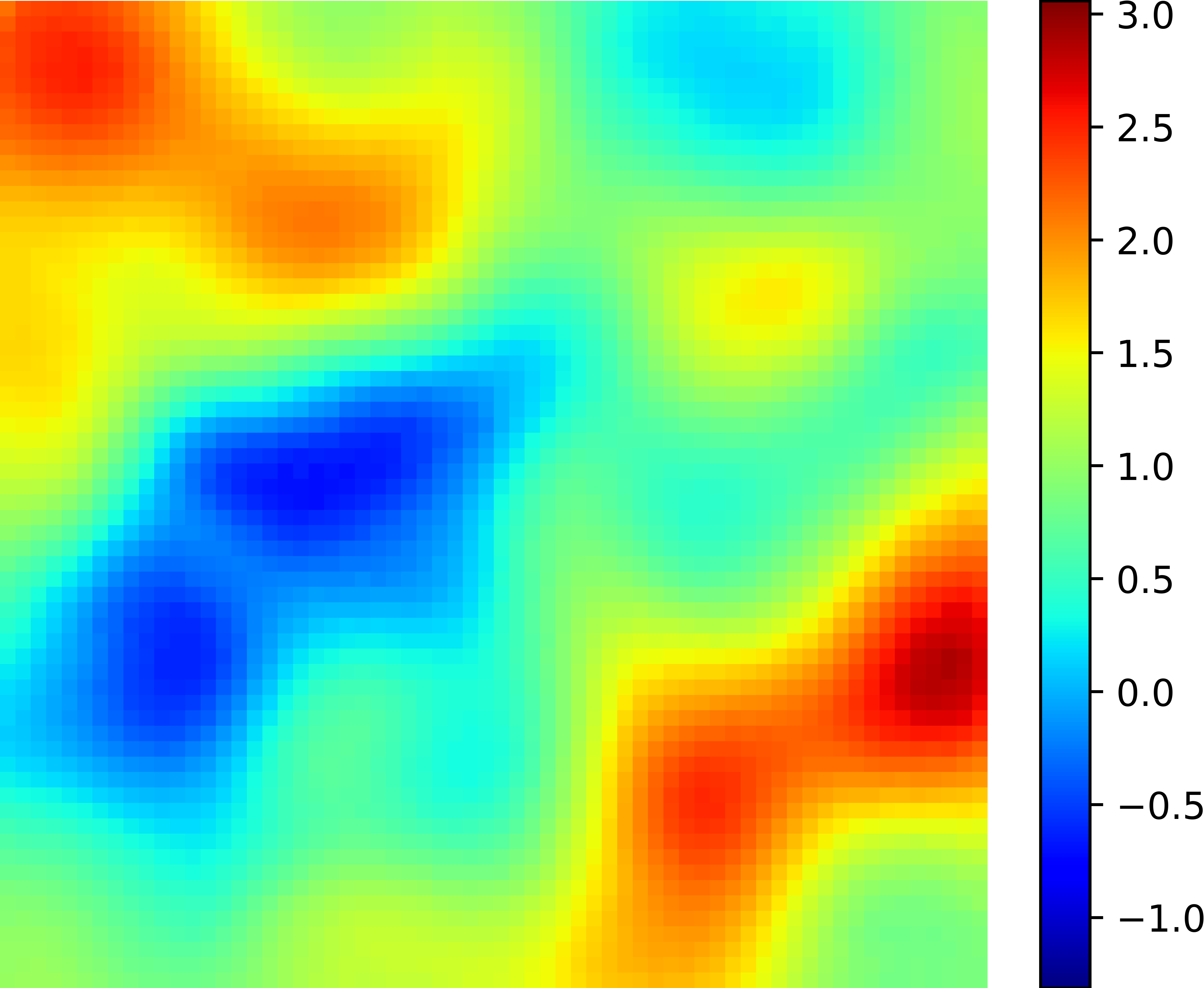}}\quad
	\subfloat[][Posterior sample 2]{\includegraphics[width=.28\textwidth]{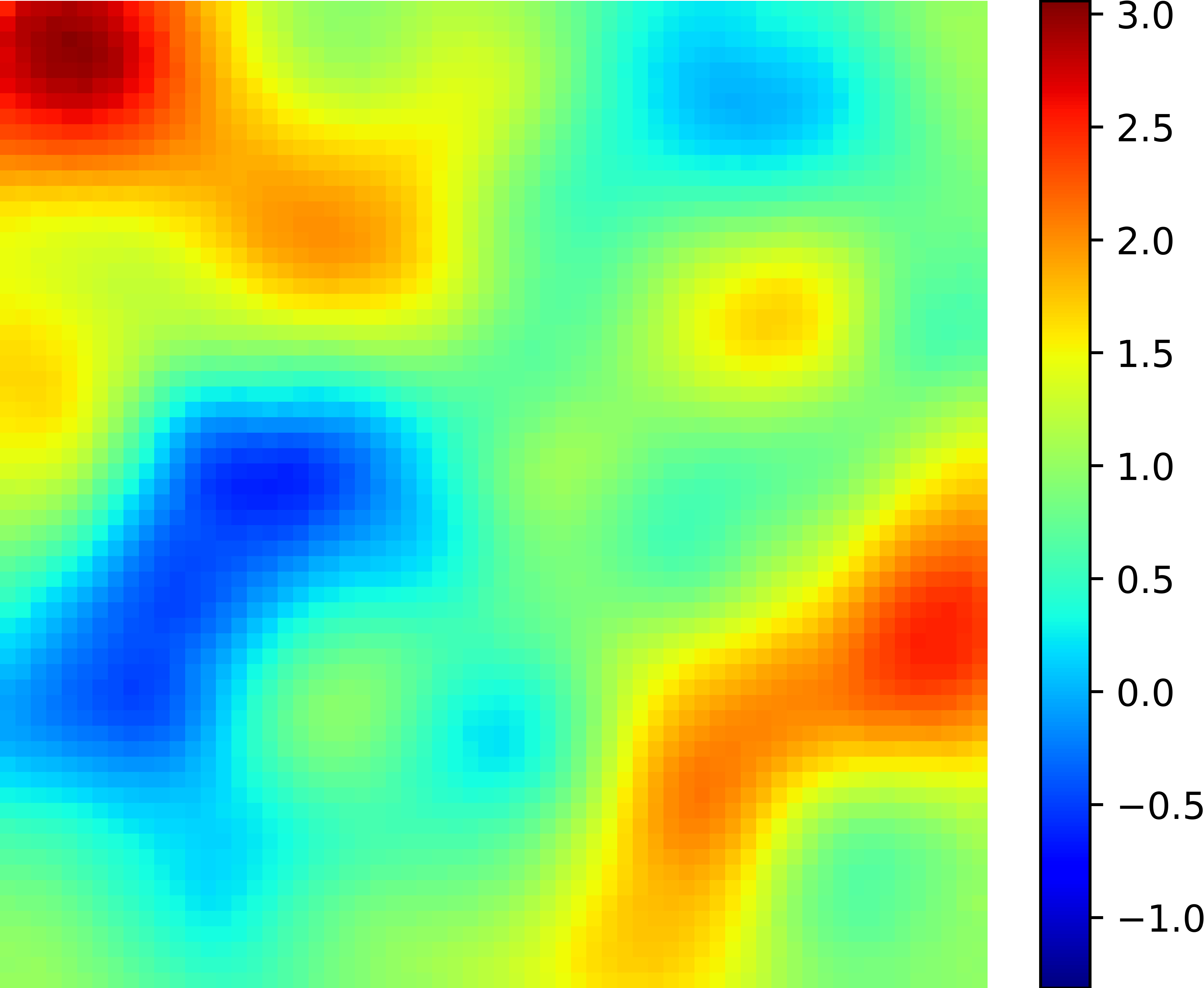}}
   \quad \subfloat[][Posterior sample 3]{\includegraphics[width=.28\textwidth]{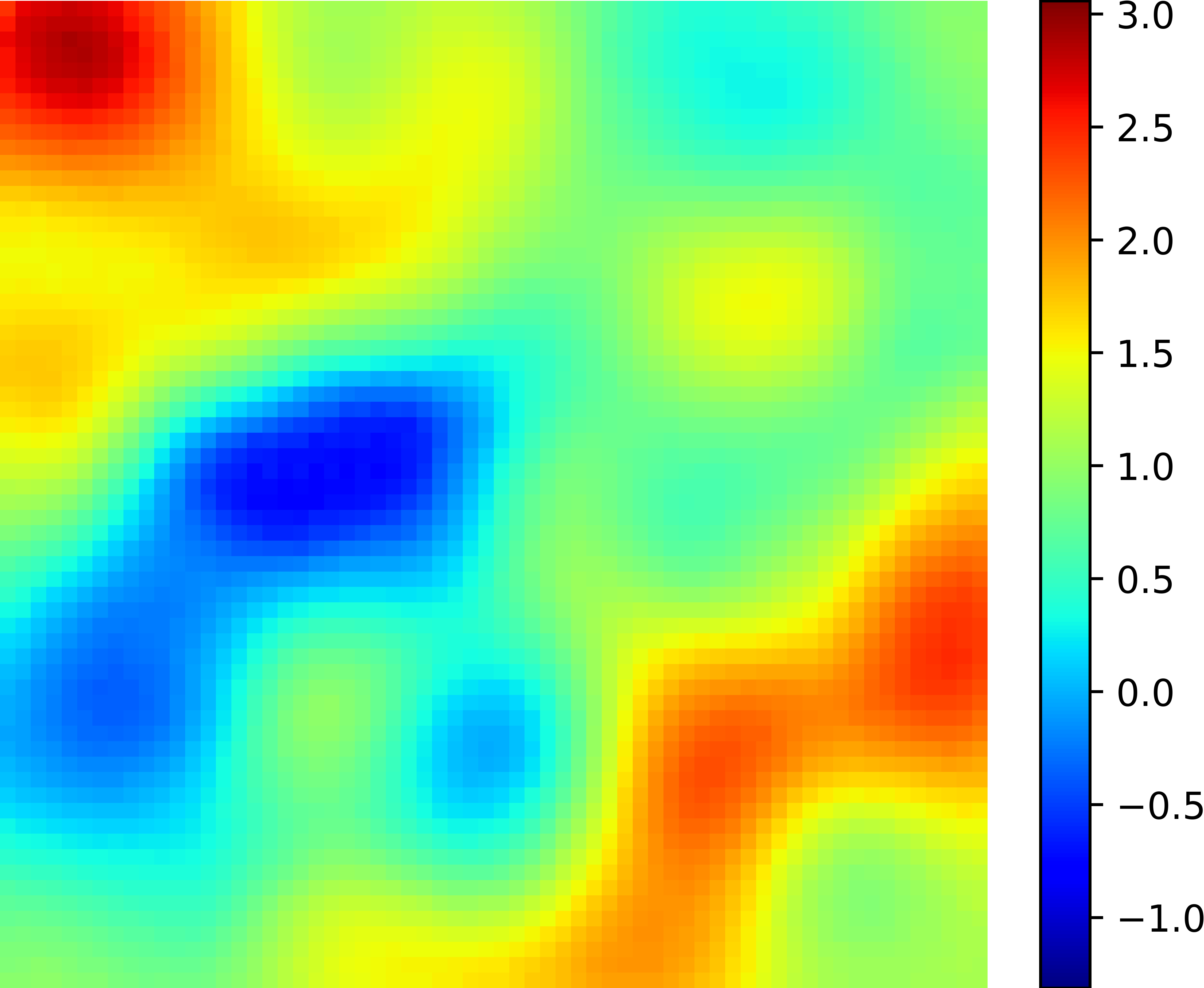}}
	\caption{The inversion results of DR-KRnet for test problem 1. }
    \label{inverse_vae}
\end{figure}

\begin{figure}
	\centering
	\subfloat[][The exact log-permeability field]{\includegraphics[width=.28
 \textwidth]{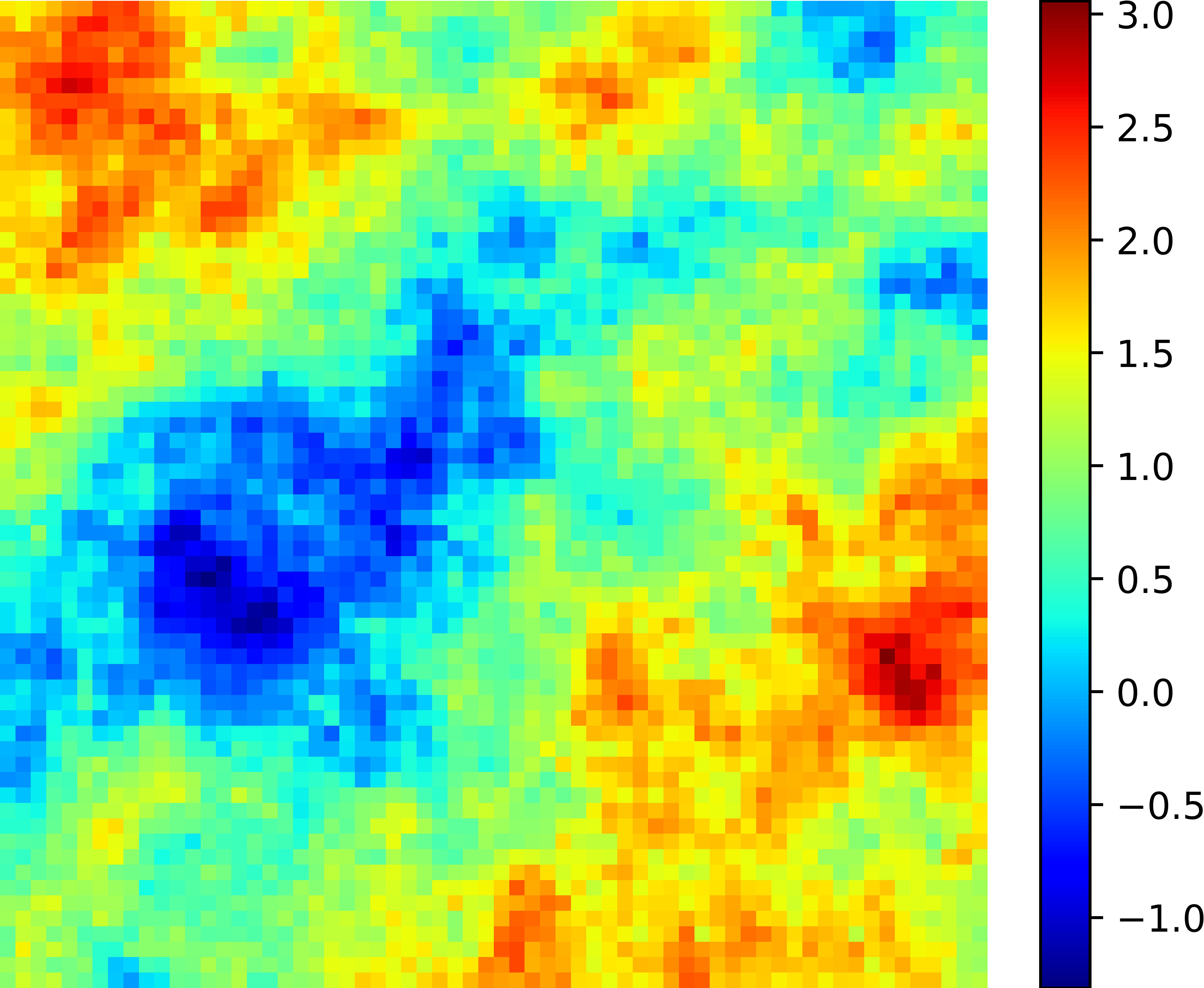}}\quad
	\subfloat[][Posterior mean]{\includegraphics[width=.28\textwidth]{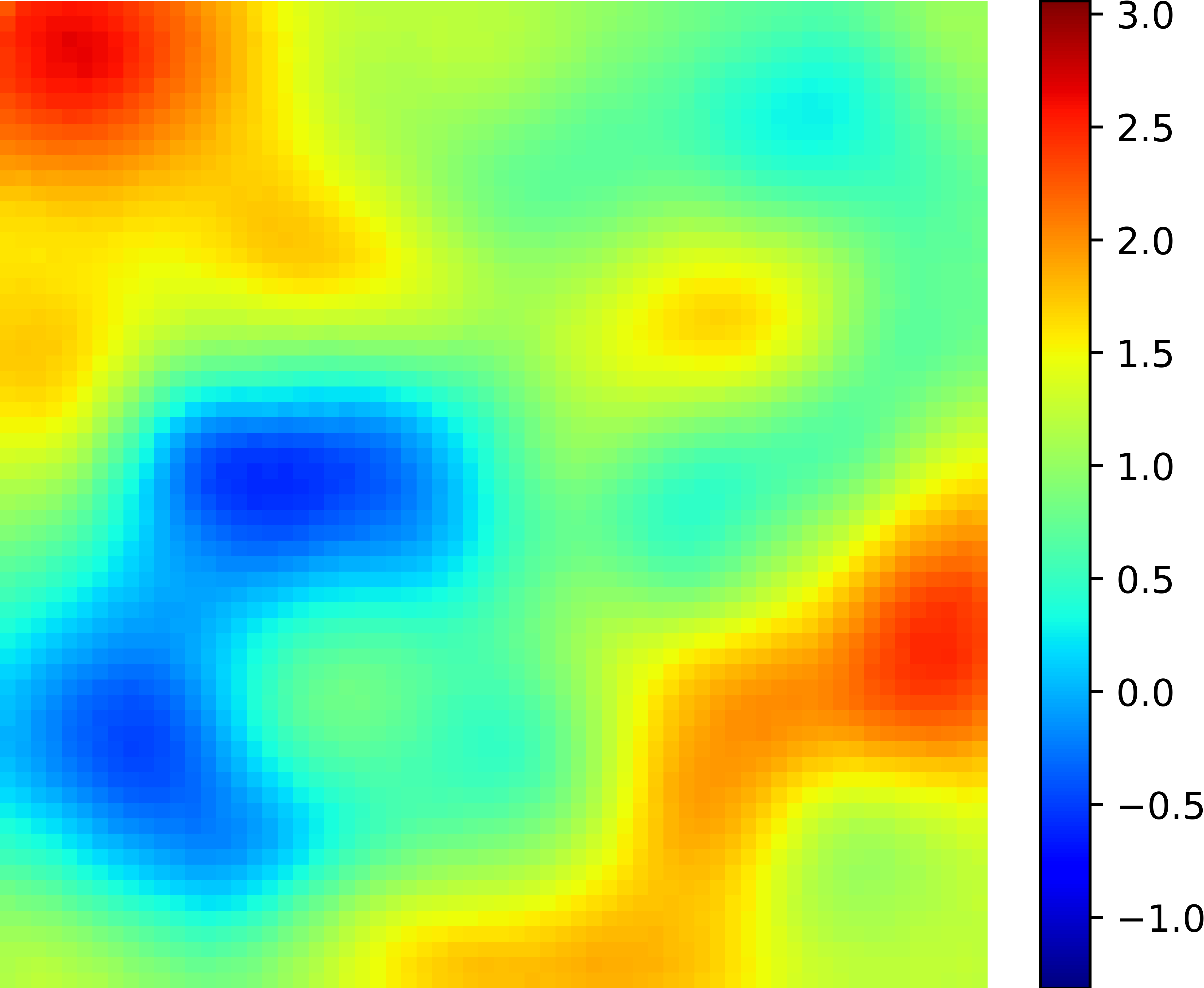}}\quad
	\subfloat[][Posterior variance]{\includegraphics[width=.28\textwidth]{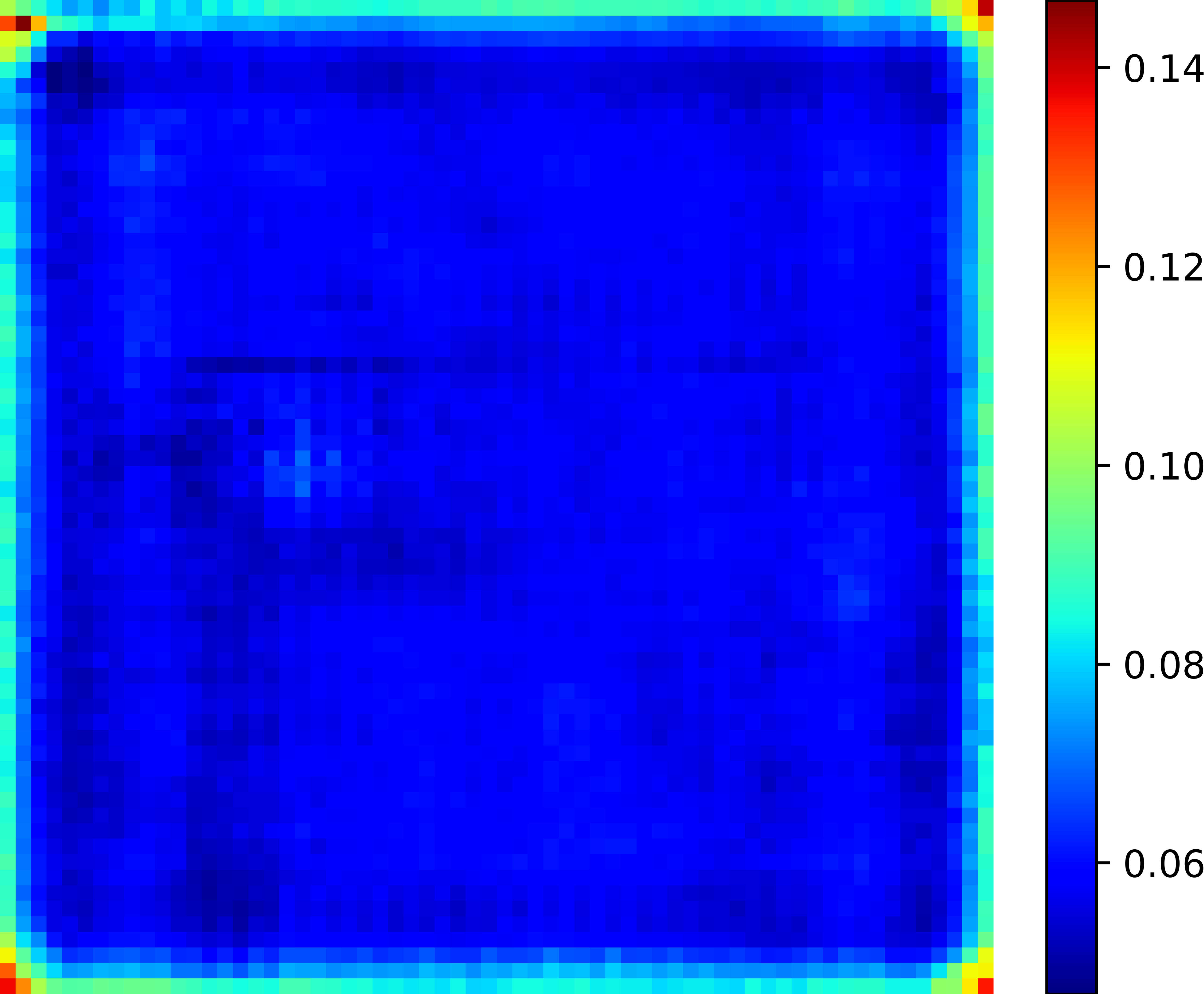}}
	\\
	\subfloat[][Posterior sample 1]{\includegraphics[width=.28\textwidth]{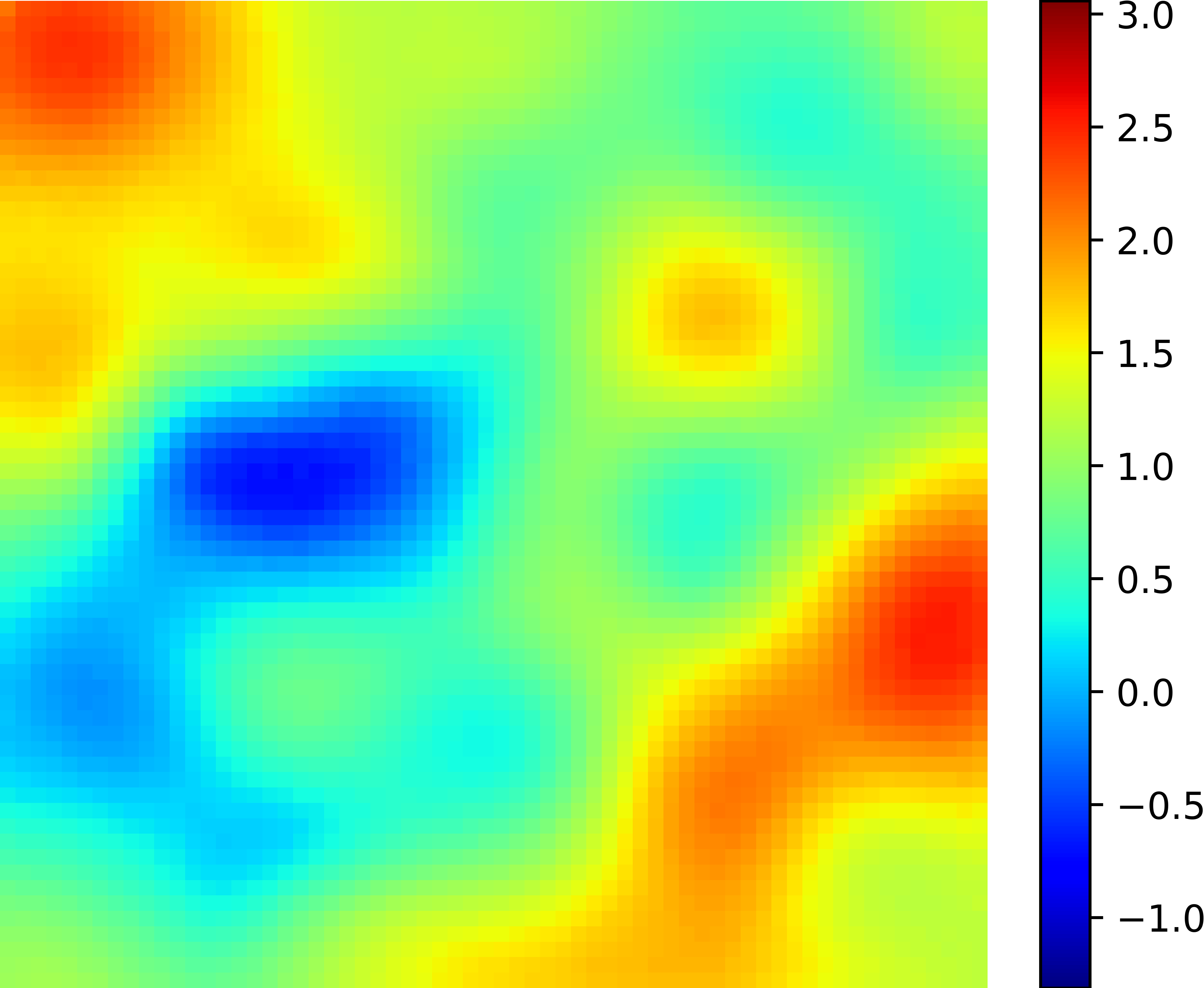}}\quad
	\subfloat[][Posterior sample 2]{\includegraphics[width=.28\textwidth]{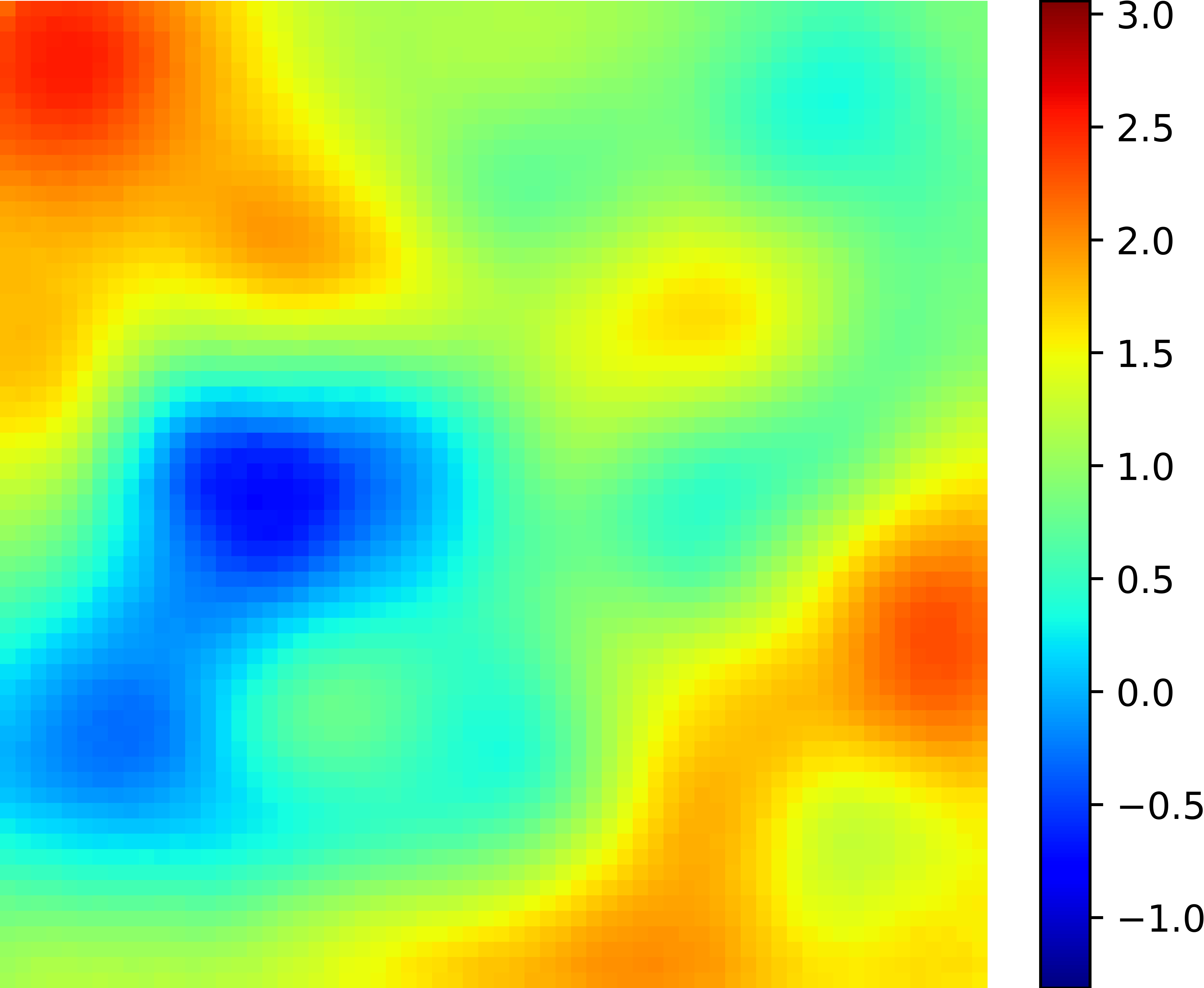}}
   \quad \subfloat[][Posterior sample 3]{\includegraphics[width=.28\textwidth]{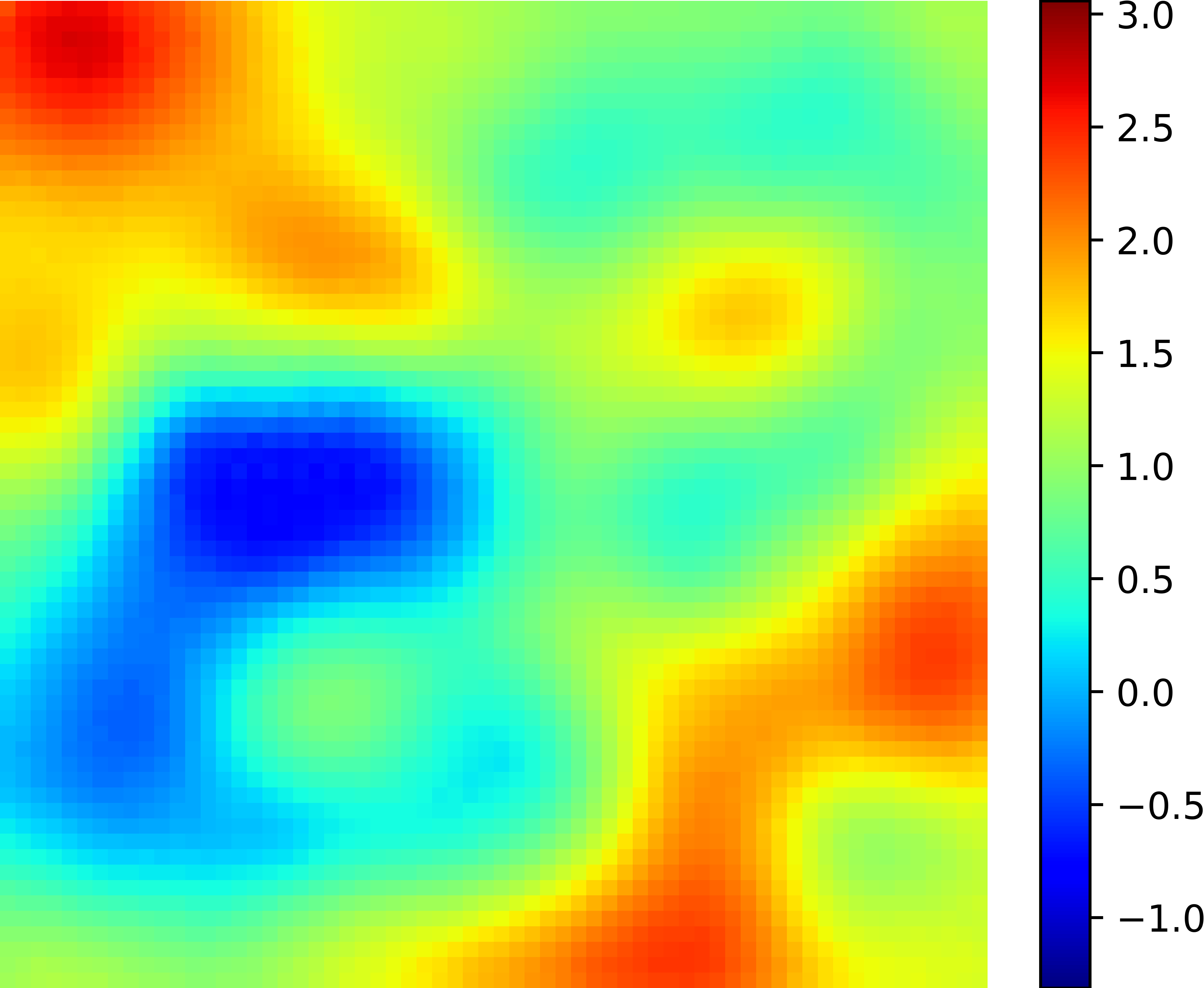}}
	\caption{The inversion results of VAEprior-MCMC for test problem 1. }
    \label{inverse_vae_mcmc}
\end{figure}

Figure \ref{inverse_vae} and Figure \ref{inverse_vae_mcmc} provide the inversion results given by DR-KRnet and VAEprior-MCMC respectively. It is seen that the two strategies yield consistent mean and variance and posterior samples. 
More details about accuracy and efficiency are presented in Table~\ref{test1}, where the relative errors are computed through \eqref{error_compute}. Time consumption for DR-KRnet (see Algorithm \ref{alg_krnet}) and VAEprior-MCMC is the computational cost of approximating the posterior of the latent variable by KRnet and MCMC respectively.
DR-KRnet yields a smaller relative error than VAEprior-MCMC with a computational cost reduced by half.



\begin{table}
		\caption{Comparisons of DR-KRnet and VAEprior-MCMC, test problem 1.}
		\label{test1}
		\centering
		\begin{tabular}{lll}
			\toprule
			 Model&  $\epsilon_{relative}$ &Time consumption\\
			\midrule
			VAEprior-MCMC &0.4014& 5.2224 minutes\\
			DR-KRnet & 0.3914&1.9608 minutes\\
			\bottomrule
		\end{tabular}
\end{table}

\subsection{Test problem 2}
\begin{figure}
	\centering
	\subfloat[][Prior sample 1]{\includegraphics[width=.28
 \textwidth]{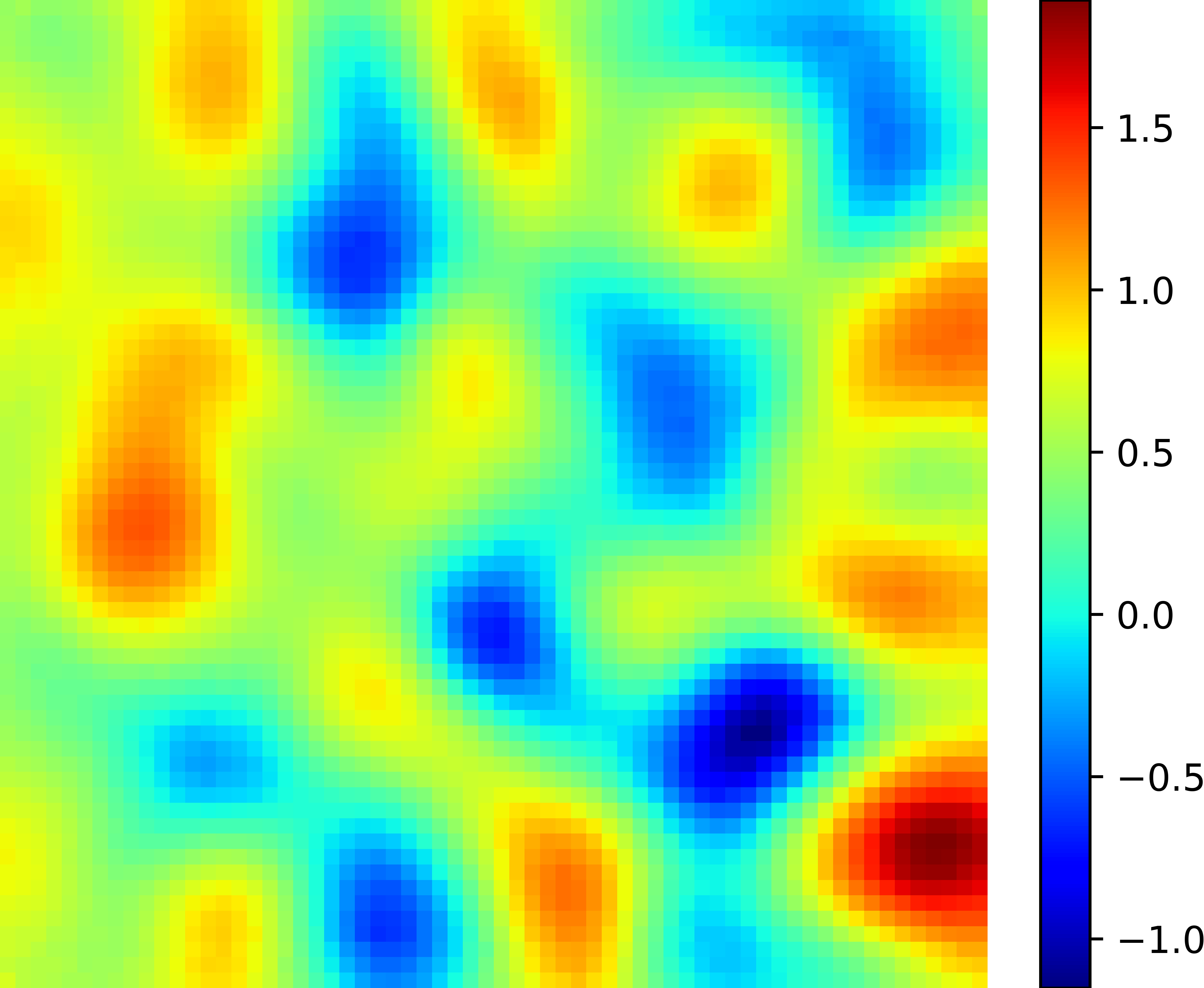}}\quad
	\subfloat[][Prior sample 2]{\includegraphics[width=.28\textwidth]{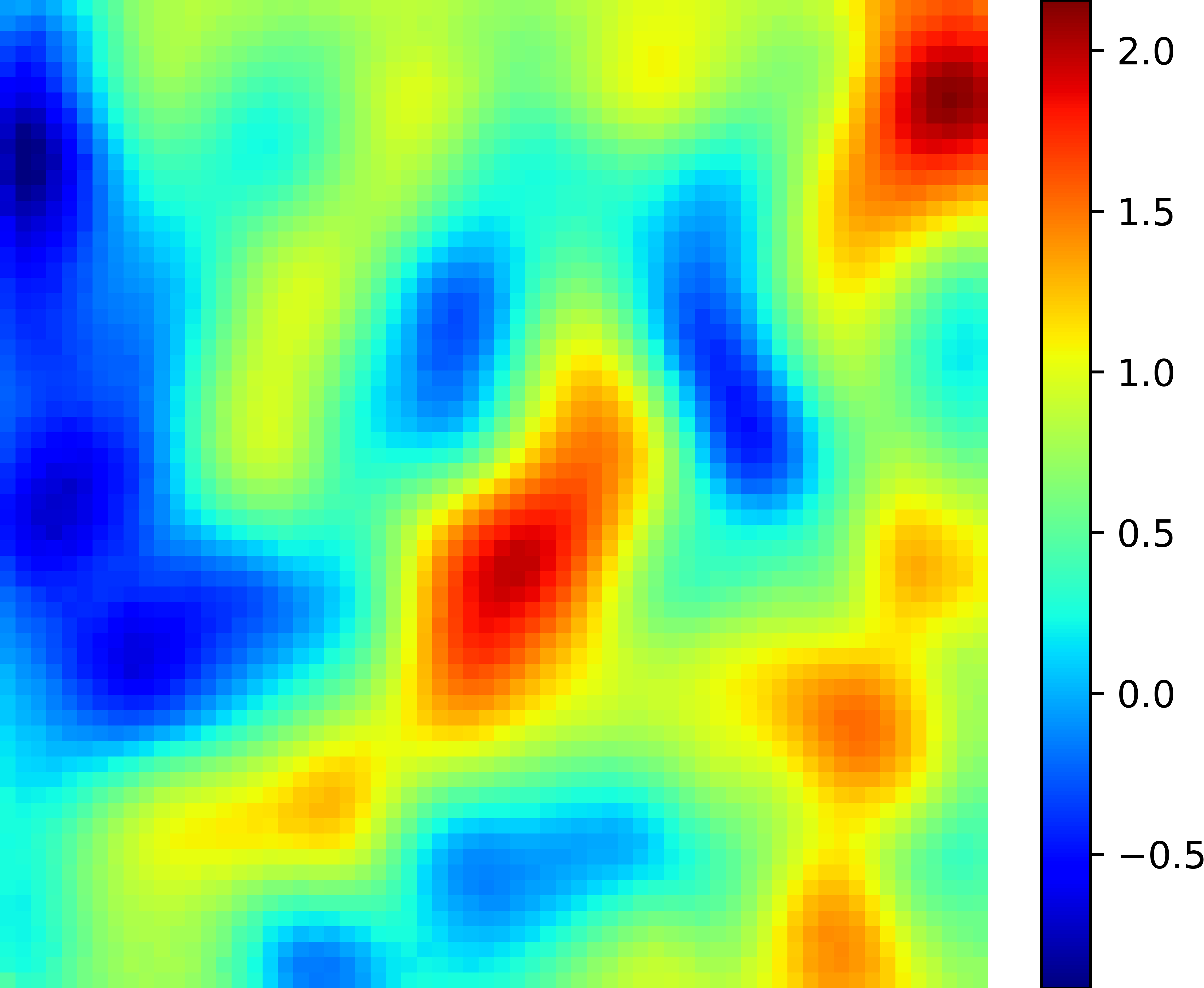}}\quad
	\subfloat[][Prior sample 3]{\includegraphics[width=.28\textwidth]{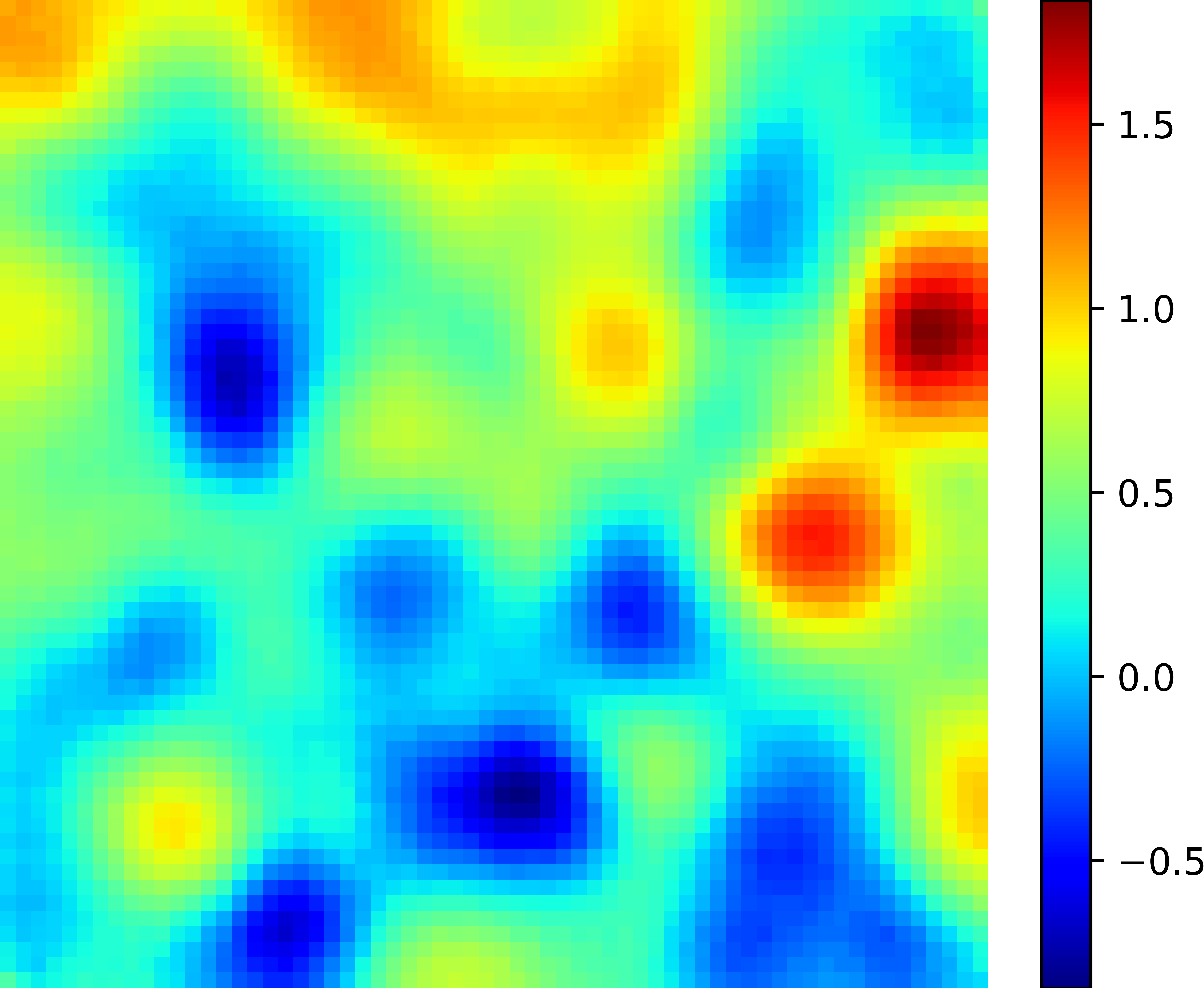}}
	\\
	\subfloat[][Prior sample 4]{\includegraphics[width=.28\textwidth]{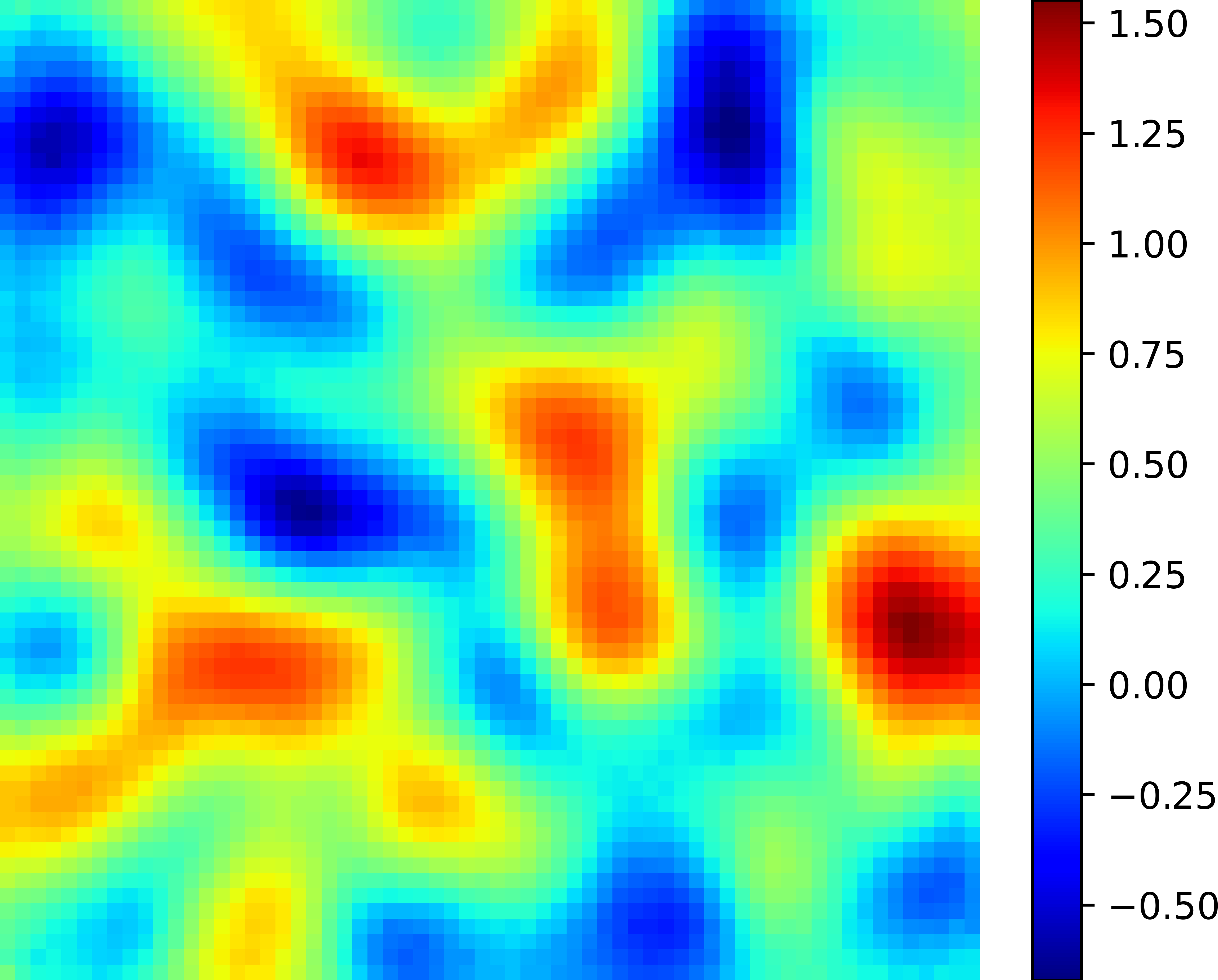}}\quad
	\subfloat[][Prior sample 5]{\includegraphics[width=.28\textwidth]{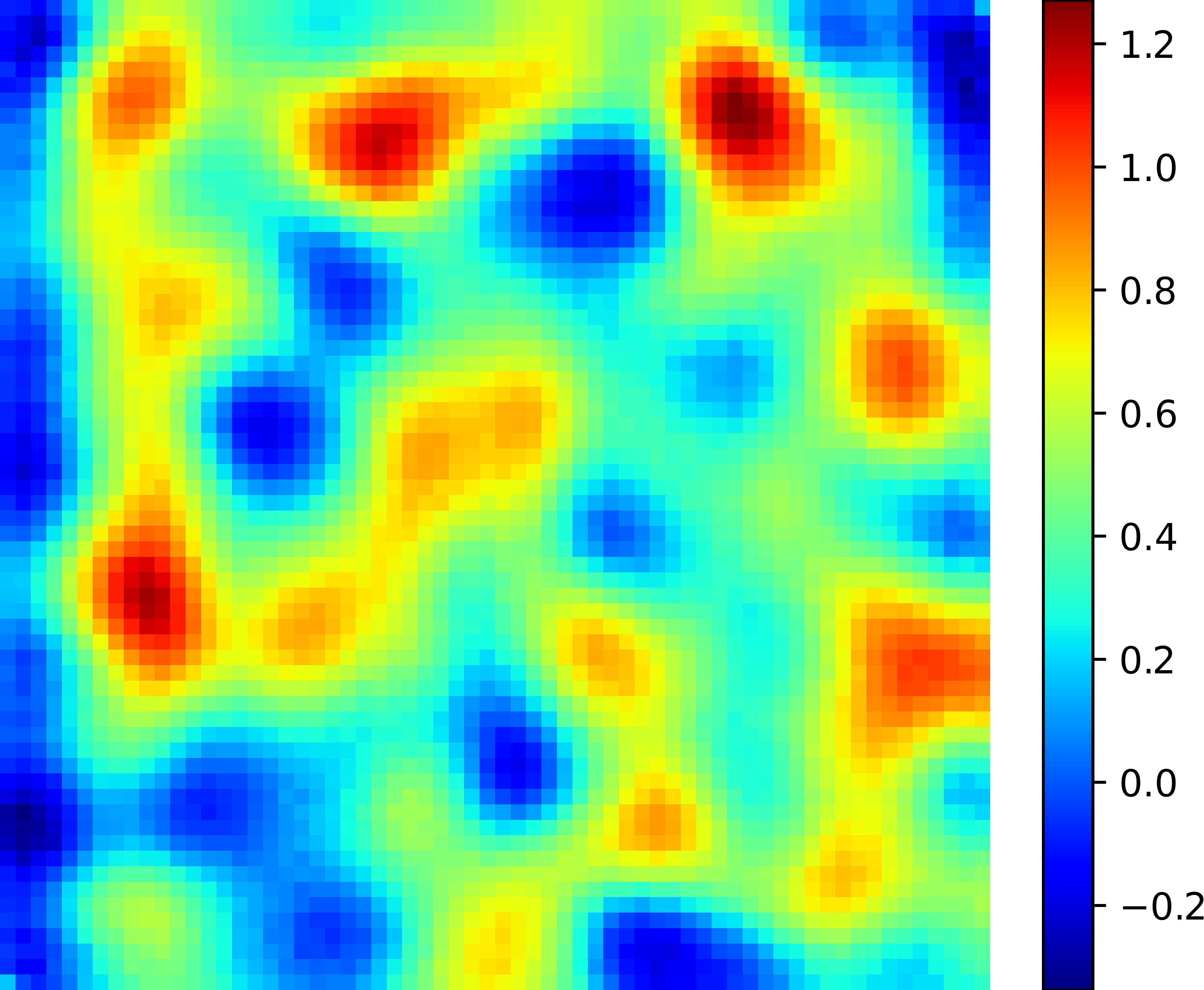}}
   \quad \subfloat[][Prior sample 6]{\includegraphics[width=.28\textwidth]{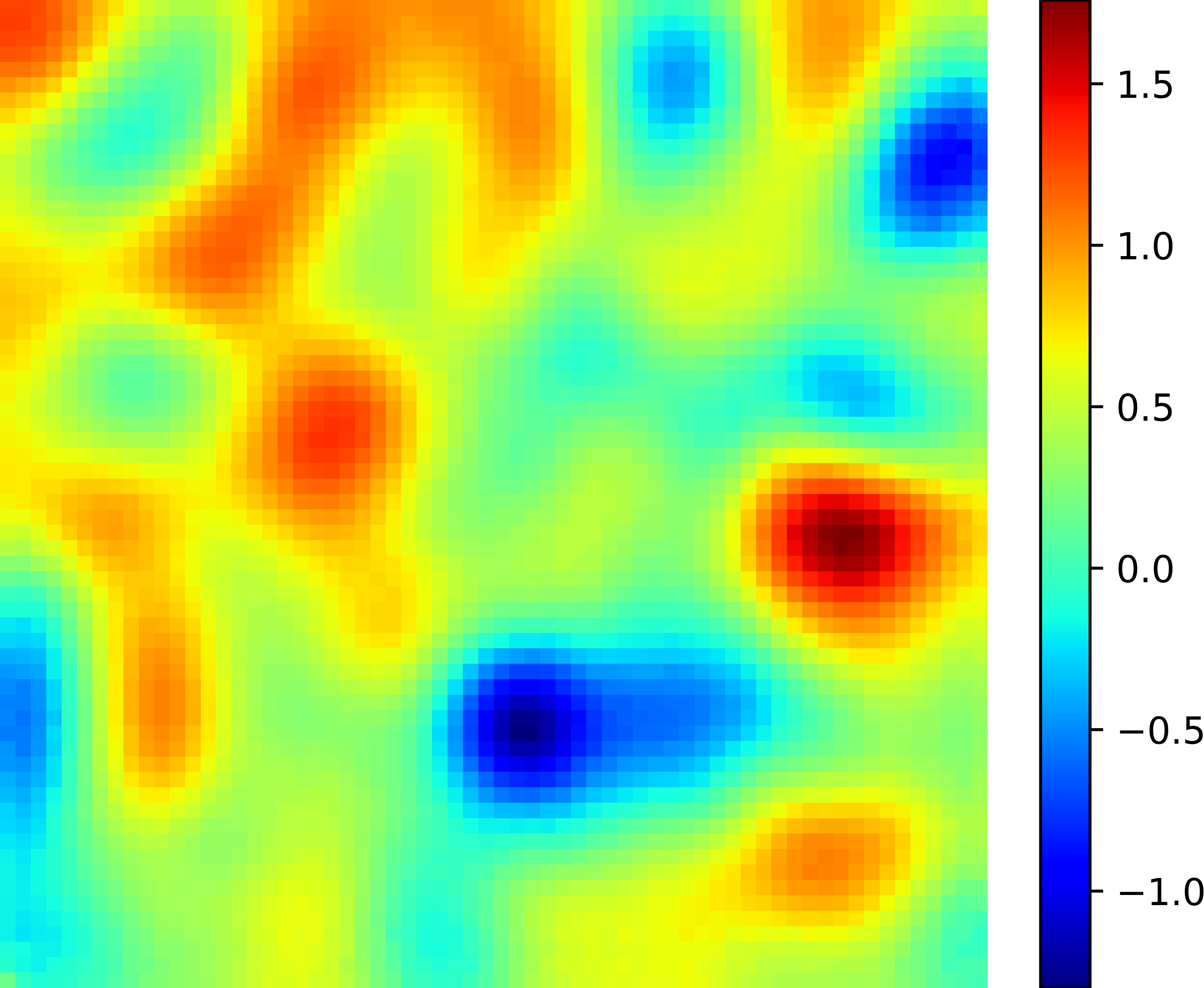}}
	
	\caption{64×64 resolution random samples generated from $p_{y|x,\theta^*}p_x$ for test problem 2, where $x\in \mathbb{R}^{64}$, $p_{y|x,\theta^*}$ is the decoder of the VAE prior and  $p_x=\mathcal{N} (0, \mathbf{I})$.}
    \label{samplesVAE_highdim}
\end{figure}

In this case, we consider a larger $d_{KL}$ subject to smaller correlation lengths when generating the prior dataset. 
Using $N=20000$ images as the prior dataset, we train the VAE priors with Algorithm \ref{alg_vae_gan}, where the architecture of the neural
networks is described in \ref{vae_nn}. Here the hyperparameters are the same as those of test problem 1 
except that the dimension of the latent variable is increased to 64. 
Figure \ref{samplesVAE_highdim} includes 6 realizations given by the decoder of the trained VAE prior. 

\begin{figure}
	\centering
	\subfloat[][The exact log-permeability]{\includegraphics[width=.28
 \textwidth]{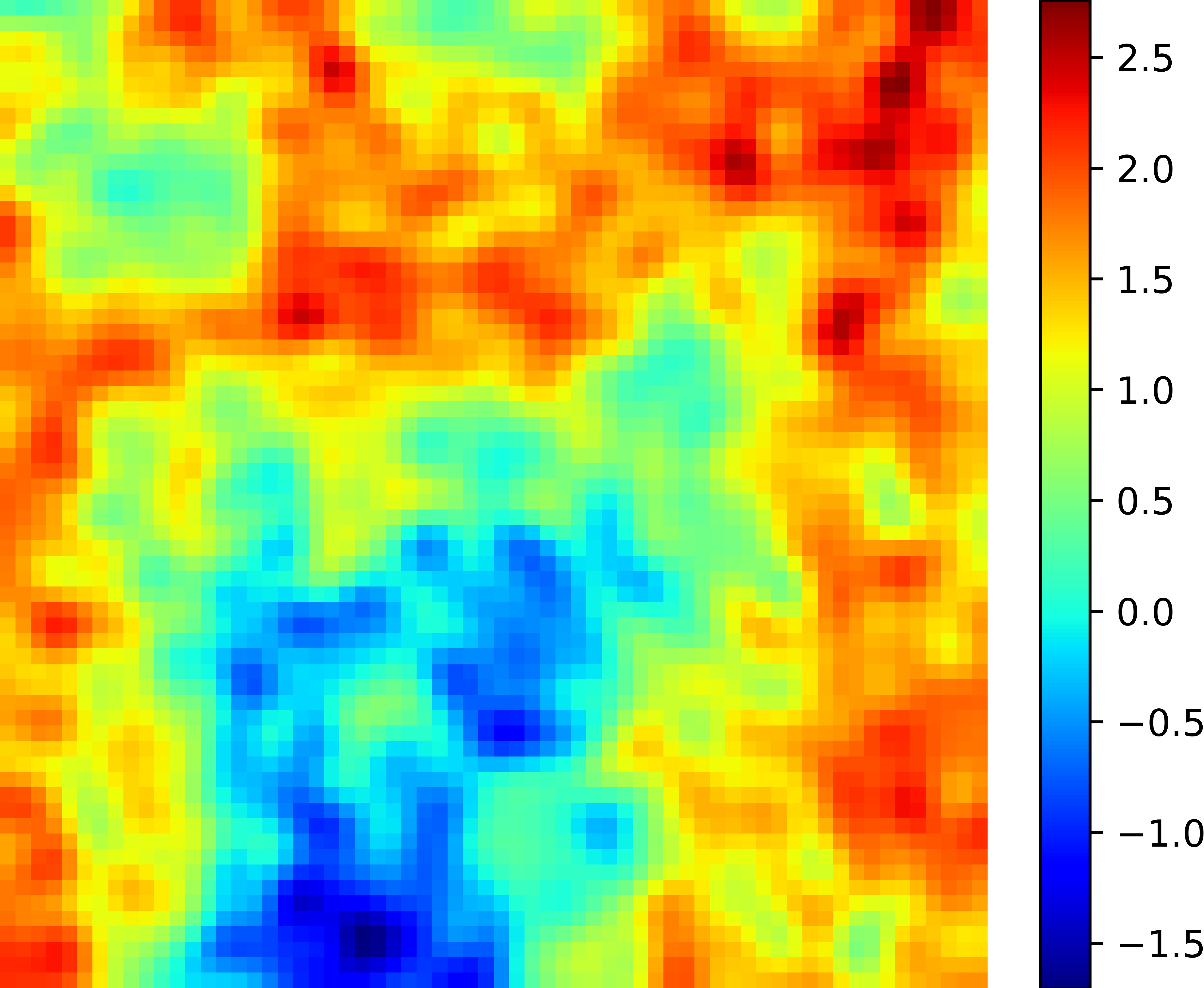}}\quad
	\subfloat[][Simulation pressure]{\includegraphics[width=.28\textwidth]{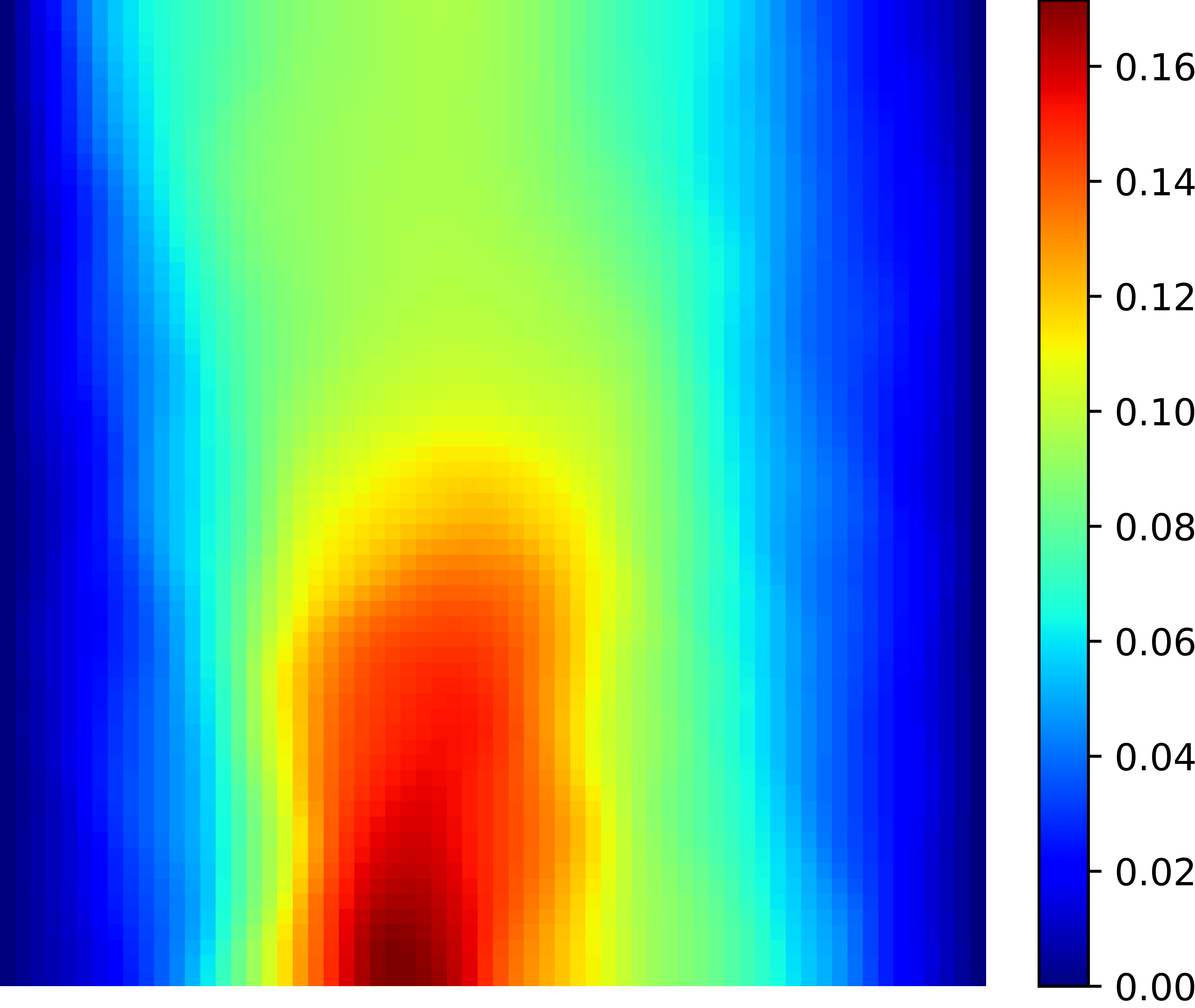}}\\
	\subfloat[][Surrogate pressure]{\includegraphics[width=.28\textwidth]{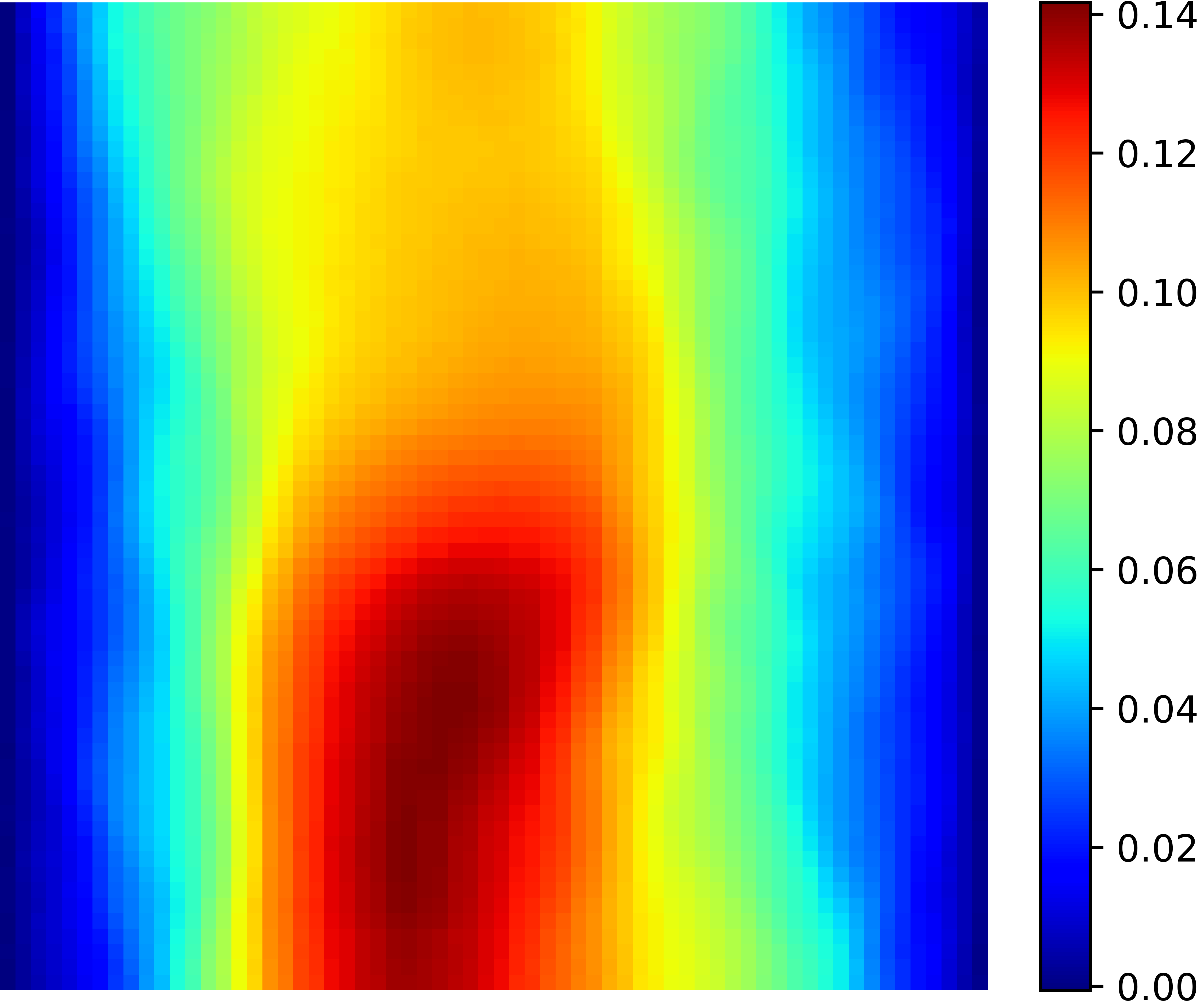}}\quad
	\subfloat[][Difference between simulation pressure and surrogate pressure]{\includegraphics[width=.28\textwidth]{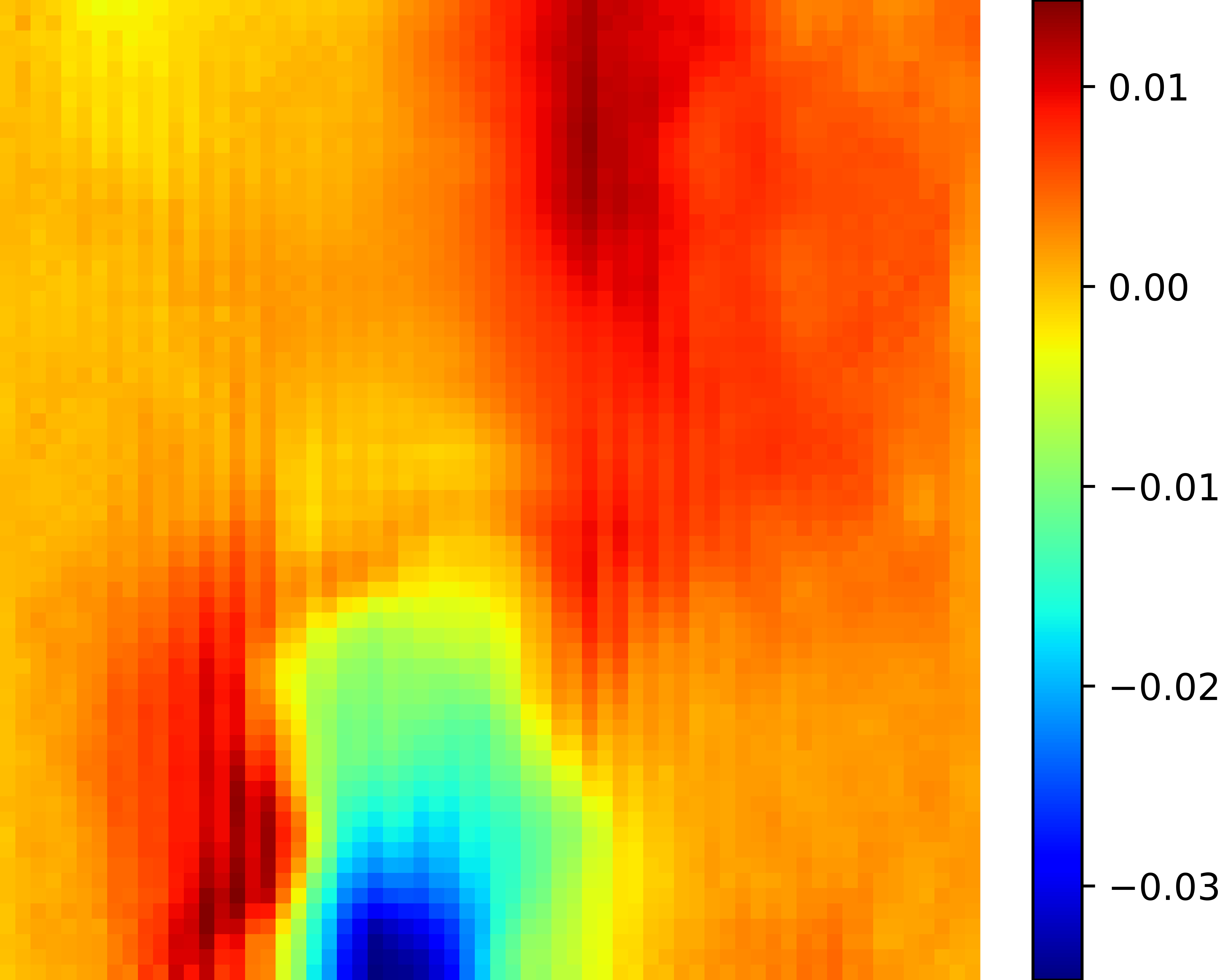}}
	\caption{Illustration of surrogate results for test problem 2.}
    \label{highdim_surrogate_plot}
\end{figure}

The setups and hyperparameters of the surrogate model are the same as those of test problem 1. The performance of the surrogate model is illustrated in Figure \ref{highdim_surrogate_plot}, where the simulated pressure field given by the finite element method and the predicted pressure field given by the surrogate model are shown in Figure \ref{highdim_surrogate_plot}(b)--(c) respectively for the log-permeability shown in Figure \ref{highdim_surrogate_plot}(a). 
The difference between the surrogate pressure $\hat{u}$ and the simulation pressure $u$ is defined by $\hat{u}-u$ shown in Figure \ref{highdim_surrogate_plot}(d).
The relative errors ($\Arrowvert \hat{u}-u \Arrowvert_2/\Arrowvert u \Arrowvert_2$) is 0.08260.
Compared to test problem 1, the prediction of the surrogate model captures the solution sufficiently well with a slight loss in accuracy.    

Since test problem 2 is more challenging than test problem 1, we consider 225 pressure observations that are uniformly located in $[0.0625 + 0.125i, 0.0625 + 0.0625i],\,i = 0, 1, 2,\dots, 14$. The observations are generated from the simulated pressure field by adding $1\%$ independent additive Gaussian noise. 

\begin{figure}
	\centering
	\subfloat[][The exact log-permeability field]{\includegraphics[width=.28
 \textwidth]{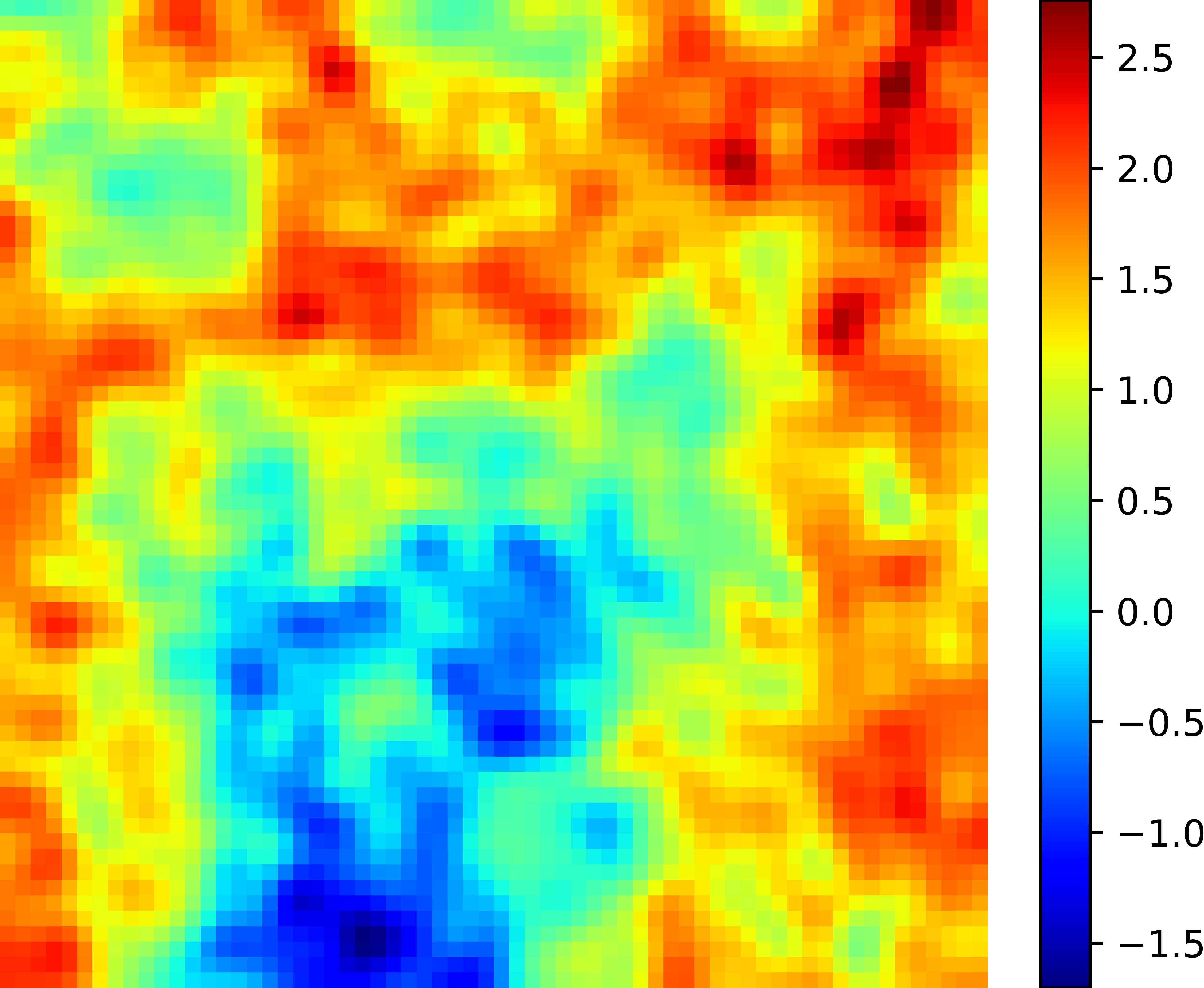}}\quad
	\subfloat[][Posterior mean]{\includegraphics[width=.28\textwidth]{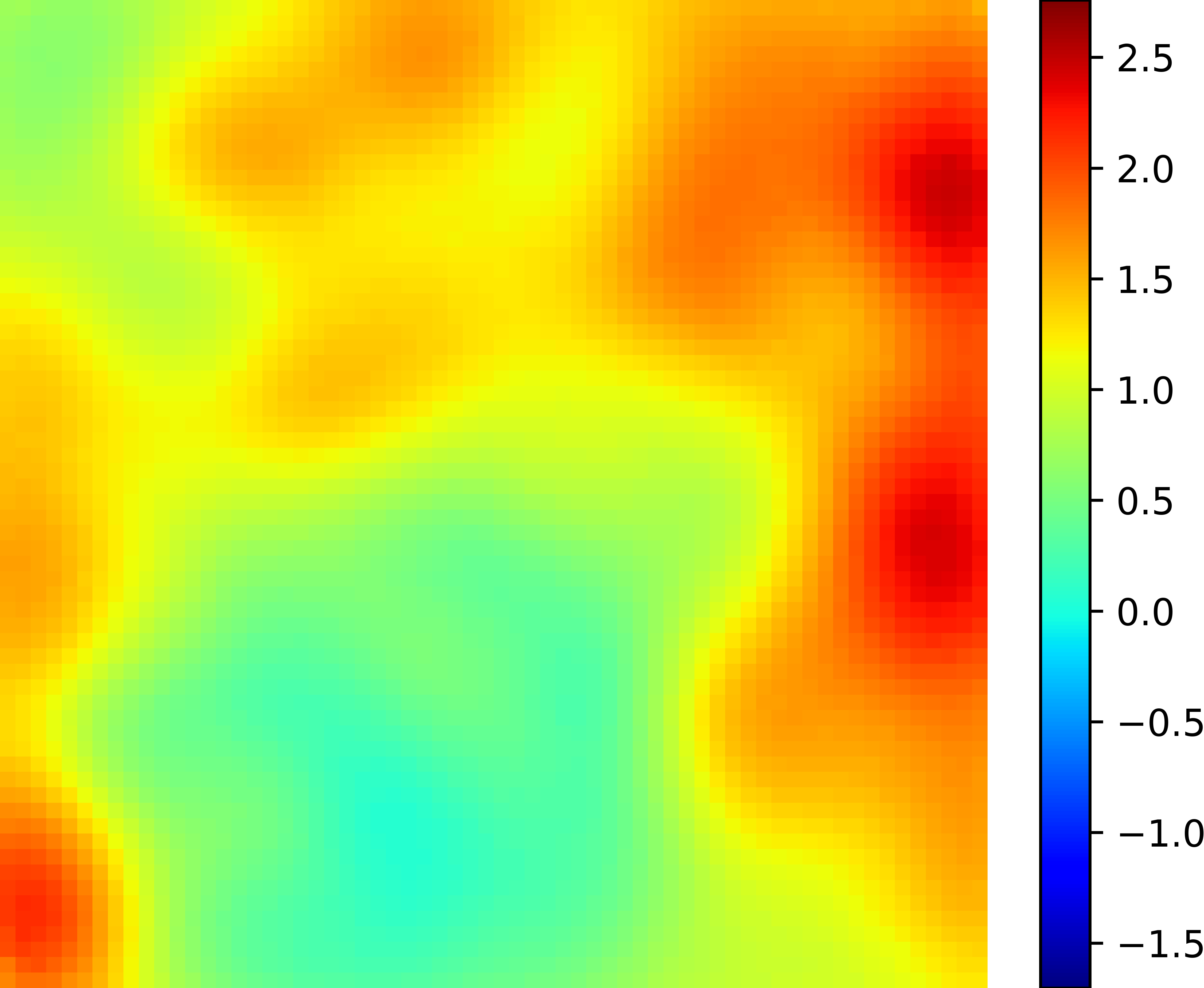}}\quad
	\subfloat[][Posterior variance]{\includegraphics[width=.3\textwidth]{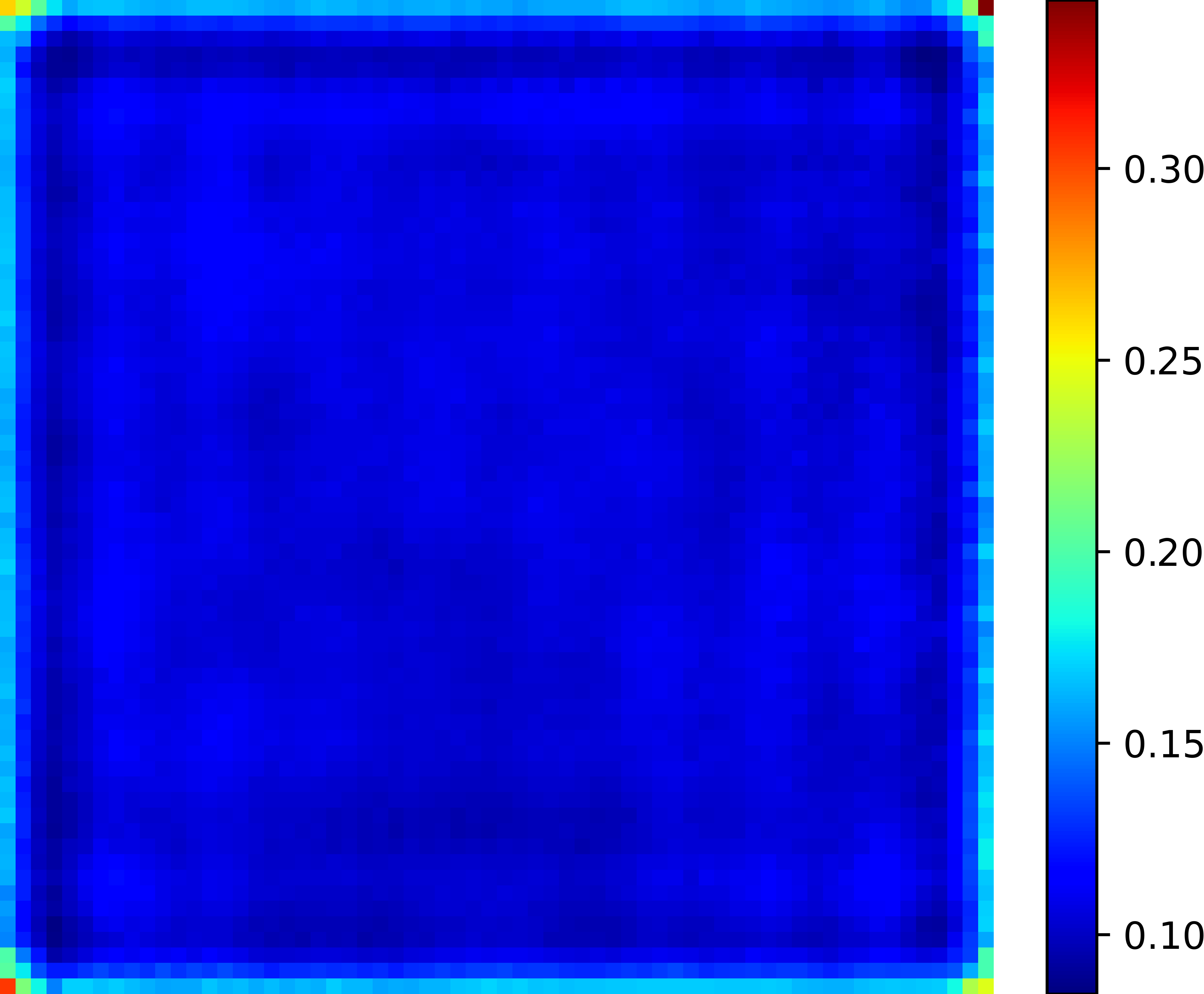}}
	\\
	\subfloat[][Posterior sample 1]{\includegraphics[width=.28\textwidth]{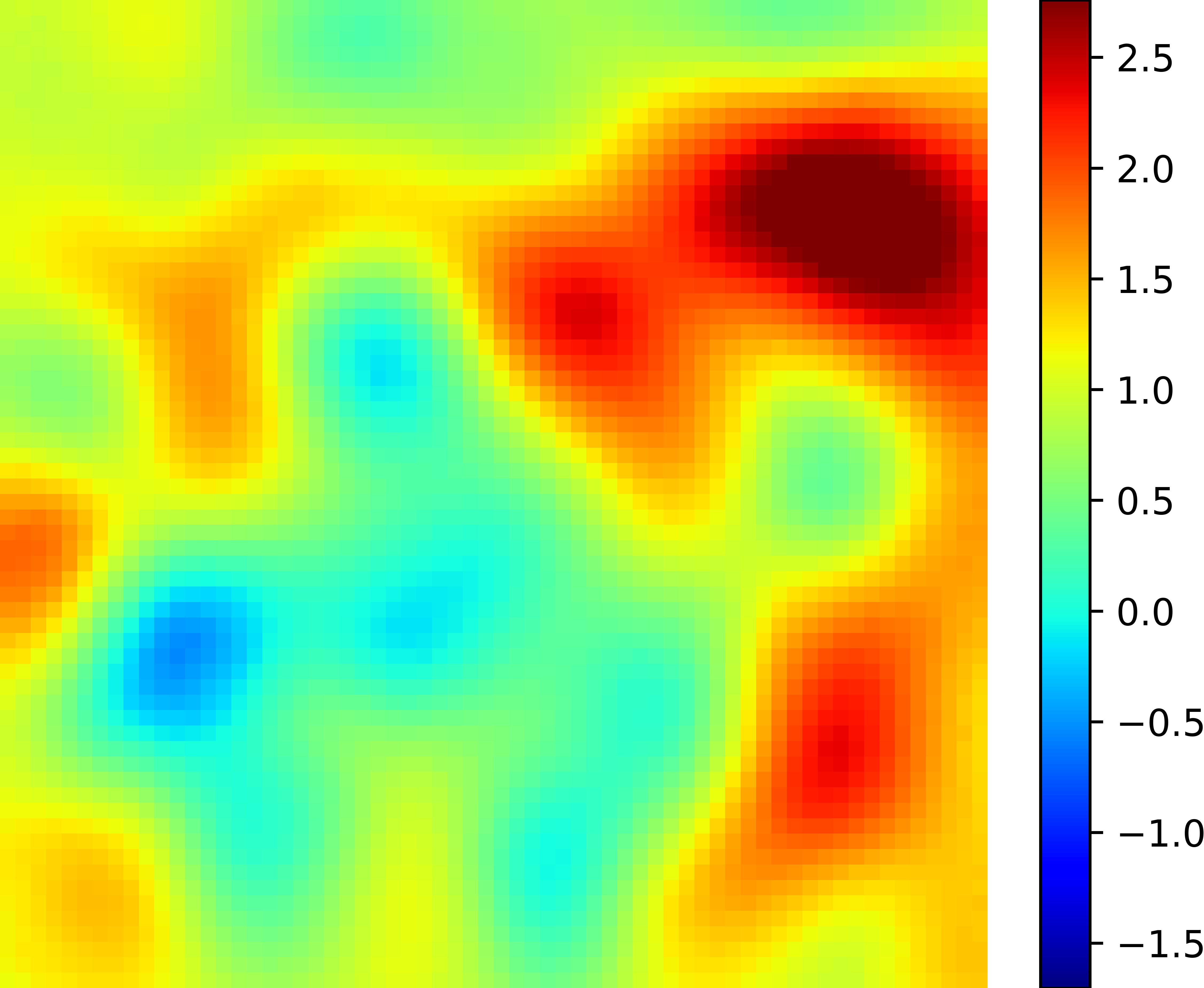}}\quad
	\subfloat[][Posterior sample 2]{\includegraphics[width=.28\textwidth]{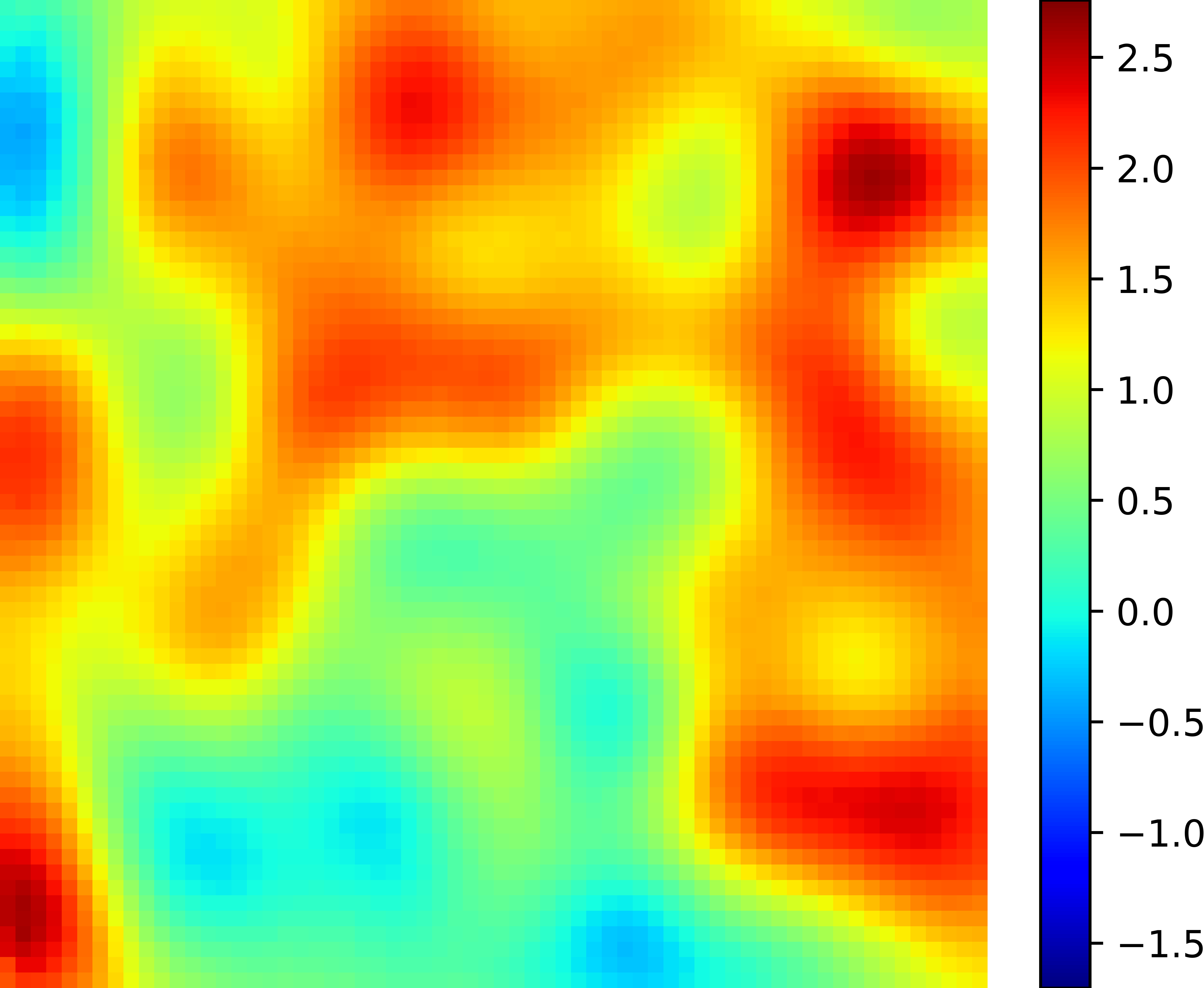}}
   \quad \subfloat[][Posterior sample 3]{\includegraphics[width=.28\textwidth]{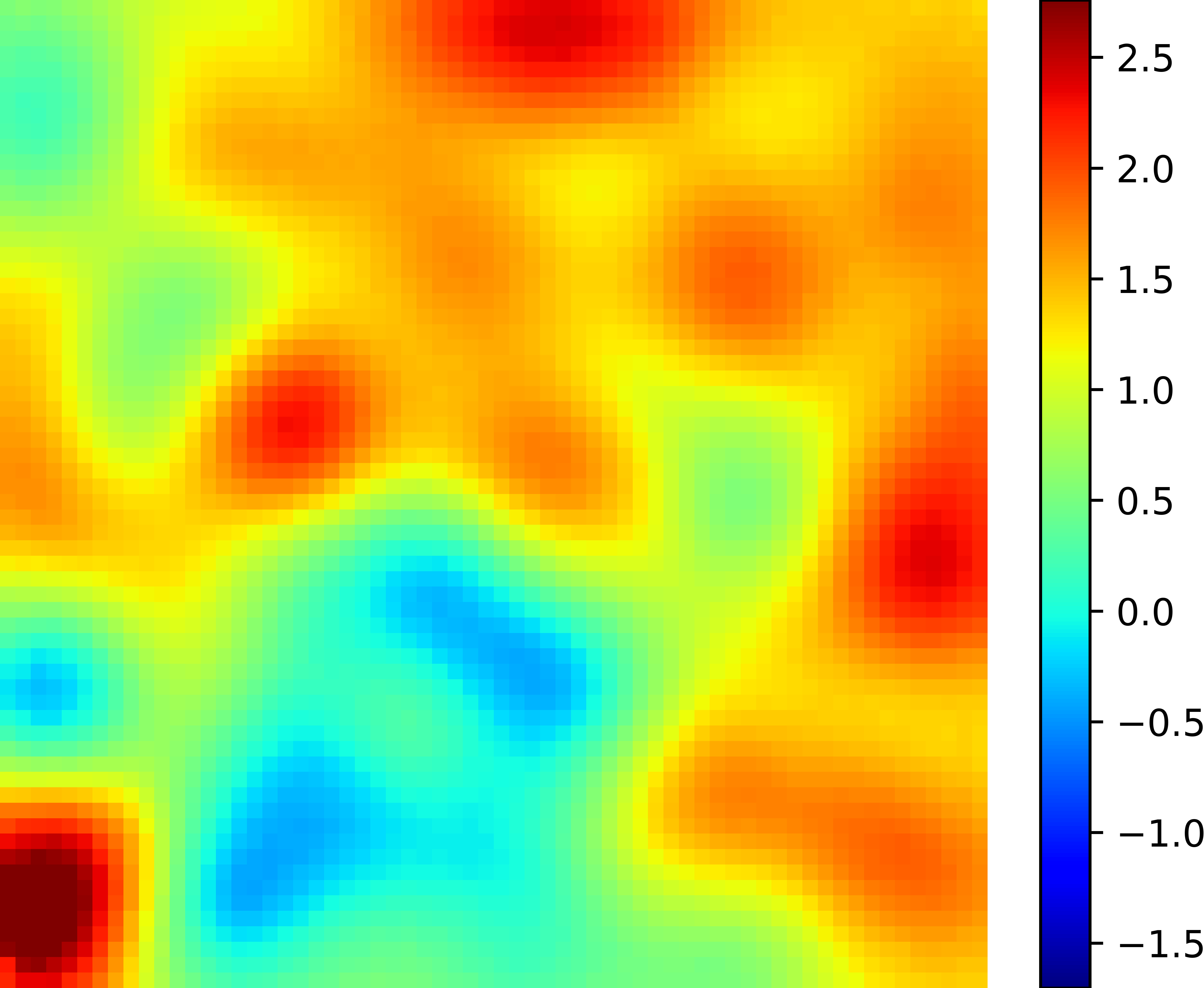}}
	\caption{The inversion results of DR-KRnet for test problem 2. }
    \label{inverse_vae_highdim}
\end{figure}

\begin{figure}
	\centering
	\subfloat[][The exact log-permeability field]{\includegraphics[width=.28
 \textwidth]{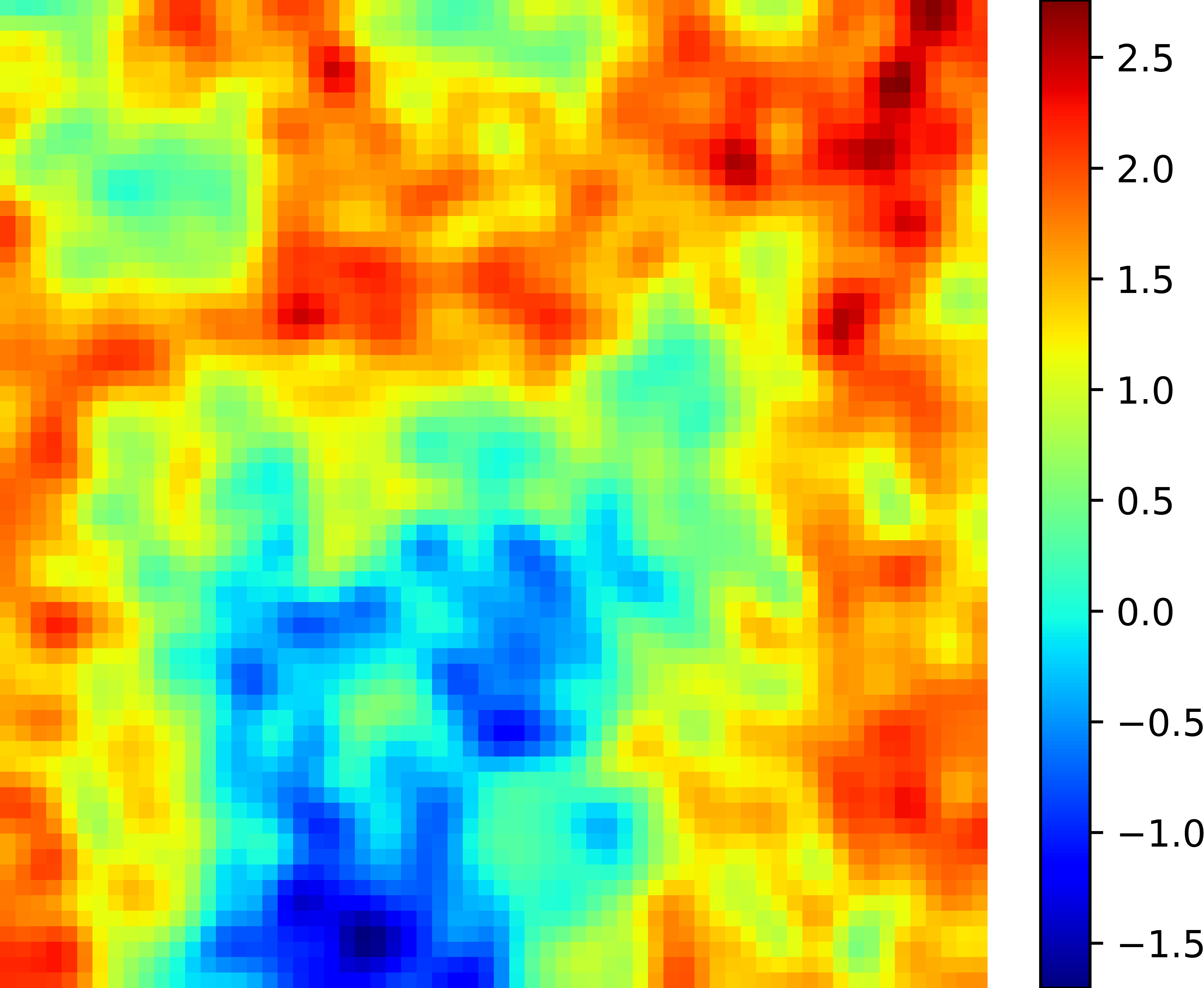}}\quad
	\subfloat[][Posterior mean]{\includegraphics[width=.28\textwidth]{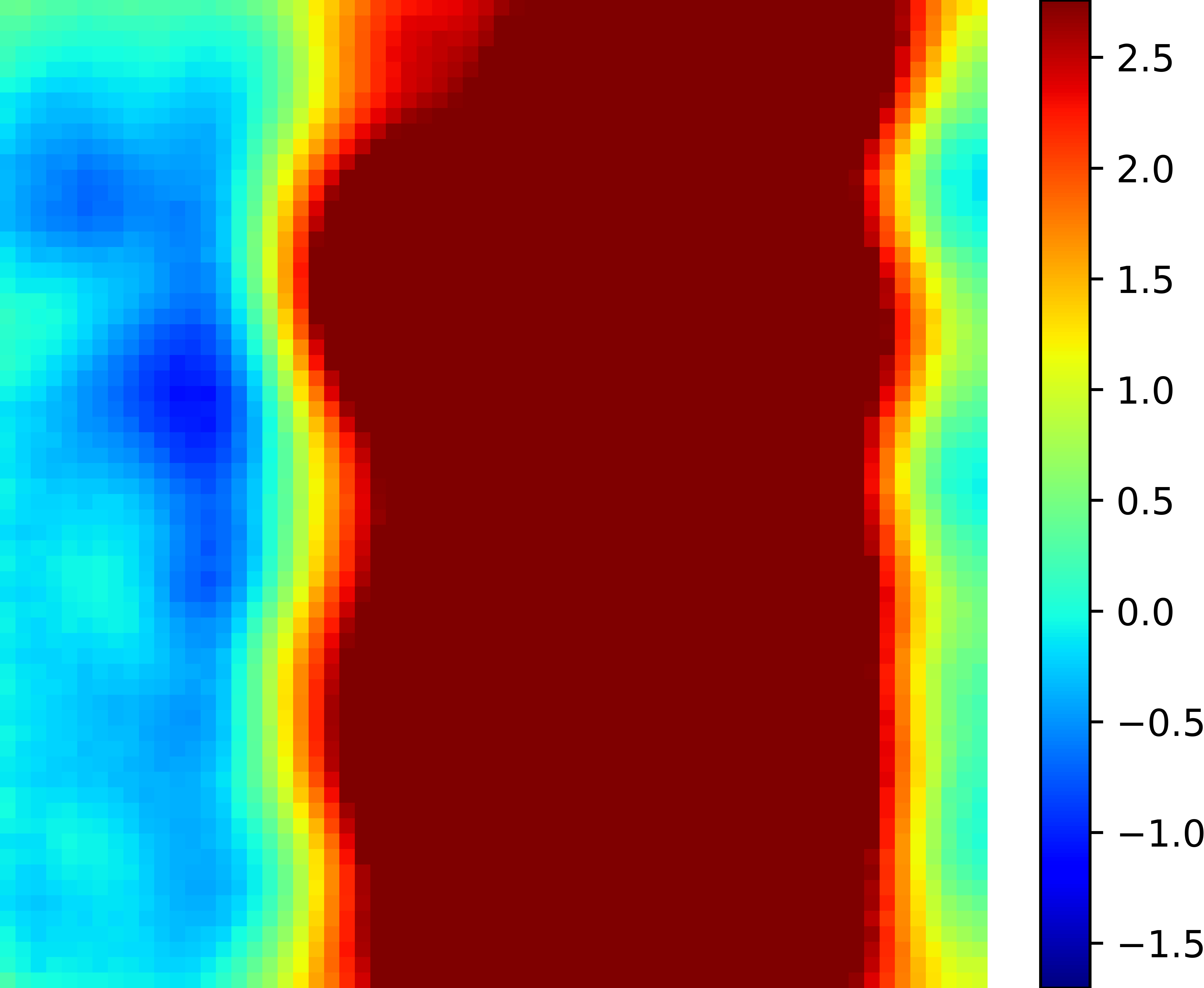}}\quad
	\subfloat[][Posterior variance]{\includegraphics[width=.3\textwidth]{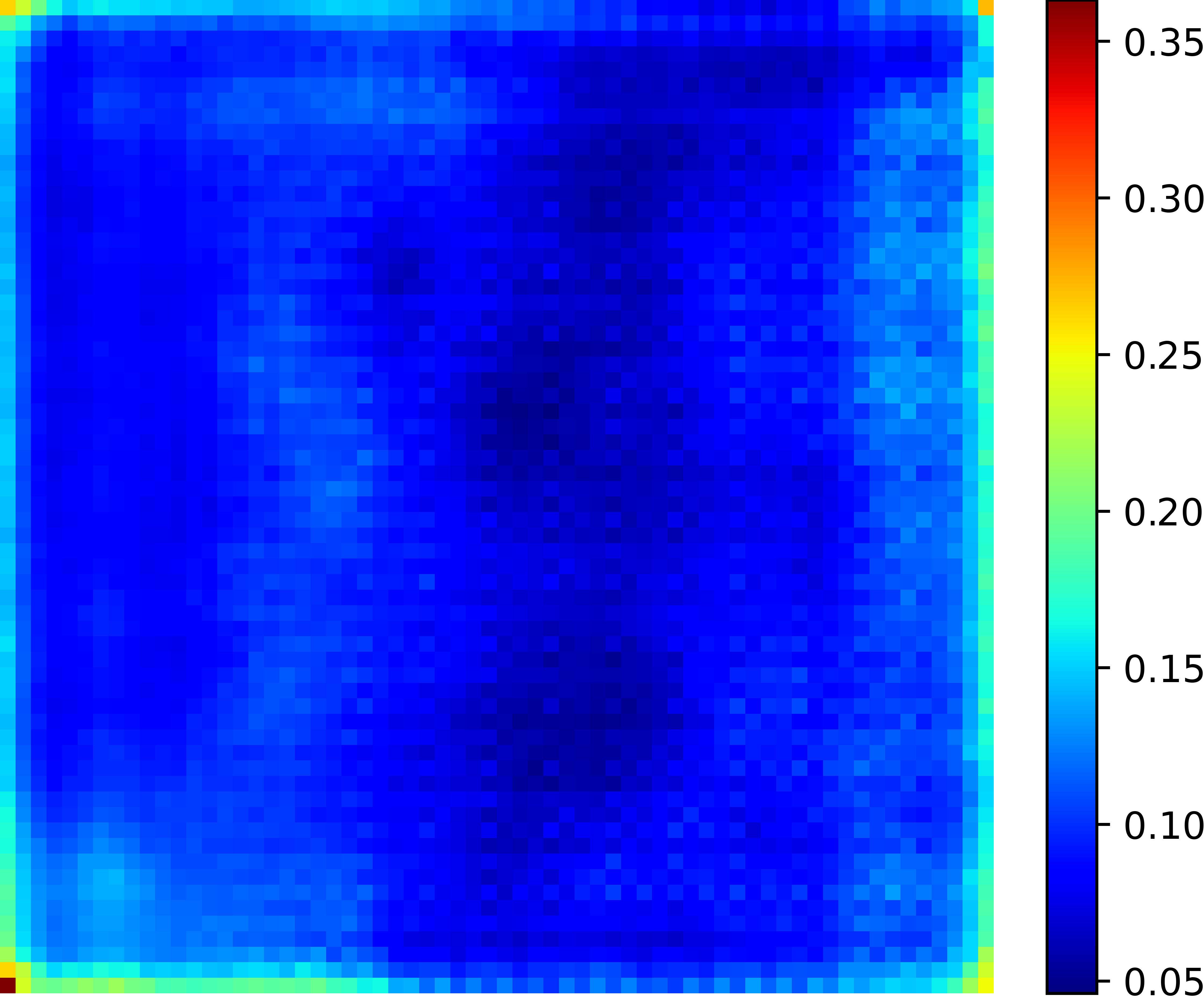}}
	\\
	\subfloat[][Posterior sample 1]{\includegraphics[width=.28\textwidth]{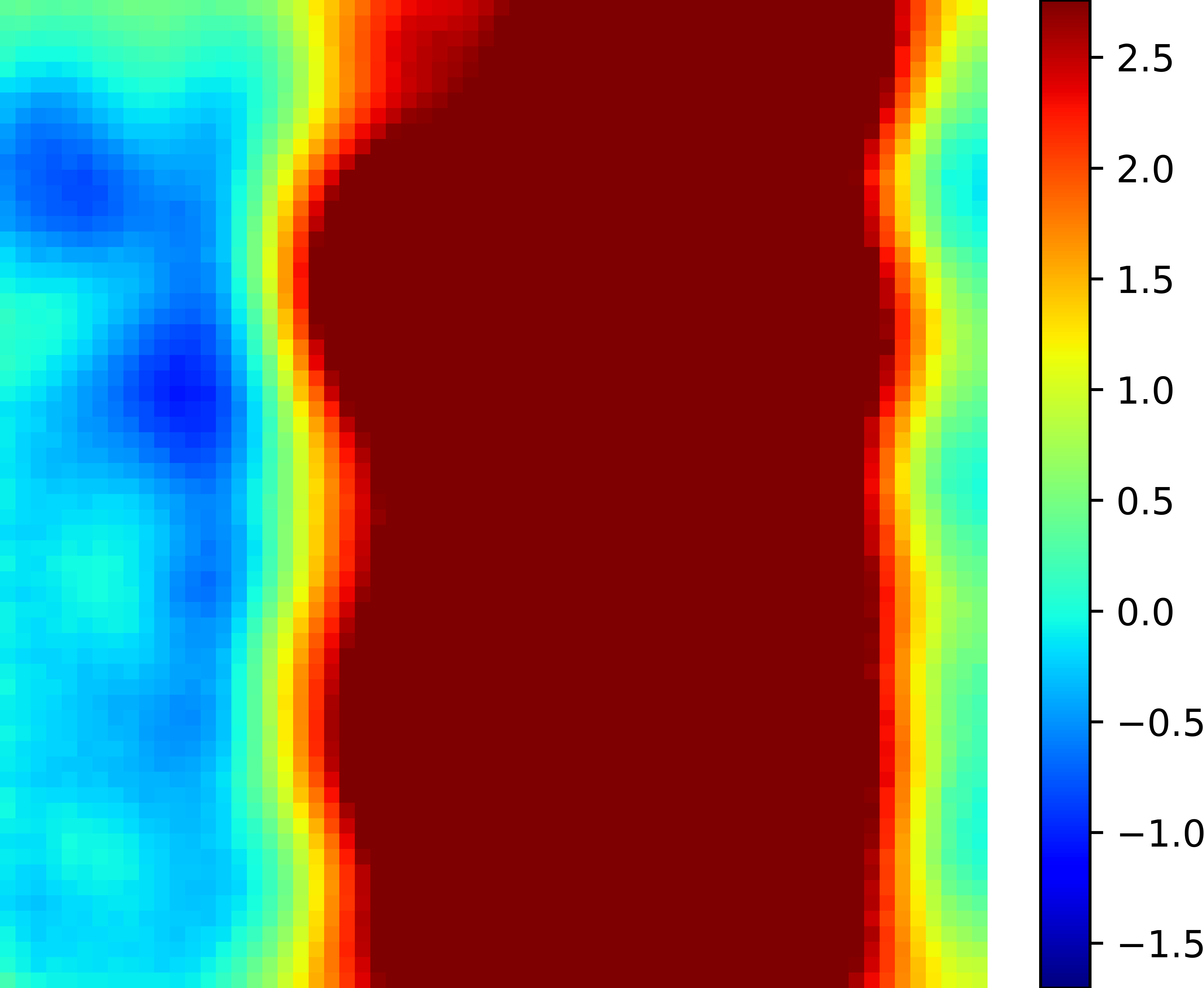}}\quad
	\subfloat[][Posterior sample 2]{\includegraphics[width=.28\textwidth]{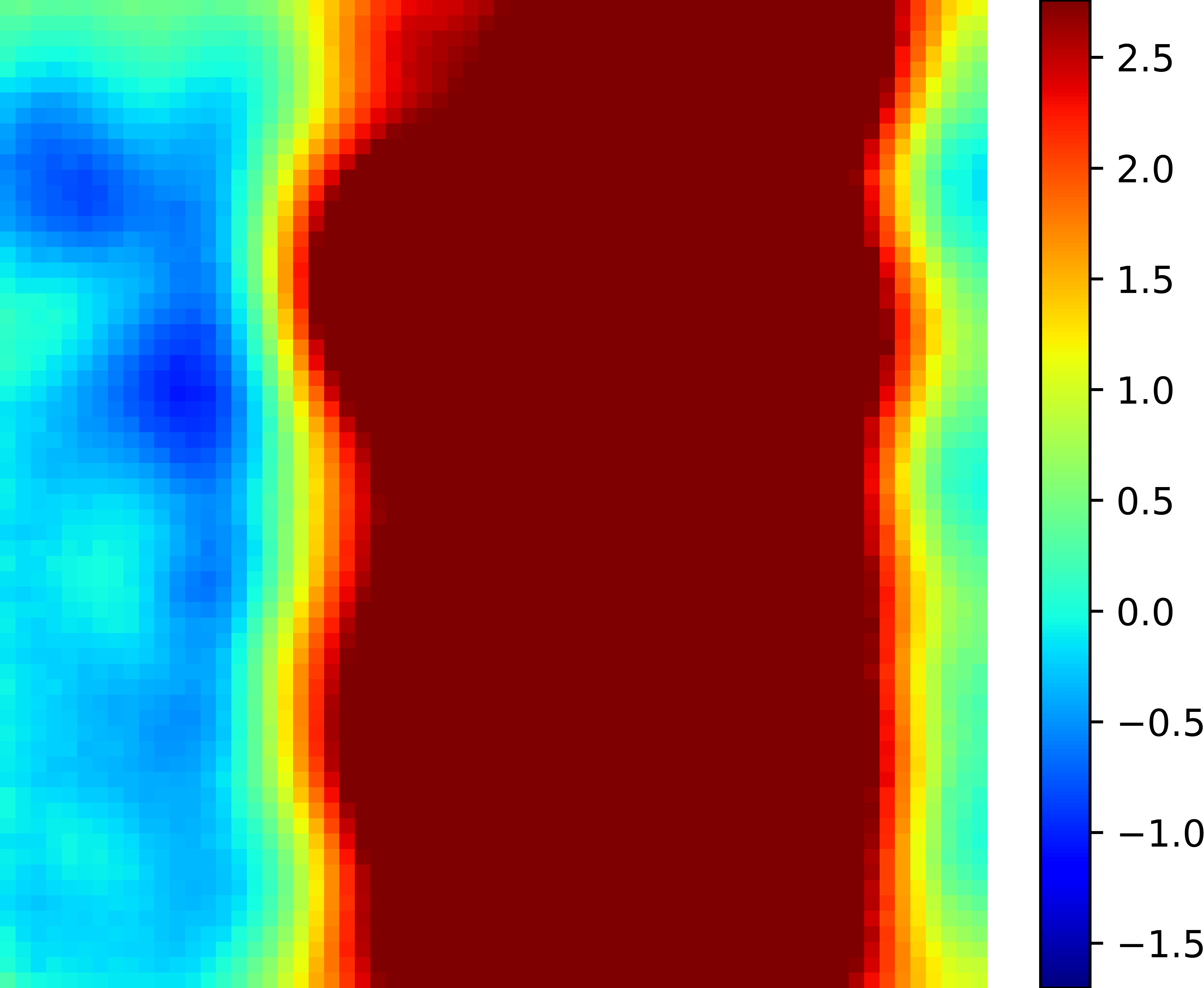}}
   \quad \subfloat[][Posterior sample 3]{\includegraphics[width=.28\textwidth]{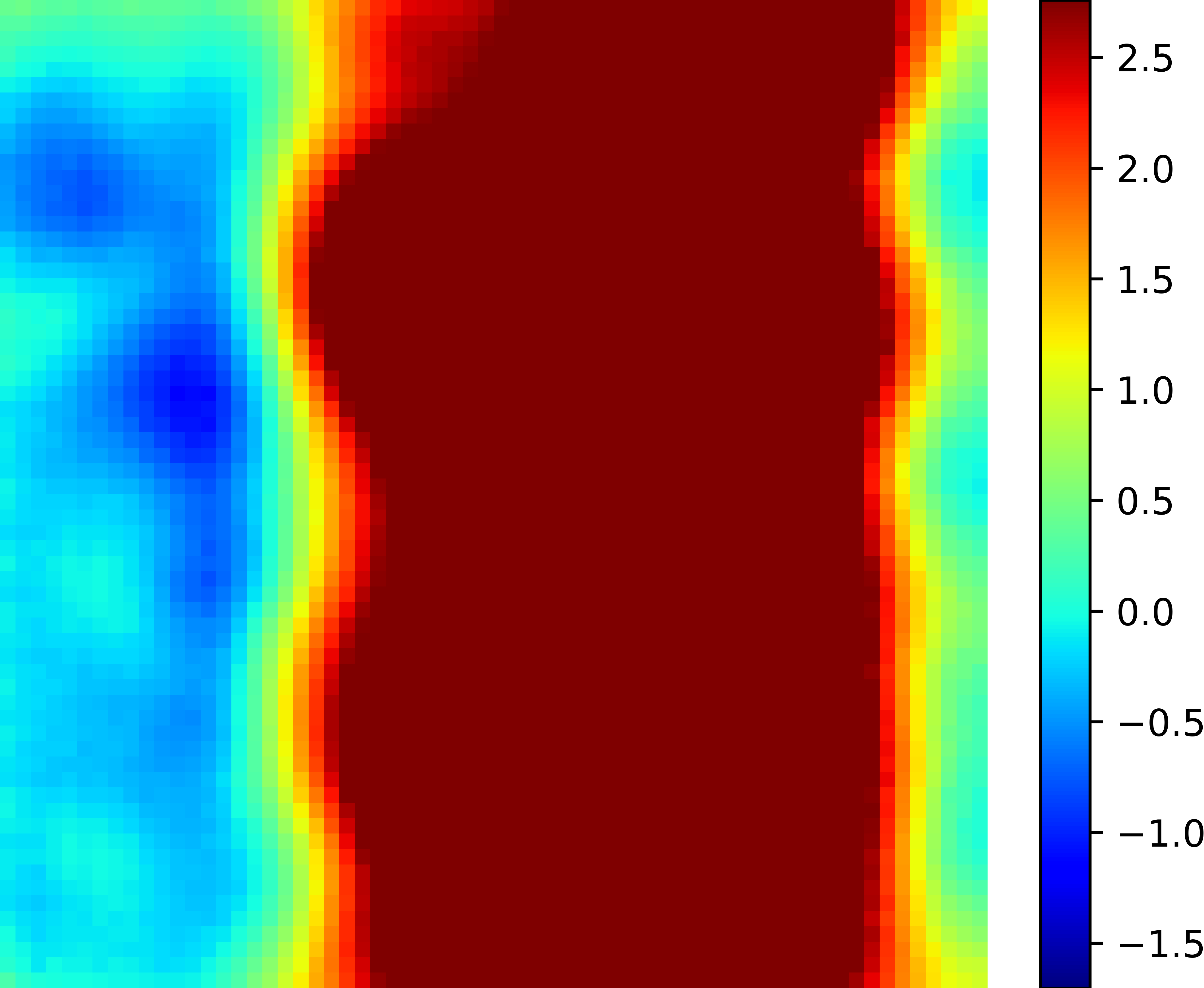}}
	\caption{The inversion results of VAEprior-MCMC for test problem 2. }
    \label{inverse_vae_mcmc_highdim}
\end{figure}

For DR-KRnet, the architecture of the KRnet is the same as that in test problem 1 except that the components of $x\in\mathbb{R}^{64}$ are divided into 8 even groups. 
For VAEprior-MCMC, 
 we run 10000 iterations and then set the last 2000 states as posterior samples,
 and the acceptance rate is $24.07\%$. The inversion results for the two methods are shown in Figures \ref{inverse_vae_highdim}--\ref{inverse_vae_mcmc_highdim}. It is seen that for this case the inversion result of DR-KRnet is consistent with the exact log-permeability but VAEprior-MCMC fails to approximate the posterior of the latent variables.  
 In Table \ref{test22}, more scenarios are considered in terms of the dimension of the latent variable $d$. It is seen that 
DR-KRnet outperforms VAEprior-MCMC in terms of both accuracy and computational cost.

\begin{table}
		\caption{Comparisons of DR-KRnet and VAEprior-MCMC for test problem 2.}
		\label{test22}
		\centering
		\begin{tabular}{ccccc}
			\toprule
			 Model&$d$ & $\epsilon_{relative}$ &Time consumption&Acceptance rate\\
			\midrule
			DR-KRnet &36 &0.3854&1.9842 minutes&-\\
			DR-KRnet & 64& 0.4206&2.6844 minutes &- \\
                \midrule
                VAEprior-MCMC &36 &2.9391&4.9152 minutes&$25.64\%$ \\
			VAEprior-MCMC &64& 2.0616&5.4462 minutes&$24.07\%$\\
                VAEprior-MCMC &128 &2.0760&5.1006 minutes&$27.97\%$\\
			VAEprior-MCMC &256& 2.6581&4.824 minutes& $30.5\%$ \\ 
                VAEprior-MCMC &512& 2.2229&5.388 minutes&$31.8\%$\\
                
   \bottomrule
		\end{tabular}
\end{table}
\section{Conclusions}\label{section_conclude}
We have presented a dimension-reduced KRnet map approach (DR-KRnet) for high-dimensional Bayesian inverse problems, which applies the KRnet to construct an invertible transport map from the prior to the posterior in the low-dimensional latent space of a VAE prior. The key idea of our approach is to employ a deep generative model, called KRnet, to approximate the posterior distribution in the latent space, which allows this approach to incorporate the dimension reduction technique into the Bayesian framework. In this way, the proposed approach can be suitable for practical problems when we only have access to high-dimensional prior data. With the aid of KRnet, our approach can provide an effective and efficient algorithm for both probability approximation and sample generation of posterior distributions. Numerical experiments illustrate that DR-KRnet can solve high-dimensional Bayesian inverse problems. Overall, inference with KRnet maps conducts with greater reliability and efficiency than MCMC, particularly in high-dimensional Bayesian inverse problems. Several promising avenues exist for future work. First, VAE is easy to train and we can couple DR-KRnet with information theory to design new data-driven priors. Second, we can apply our approach to more challenging problems such as petroleum reservoir simulation.

\bigskip
\textbf{Acknowledgments:}
The authors thank Yingzhi Xia for helpful suggestions and
discussions.

\bigskip
\textbf{Funding:}
Y. Feng and Q. Liao are supported by the National Natural Science Foundation of China (No. 12071291), 
the Science and Technology Commission of Shanghai Municipality (No. 20JC1414300), and the Natural Science Foundation of Shanghai (No. 20ZR1436200). K. Tang is supported by the China Postdoctoral Science Foundation under grant 2022M711730,
and X. Wan’s work was supported by the National Science Foundation under grant DMS-1913163.

\begin{appendix}
\section{The neural network architecture of VAE priors}\label{vae_nn}
In section \ref{vae_gan_section}, convolutional neural networks (CNN) are applied to construct the encoder and decoder of VAE priors. Table \ref{vae_ar} presents the neural network architectures of VAE priors, where the dimension of latent variables in test problem 1-2 is $d=36$ and $d=64$, respectively.
\begin{table}[h]
    \centering
    \small
    \caption{The neural network architecture of VAE priors for test problem 1--2.}
     \label{vae_ar}
    \begin{tabular}{|c|c|}
    \hline
       Encoder  & Decoder \\
    \hline
       Input: $y$  & Input: $x$ \\
    \hline
    BatchNormalization&Dense($8*8*48$, activation=`relu')\\
    \hline
    Conv2D(16,2,2,activation=`relu')&Reshape((48, 8, 8))\\
    \hline
    BatchNormalization&BatchNormalization\\
    \hline
    Conv2D(16,3,1,padding=`same',activation=`relu')&Conv2DTranspose(64,3,2,padding=`same',activation=`relu')\\
    \hline
    BatchNormalization&BatchNormalization\\
    \hline
    Conv2D(32,2,2,activation=`relu')&Conv2DTranspose(64,3,1,padding=`same',activation=`relu')\\
    \hline
    BatchNormalization&BatchNormalization\\
    \hline
    Conv2D(32,3,1,padding=`same',activation=`relu')&Conv2DTranspose(32,3,2,padding=`same',activation=`relu')\\
    \hline
    BatchNormalization&BatchNormalization\\
    \hline
    Conv2D(64,2,2,activation=`relu')&Conv2DTranspose(32,3,1,padding=`same',activation=`relu')\\
     \hline
    BatchNormalization&BatchNormalization\\
    \hline
Conv2D(64,3,1,padding=`same',activation=`relu')&Conv2DTranspose(16,3,2,padding=`same',activation=`relu')\\
    \hline 
    Flatten&BatchNormalization\\
    \hline 
Dense($2d$)&Conv2DTranspose(16,3,1,padding=`same',activation=`relu')\\
    \hline
    \multirow{3}{*}{
    Output: $\left(\mu_{en},\log\left(\sigma_{en}^2\right)\right)$}&BatchNormalization\\
    &Conv2DTranspose(2,3,1,padding=`same')\\
    &Output: $\left(\mu_{de},\log\left(\sigma_{de}^2\right)\right)$\\
    \hline
    \end{tabular}
   
\end{table}
\section{The neural network architecture of physics-constrained surrogate model}\label{surrogate_nn}
In this paper, we apply convolutional neural networks (CNN) for physics-constrained surrogate model. The neural network architectures of the surrogate are listed in Table \ref{surrogate_archi_table}. For Darcy flows, the equation loss and boundary loss of the loss function \eqref{surrogate_loss} are defined as:
\begin{align}
    {\left\Arrowvert \mathcal{R}\left(\hat{\mathcal{F}}_{\theta}\left(y^{(i)}\right),y^{(i)}\right)\right\Arrowvert}_2^2={\left\Arrowvert \nabla \cdot\tau\left(y^{(i)}\right)-h\right\Arrowvert}_2^2 +  {\left\Arrowvert\tau\left(y^{(i)}\right)+\exp\left(y^{(i)}\right)\nabla u\left(y^{(i)}\right)\right\Arrowvert}_2^2,\\
    {\left\Arrowvert\mathcal{B}\left(\hat{\mathcal{F}}_{\theta}\left(y^{(i)}\right)\right)\right\Arrowvert}_2^2 =  {\left\Arrowvert u\left(y^{(i)}\right)\right\Arrowvert}_2^2 +  {\left\Arrowvert \exp\left(y^{(i)}\right)\nabla u\left(y^{(i)}\right)\cdot \mathbf{n}\right\Arrowvert}_2^2.    
\end{align}
In addition, the weight $\beta$ in  \eqref{surrogate_loss} set to 100 for test problem 1--2.
\begin{table}[H]
    \centering
    \caption{The neural network architecture of PDE surrogate for test problem 1--2.}
     \label{surrogate_archi_table}
    \begin{tabular}{|c|c|}
    \hline
        Networks& Feature maps  \\
    \hline 
    Input: $y$ &$(1,64,64)$\\
    \hline
    Conv2D&(48,2,2,activation=`relu')\\
    \hline 
    Conv2D&(144,3,1,padding=`same',activation=`relu')\\
    \hline 
    Conv2D&(72,2,2,activation=`relu')\\
    \hline 
    Conv2D&(200,3,1,padding=`same',activation=`relu')\\
    \hline 
    UpSampling2D&2\\
    \hline 
    Conv2D&(100,3,1,padding=`same',activation=`relu')\\
    \hline 
    Conv2D&(196,3,1,padding=`same',activation=`relu')\\
    \hline 
    UpSampling2D&2\\
    \hline 
    Conv2D&(3,3,1,padding=`same',activation=`relu')\\
    \hline
    Output: $(u(y),\tau_1,\tau_2)$ &$(3,64,64)$   \\ \hline
    \end{tabular}
\end{table}

\end{appendix}


\bibliography{feng}

\end{document}